\documentclass{brownthesis}
\usepackage[margin=1.2in]{geometry} 
\usepackage{graphicx} 

\usepackage[authoryear,round]{natbib}
\usepackage[bookmarks=true,hidelinks]{hyperref}

\usepackage{amsmath,amssymb}
\usepackage{stfloats}
\usepackage[dvipsnames]{xcolor}
\usepackage{algorithm}
\usepackage{algorithmic}
\usepackage{caption}
\usepackage{lettrine}
\usepackage{booktabs}
\usepackage{makecell}
\usepackage{siunitx}
\usepackage{listings} 
\usepackage{wrapfig}
\usepackage{xspace}
\usepackage{multirow}
\usepackage{tabularx}
\definecolor{codeblue}{rgb}{0.38039216, 0.61568627, .8       }
\definecolor{codepink}{rgb}{1.        , 0.41568627, 0.83529412}
\definecolor{codegray}{rgb}{0.5,0.5,0.5}
\definecolor{codepurple}{rgb}{0.58,0,0.82}
\definecolor{backcolour}{rgb}{0.98,0.98,0.98}
\lstdefinestyle{mystyle}{
    backgroundcolor=\color{backcolour},   
    commentstyle=\color{codeblue},
    keywordstyle=\color{codepink},
    numberstyle=\tiny\color{codegray},
    stringstyle=\color{codepurple},
    basicstyle=\ttfamily\footnotesize,
    breakatwhitespace=false,         
    breaklines=true,                 
    captionpos=b,                    
    keepspaces=true,                 
    numbers=left,                    
    numbersep=5pt,                  
    showspaces=false,                
    showstringspaces=false,
    showtabs=false,                  
    tabsize=2
}
\lstdefinelanguage{gptprompt}{
  basicstyle=\ttfamily\scriptsize,
  keywords = {generated_example_instruction,generated_example_LTL, given_instruction, stage1_ltl_response, last_response, given_error_clarification, verification_type},
  keywordstyle=\color{red},
  keywords = [2]{isbetween, isabove, isToTheLeft, isbelow, isleftof, isrightof, isnextto, isinfrontof, isbehind},
  keywordstyle= [2]\color{Fuchsia},
  keywords = [3]{Input, Output, Input_instruction, Input_ltl, previous_output, error_clarification, correct_output},
  keywordstyle = [3]\color{blue},
  keywords = [4]{near, pick, release},
  keywordstyle = [4]\color{TealBlue},
}

\lstdefinelanguage{instructions}{
  basicstyle=\ttfamily\scriptsize,
  keywords = {NLMD,NLMC,OKRB,CT,CST},
}

\newcommand{\approachname}{\textsc{Pacts}}
\newcommand{\method}{\textsc{Realm}\xspace}

\newcommand{\objects}{\mathcal{O}}
\newcommand{\gaussians}{\mathcal{G}}
\DeclareMathOperator{\SE}{SE}

\title{Scaling Trusted Autonomy: Towards Generally Capable Verifiable Robots}
\author{Benedict Quartey}
\dept{Computer Science} 
\principaladvisor{George Konidaris} 
\reader{Stefanie Tellex} 
\reader{Srinath Sridhar} 
\dean{David P. Lindstrom} 
\submitdate{October 2026} 
\copyrightyear{2026} 

\abstract{Robots powered by foundation models are increasingly able to interpret open-ended language, recognize unfamiliar objects, and operate across diverse environments. Yet broader capability does not by itself establish that a robot has understood a request, that its actions will produce the intended effects, or that its internal representation still corresponds to reality. Classical symbolic planning provides explicit structure for composition and verification, but its application to robotics has traditionally depended on hand-engineered skills and representations specified in advance. Learned policies can broaden the robot’s repertoire, but provide weak guarantees about what will happen and why. This thesis argues that scaling trusted autonomy requires a neurosymbolic approach that unifies the generality of learned models with the symbolic structure needed for robust robot behavior. It explores three complementary directions toward this goal. First, LIMP uses foundation models and spatial reasoning to ground natural-language instructions in explicit task constraints and instruction-specific semantic maps, enabling task-and-motion planning while exposing the robot’s interpretation for inspection. Second, REALM extends this framework to an evolving world. It maintains an editable, photorealistic 3D Gaussian scene, predicts the consequences of planned manipulation, and locally reconciles the scene so that subsequent instructions obtain their grounding and planning geometry from an updated state. Third, PACTS learns visuomotor skills that jointly generate action trajectories and predictions of their symbolic effects, supporting skill composition, progress monitoring, and recovery through replanning. Together, these contributions provide mechanisms for constructing task-relevant abstractions when needed, connecting learned behavior to its predicted symbolic effects, and maintaining an evolving world representation for subsequent tasks. This body of work preserves explicit structure for reasoning and verification while broadening how it is obtained, providing a foundation for generally capable robots whose behavior can be assessed under the assumptions of their models.}

\begin{document}


\beforepreface

\thispagestyle{plain}
\vspace*{\fill}
\begin{center}
  \scalebox{1.3}{\itshape For Aleen, my beautiful wife, and my parents, Janis and Ignatius Quartey}
\end{center}
\vspace*{\fill}
\clearpage

\abstractpage

\prefacesection{Acknowledgments}
\begingroup
\clubpenalty=10000
\widowpenalty=10000
Long before this dissertation took shape, there were people who believed in me and helped me find my way into research. I am deeply grateful to them, and to everyone who has made these years so meaningful.

My advisor, George Konidaris, took a chance on a young man from Ghana during a pandemic that shook the entire world. We had never met in person, and my path into research had been far from conventional. Yet, George looked beyond what could be seen on paper and believed in what I could become. He gave me the opportunity to come to Brown and supported me at every stage of this PhD. I am eternally grateful for his support, advice, and mentorship.

Stefanie Tellex has been an extraordinary mentor at Brown and beyond. Stefanie gave me a shot to do research at the Robotics and AI Institute, believing in my ideas before they could materialize; her encouragement and mentorship have been invaluable. I also thank Srinath Sridhar, whose lessons and intuitive explanations gave me the foundations to pursue the vision work in this thesis, and Michael Littman, who introduced me to reinforcement learning in my first year and served on my first candidacy committee.

Many years ago during my undergraduate studies at Ashesi, Dr. Ayorkor Korsah introduced me to research and helped me see a PhD as a possibility. Thank you for encouraging that curiosity before I knew where it might lead.

I am also very grateful to Eric Rosen, who worked with me on my first major PhD paper; I learned a great deal from our collaboration. Thank you for sharing your time and knowledge so generously. To Sebastian Castro, Wil Thomason, and my colleagues at the Robotics and AI Institute: thank you for making my time there so rewarding. Working together was a pleasure and I will always value the friendships that grew alongside research.

To my friends, labmates, and colleagues in the Intelligent Robot Lab and the Humans to Robots Lab, thank you for the conversations, shared enthusiasm, and laughter that made Brown a place where I belonged.

Wasiwasi and Amina, we began our PhDs together as three Africans finding our way in a new place. You have grown from friends into family, and I am grateful that we have shared so much of this experience. Tuluhan and Yong, our friendship began even before our Ph.D. journey. Being roommates and spending those early years together gave me brothers, not just friends. Thank you for the companionship and all the memories I will carry with me.

To my mom and dad, thank you for the love and faith that have sustained me throughout these years. You nourished my curiosity from childhood and went above and beyond, sacrificing so much to give me the opportunities that led me here. Thank you so much. Living so far from home has made me appreciate even more how much of who I am comes from you. I hope you can feel your part in this achievement. To my friends near and far, thank you for keeping me connected to the people and places I love.

Finally, to my wife, Aleen: thank you for your patience through the long nights and for the love that has filled these years with happiness. So much of the joy and enthusiasm that others see in me begins with you. You have shared the demands of this PhD with a generosity I will never take for granted. I am grateful that I get to share this moment, and my life, with you. I love you. This dissertation is for you and for our family. Thank you Jesus, we made it!
\par\clearpage
\endgroup

\afterpreface 
\pagestyle{plain} 
\clubpenalty=10000
\makeatletter
\@clubpenalty=10000
\makeatother
\widowpenalty=10000



\chapter{Introduction}
\label{ch:introduction}

\begingroup
\clubpenalty=10000
\makeatletter
\@clubpenalty=10000
\makeatother
\widowpenalty=10000

A central ambition of robotics is to build machines that can perform a wide range of useful tasks in the environments people inhabit.
Such robots would allow people to specify goals and constraints without prescribing each movement or engineering a separate system for every activity.
Their usefulness would depend on the ability to interpret unfamiliar requests, apply learned abilities in new circumstances, and sustain purposeful behavior as their surroundings change.
The challenge is to make this breadth of capability compatible with behavior that people can understand, assess, and rely upon.

Foundation models have opened new possibilities for pursuing this goal.
Their training on broad collections of data allows knowledge acquired across tasks and settings to inform the behavior of individual robot systems~\citep{bommasani2021foundation}.
They support interfaces through which people can describe tasks and objects beyond a fixed vocabulary, and connect pretrained language and visual knowledge to robot control across diverse tasks and settings~\citep{minderer_simple_2022,brohan2023rt,kim2024openvlaopensourcevisionlanguageactionmodel}.
This progress changes what can reasonably be asked of a robot.
It also brings a longstanding requirement into sharper focus: as robots undertake a broader range of activities, there must be a corresponding basis for assessing whether their behavior meets the requirements of those activities.

Greater capability does not automatically provide that assurance.
A robot may respond fluently to an instruction while leaving its interpretation implicit, or produce effective movements without exposing the conditions those movements establish.
The consequences matter when people delegate tasks in homes and workplaces, where intentions must be communicated, constraints respected, and decisions made from incomplete information.
The scientific problem is therefore to extend the scope of autonomous behavior while retaining the means to examine what the robot understood, reason about what it will do, and assess whether continued reliance on its decisions is warranted.
Generality and assurance must advance together.

Classical planning provides an important foundation for this objective.
Explicit task representations and action models support the composition of behaviors, while formal specifications make requirements available for checking and behavior synthesis~\citep{fikes_strips_1971,kress-gazit_temporal-logic-based_2009}.
Task and motion planning connects this reasoning to geometric feasibility~\citep{garrett_integrated_2021}.
These strengths, however, depend on suitable descriptions of the task, the robot's skills, and the world.
Constructing and maintaining those representations across varied environments has traditionally required substantial engineering, limiting the settings in which structured reasoning can be applied. 
Learned models offer ways to broaden perception and behavior; planning offers structure for reasoning about their use.
The challenge is to bring these strengths together without allowing the assumptions that support reliable reasoning to become hidden inside increasingly capable components.

\begin{samepage}
The central research question of this thesis is therefore: \emph{how can robot capability be extended across tasks and environments while preserving the explicit structure needed to interpret, compose, and check its behavior?}
This thesis develops an answer through grounded language understanding, composable state-reporting skills, and evolving representations of the world.
These contributions connect learned capabilities to representations that a planner can inspect and use.
\par
\end{samepage}

\section{From Tasks to Robot Behavior}
\label{sec:intro:paradigms}

Two complementary approaches help explain both the opportunity and the difficulty of this problem.
Planning constructs behavior by reasoning over a model of the task and environment, whereas policy learning acquires behavior from experience.
The former makes the organization of behavior explicit; the latter provides a way to acquire the perception and control needed to realize it.
Understanding their respective strengths requires examining how each turns a requested outcome into physical action, and how knowledge acquired for one task becomes useful in another.

\subsection{Explicit Structure and Compositional Planning}
\label{sec:intro:planning}

A physical scene contains far more information than is needed for any particular decision.
Symbolic planning makes reasoning manageable by representing the objects, properties, and relationships relevant to a task, together with models of the actions that can change them~\citep{fikes_strips_1971}.
Predicates express conditions that can be evaluated as true or false, such as whether an object is on a surface or held by the robot.
An action model specifies the conditions under which an action is applicable and the effects it is expected to produce.
The initial state describes the conditions that currently hold, while the goal describes those the planner must establish.
This abstraction allows the planner to reason about changes in the world without representing every detail of their physical realization.

For example, the instruction ``bring the red book on the table to the sofa'' can be expressed using objects for the book, table, and sofa, and predicates describing the book's location and whether the robot is holding it.
Picking up the book might require an available gripper and a reachable book; its modeled effects include holding the book and removing it from the table.
The planner can then search for a sequence in which the effects of each action establish the conditions required by the next: approach the table, pick up the book, approach the sofa, and place the book.
Figure~\ref{fig:intro:planning} illustrates this connection between a requested outcome and a compositional description of behavior.

\begin{figure}[htbp]
    \centering
    \includegraphics[width=\textwidth]{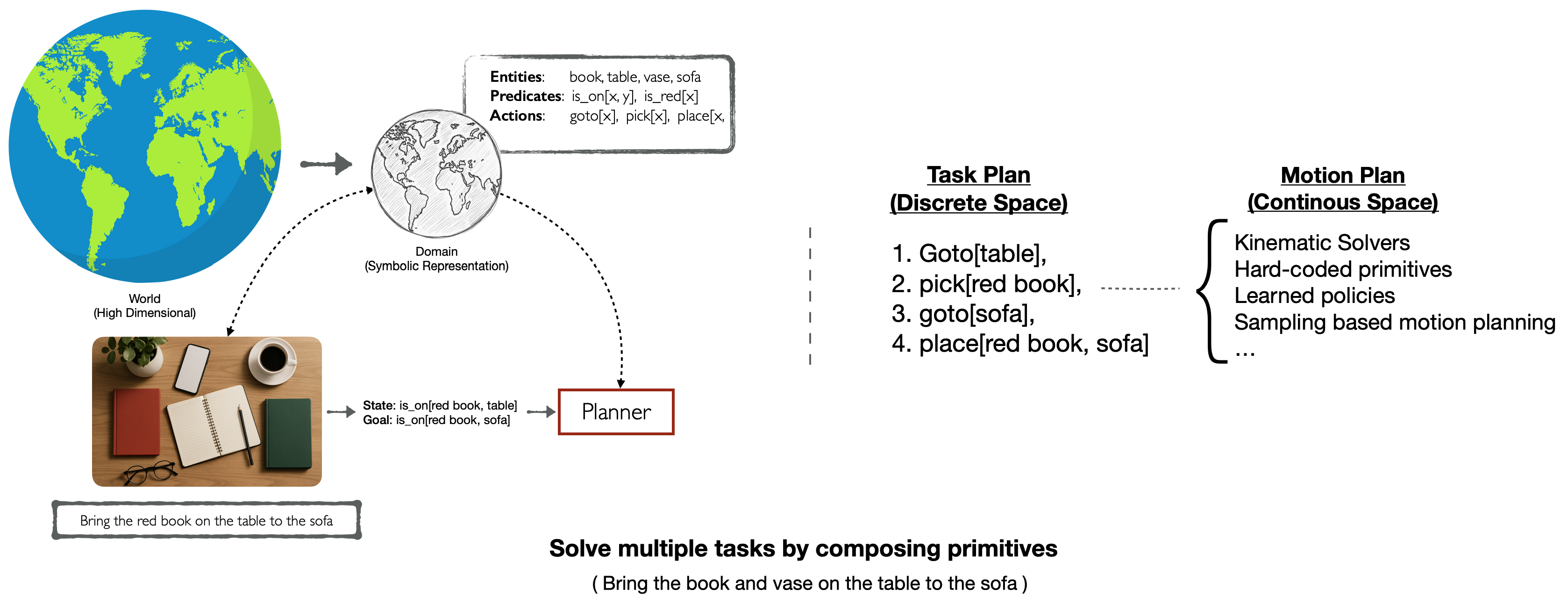}
    \caption[Explicit representations for the book-transfer task]{Explicit representations for the book-transfer task. A symbolic description connects the observed scene and desired goal to a sequence of task-level actions, whose physical realization requires motion planning and control. The book-and-vase extension illustrates how the same action models support a new composition. Learned policies can implement individual skills within this structure.}
    \label{fig:intro:planning}
\end{figure}

\begin{samepage}
The value of this representation extends beyond finding one successful sequence.
If the request changes to transporting both the book and the vase, the planner can reuse the same action models with different object instances and arrange their applications to satisfy the new goal.
There is no need to store a separate procedure for every combination of familiar objects and goals: the model describes reusable relationships between conditions and actions~\citep{konidaris2018skillstosymbols}.
This form of composition gives structured planning its own source of generality: a domain can support many tasks that were never individually specified as complete behaviors.
\par
\end{samepage}

\begin{samepage}
Physical realization requires a further connection between discrete decisions and continuous control.
The robot must select grasps, configurations, and collision-free motions that support the proposed sequence; a failed geometric search may require the task plan itself to change.
Task and motion planning reasons across these levels~\citep{garrett_integrated_2021}.
Its actions can be implemented through reusable skills, such as grasping or navigation, that unfold over multiple control steps and have conditions governing when they can begin and terminate~\citep{sutton_between_1999}.
A skill may use an analytic controller, a motion planner, or a learned policy.
Planning and learning can therefore contribute to the same behavior, with planning organizing the skills and learning supplying some of their control policies.
\par
\end{samepage}

Explicit task structure also permits requirements to shape the course of an activity.
For the book-transfer task, reaching the sofa while avoiding a specified region is a different requirement from merely ending at the sofa.
A final-state goal captures the destination but does not, by itself, express every constraint on the intervening behavior.
Formal methods ideas like temporal logic provide a language for requirements such as ordering and avoidance, and support the synthesis of behavior that satisfies a specification under a given model~\citep{pnueli_temporal_1977,kress-gazit_temporal-logic-based_2009}. Planners can thus reason about both what should be achieved and which sequences of events are acceptable.

The reach of this reasoning depends on the representations supplied to it.
The planning system needs a formal task grounded in the relevant objects and skills equipped with appropriate action models.
A prebuilt semantic map can supply associations between spatial structure and meaning, but future requests may refer to properties or objects that were not recorded when the map was created.
Such a map is not inherently closed-world: that assumption arises when represented facts are treated as complete and unrepresented facts as false.
Even a detailed map may require revision as the environment changes.
Extending planning to less structured settings therefore requires ways to construct and maintain its inputs without engineering them anew for every task.

\subsection{Learning Behavior from Experience}
\label{sec:intro:learning}

Learning from demonstration offers a way to acquire control from examples of successful behavior.
For the same book-transfer task, an expert supplies trajectories containing observations and robot actions, from which a visuomotor policy learns to produce actions in response to visual input and, where available, robot state and a task description~\citep{robomimic2021}.
Figure~\ref{fig:intro:learning} illustrates this route from demonstrated experience to a control policy.
During deployment, a closed-loop policy repeatedly receives current observations, allowing its movements to respond to the scene as execution proceeds.
The acquired behavior is consequently a mapping from situations to actions, rather than a recording of one sequence of movements.

\begin{figure}[htbp]
    \centering
    \includegraphics[width=\textwidth]{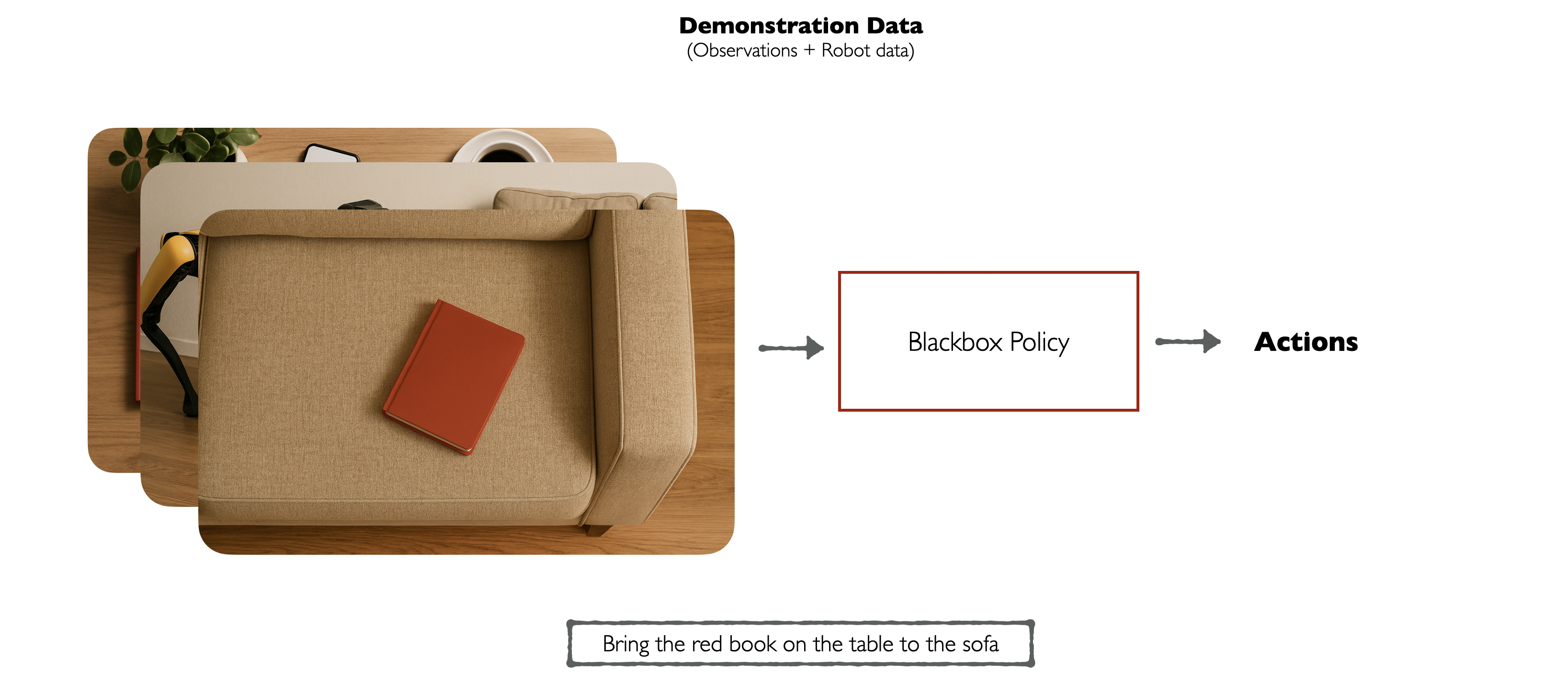}
    \caption[Learning the book-transfer behavior from demonstrations]{Learning the book-transfer behavior from demonstrations. Observation--action examples train a policy that produces robot actions. The schematic emphasizes behavior acquisition; a deployed closed-loop policy receives current observations. Comparing this output with Figure~\ref{fig:intro:planning} identifies the additional task and symbolic-state information needed to use the learned behavior within a planner.}
    \label{fig:intro:learning}
\end{figure}

\begin{samepage}
A useful control representation must accommodate the fact that several different movements can accomplish the same task.
There may be multiple valid approaches to an object, and averaging their trajectories need not yield a valid approach.
Generative policies address this multiplicity by modeling distributions over actions or action sequences.
Diffusion Policy, for example, generates observation-conditioned action trajectories and executes them through receding-horizon control~\citep{chi2024diffusionpolicy}.
By generating a trajectory, executing part of it, and then responding to another observation, the policy combines coordinated movement with feedback.
Such methods provide a basis for acquiring manipulation skills whose detailed motions would be difficult to specify by hand.
\par
\end{samepage}

The same policy can also cover several tasks when its output is conditioned on the requested behavior.
A task identifier can select an activity, a goal image can depict an outcome, and language can describe what should be accomplished.
In the shared example, conditioning distinguishes bringing only the book from bringing both the book and the vase.
How well the policy responds to an unfamiliar request depends on the coverage of its training experience and its ability to generalize beyond that experience.
This makes the source and organization of training data central to the breadth of learned behavior.

Longer activities place additional demands on that behavior.
Errors can change the observations encountered later in an execution, taking a policy into situations poorly represented in its demonstrations~\citep{ross2011reduction}.
Feedback may correct a deviation within a skill, while recovery at the task level may require selecting a different skill or revising their order.
A planner making that decision needs to know which relevant conditions hold and which actions can restore progress.
An action-producing policy may use such information internally without exposing it in a form that another component can interpret.
The ability to perform a behavior and the ability to supply the information needed to compose it are therefore distinct aspects of a learned skill.

\section{The Capability--Assurance Gap}
\label{sec:intro:assurance}

Planning allows familiar actions to be composed for new goals; learning broadens the behavior a robot can acquire from experience.
Both extend what the robot can do, but neither alone establishes whether a particular behavior satisfies the requirements of the task.
Capability concerns what a robot can accomplish and how broadly that competence extends across tasks and conditions.
Assurance concerns the evidence and reasoning that support confidence in its behavior under stated conditions.
For example, recognizing the objects mentioned in an instruction does not establish that the robot has resolved the instructor's intended referents, and a high task success rate does not establish compliance with every constraint in a particular execution.
An account of dependable autonomy must connect evidence about individual capabilities to the requirements of the activity in which they are used.

Symbolic representations make these requirements available for reasoning, but the conclusions of that reasoning remain conditional on the model.
A plan can satisfy a specification even when the specification misrepresents the instruction; similarly, a valid transition in an action model may fail to describe what occurred during execution.
Closing the capability--assurance gap requires connecting an instructor's intended task to the robot's interpretation, and the effects anticipated by the planner to what happens during execution.
The interfaces between perception, planning, and control must expose enough information to check these connections.

This thesis uses \emph{trusted autonomy} to describe the objective of making reliance on task-level robot behavior warranted by such evidence and reasoning.
The term \emph{verifiable} refers to explicit specifications and representations against which aspects of behavior can be checked.
Formal reasoning establishes properties within a model, while empirical evaluation examines how well grounding, control, and state predictions correspond to the environment.
These forms of evidence support different claims; together, they identify both the basis for a decision and the assumptions on which that basis depends.

\section{Three Interfaces for Trusted Autonomy}
\label{sec:intro:interfaces}

These connections raise three questions about the information on which a robot relies.
Before planning, the robot needs a representation of what the user intends.
During execution, it needs to determine whether the effects of its actions support continued progress.
Across actions and successive tasks, it needs to maintain the world state on which further interpretation and planning depend.
These responsibilities define the \emph{Intent}, \emph{Action}, and \emph{World} interfaces developed in this thesis.
Each interface addresses a different part of the same activity: interpreting the task, assessing progress, and maintaining the context for further decisions.

\subsection{Intent: Did the Robot Understand What the User Meant?}

Language communicates a task through descriptions whose meaning depends on the environment and on their relationships to one another.
In the book example, ``the red book on the table'' identifies an instance through an attribute and a spatial relation.
An instruction may additionally constrain the order of operations or specify what the robot must avoid.
Grounding must connect these expressions to objects, locations, and behavioral requirements, so that the planner reasons about the requested task.

An explicit task specification exposes the interpreted requirements, while grounded referents make the selected objects and locations available for inspection.
This permits the interpretation to be examined separately from the plan constructed to satisfy it.
The research problem at the intent interface is to obtain these representations from expressive language and perceptual observations without assuming that every relevant concept was encoded in a complete semantic map beforehand.
The corresponding contribution is \emph{grounded language understanding}.

\subsection{Action: Did the Robot's Actions Produce the Intended Effects?}

Composition relies on the effects of one skill establishing the conditions required by another.
For the robot to carry the book, its preceding grasp must establish that the book is held; a command to close the gripper is not sufficient evidence of that condition.
The planner needs a state description at the level of its action models, even when the skill itself operates through images and continuous motor commands.
This creates an interface between the quantities used for control and the conditions used for task reasoning.

For learned behavior, the research problem is to acquire skills that expose those conditions as execution unfolds.
The thesis develops \emph{composable state-reporting skills}, which generate control actions together with predictions of the associated symbolic state.
Such reports allow a planner to monitor expected effects and revise a skill sequence using updated state estimates.
Their usefulness depends on how well they agree with the consequences of the generated actions: evaluating predicate accuracy alone does not establish action--predicate coherence.
The symbolic report must therefore be studied as part of the skill's behavior.

\subsection{World: Does the Robot's Internal World Still Match Reality?}

Task-level state reports do not by themselves maintain the spatial context required for sustained autonomy.
A later instruction may refer to an object by its relationship to other objects, while geometric planning needs its current shape and location.
Although these decisions use different abstractions, they must draw on a compatible account of the environment.
As the robot intervenes in that environment, the representation must preserve relevant changes, object identity, and unaffected scene structure across tasks.

The research problem at the world interface is to maintain that continuity through a representation that can predict task-relevant changes and supply updated inputs to grounding and planning.
This motivates \emph{evolving representations of the world}, in which action-conditioned transitions connect the state used for one task to the state available for the next.
The world interface provides the persistent spatial context that complements the action interface's reports of skill-level conditions.
For these interfaces to support continued operation in the physical world, predicted effects must be reconciled with observations, so that later decisions reflect what actually happened.

\section{Thesis and Contributions}
\label{sec:intro:contributions}

The central argument of this dissertation is:
\begin{quote}
\emph{Scaling trusted autonomy requires unifying grounded language understanding, composable state-reporting skills, and evolving representations of the world.}
\end{quote}
The common design principle is to introduce explicit symbolic and spatial structure where learned outputs become inputs to robot reasoning.
At these interfaces, a system can expose its interpreted requirements, expected action effects, and persistent world state for inspection, composition, and checking.
The contributions develop this principle through three complementary forms of generality: broader instruction grounding, continuity across world changes, and new compositions of learned skills.

The technical chapters follow the order \textbf{LIMP $\rightarrow$ REALM $\rightarrow$ PACTS}.
LIMP establishes the language-to-plan framework and exposes two assumptions that limit its continued use: a world representation that remains valid after mapping, and an available library of skills with suitable planning interfaces.
REALM addresses the first by evolving the world state used for subsequent tasks; PACTS addresses the second by learning skills that expose planning-relevant state.

Chapter~\ref{ch:background} develops the shared foundations needed to examine the contributions: foundation models, task specifications, task and motion planning, semantic grounding, spatial world representations, and learned visuomotor skills.
These concepts provide the basis for the technical chapters, where each method is also situated within its related work.
Chapters~\ref{ch:limp}--\ref{ch:pacts} present LIMP, REALM, and PACTS in the dependency order described above.
The concluding Chapter~\ref{ch:scaling} brings these contributions together around the construction and maintenance of task-relevant abstractions.
It returns to the capability--assurance question through the three trust interfaces and develops a research agenda for continued autonomy, including physical integration, uncertainty management, and recovery.

\subsection{LIMP: Grounded Language Understanding}

Chapter~\ref{ch:limp} introduces Language Instruction Grounding for Motion Planning (LIMP), which connects expressive instructions to explicit task constraints and spatially grounded plans.
Language and vision-language models supply descriptions of the task and its referents, while spatial reasoning resolves object instances and connects the resulting specification to motion planning.
The central advance is to construct the semantic information required by an instruction at runtime, using observations retained during metric mapping, while preserving an explicit representation of its temporal and spatial requirements.
This makes learned semantic breadth available to structured planning without requiring a complete prebuilt semantic map.

\subsection{REALM: Evolving Representations of the World}

Chapter~\ref{ch:realm} develops REALM, an action-conditioned spatial world model that carries a persistent representation across successive tasks.
It derives the semantic and geometric abstractions needed for grounding, planning, and rendered inspection from a common 3D Gaussian scene~\citep{kerbl2023gaussians}.
Action-conditioned transitions predict geometric change and update the affected parts of the scene while preserving object identity and unaffected scene content.
These predicted changes provide the world state used to ground and plan subsequent tasks.
The scene becomes a world model through these predictive transitions and its reuse for later decisions.
The contribution is a persistent spatial basis for planning whose evolution remains explicit.

\subsection{PACTS: Composable State-Reporting Skills}

Chapter~\ref{ch:pacts} introduces Predicate-Action Skills (PACTS), which learn control and its symbolic interface through joint generation of action trajectories and temporally aligned predicate-belief trajectories.
Conditioned on current observations, each generated sample pairs proposed actions with the symbolic conditions the skill expects as those actions unfold.
These reports supply an off-the-shelf planner with state estimates for sequencing skills, checking effects, and replanning.
They are refreshed through observation-conditioned rollouts, but remain predictions rather than independent measurements of action success.

Together, these contributions address three requirements for extending robot capability while preserving a basis for assurance.
Broader language understanding must remain connected to explicit task requirements; learned skills must expose the state needed to assess and compose their behavior; and the world representation used for planning must account for change.
By making these connections explicit, this thesis develops a foundation for robots whose capabilities can expand while their interpretation and behavior remain open to examination.
\par
\endgroup


\chapter{Background}
\label{ch:background}
\begingroup
\clubpenalty=10000
\widowpenalty=10000
\makeatletter
\@clubpenalty=10000
\makeatother

Robots become more useful when they can carry what they have learned into tasks that were never individually programmed or demonstrated.
The introduction argued for pursuing this breadth through a combination of learning and planning.
Learned models can bring new instructions, objects, and skills within a planner's reach, while planning can organize those capabilities to accomplish new goals.
The challenge is to preserve the meaning of a task as it passes from a person's request into the robot's decisions and, ultimately, into physical action.

Making this connection requires choices about what the robot represents and what it can infer from those representations.
A planner can reason that a grasp must precede a placement only if the consequences of grasping are expressed in terms it can use.
Perception must establish which object those terms refer to, and execution must supply information about whether the conditions for continuing the task hold.
The representations connecting these processes determine both how their capabilities can be combined and what can be checked about the resulting behavior.

This chapter develops the ideas needed to understand and construct those connections.
It begins with the breadth of knowledge available through foundation models and examines how that knowledge can enter explicit reasoning about robot tasks.
Following this connection from interpretation to action leads to the physical grounding of task specifications, the maintenance of a changing world, and the acquisition of skills that a planner can reuse.
Together, these foundations explain how the opportunities for broader capability give rise to the interface problems addressed by the technical chapters.

\section{Foundation Models and Broader Generality}
\label{sec:background:foundation-models}

The knowledge required for a robot task often extends beyond the robot's own experience.
A person may describe an object through its appearance, its use, or its relationship to another object, using words that were never included in a task-specific training set.
Foundation models provide a way to bring knowledge learned from broad collections of data into such settings.
The term describes models trained at scale on broad data and adapted to a range of downstream uses~\citep{bommasani2021foundation}.
For robotics, the important change is that much of the representation learning can precede the definition of a particular task or deployment environment.

\subsection{Language and Visual Representations}

A language model can be asked to continue an ordinary sentence or to express an instruction in a formal language.
An autoregressive model approaches both through the same prediction problem: generating the next token, a unit of text, from the prompt and the tokens already produced.
With learned parameters $\theta$, the probability of a sequence $w_{1:n}$ conditioned on a prompt $c$ factorizes as
\begin{equation}
 p_\theta(w_{1:n}\mid c)
 =\prod_{i=1}^{n}p_\theta(w_i\mid c,w_{1:i-1}).
 \label{eq:bg:language-model}
\end{equation}
The prompt may include an instruction, descriptions of available tools, and examples of the desired output.
This makes it possible to request a structured expression, such as a program or a logical formula, using the same sequence-generation mechanism that produces ordinary text.
In-context examples specify how the model should respond without themselves changing its trained parameters~\citep{brown2020fewshot}.
They can guide the form and content of a translation, but the probability assigned to a token sequence is not a proof that the translation preserves the user's meaning.

Visual representations supply a complementary connection to the scene.
In an open-vocabulary detector, a text query can specify the object category or description to localize, rather than selecting only from a fixed list of detector labels.
OWL-ViT, for example, adapts image--text pretraining to text-conditioned object detection~\citep{minderer_simple_2022}.
Detection proposes image regions; segmentation identifies the pixels belonging to a candidate region.
These are distinct operations: a box around a book may include part of the table, whereas a segmentation mask can provide more selective support for projecting that book into three dimensions~\citep{kirillov_segment_2023}.

Open vocabulary expands the descriptions that can be submitted to perception; it does not imply exhaustive recognition of every concept or reliable identification of every instance.
Several books may match ``red book,'' and a correct category detection may still select the wrong book for a relational instruction.
Language understanding, object detection, and referent resolution therefore contribute different information.
The task becomes grounded only when the expressions in the instruction are associated with the relevant entities and conditions in the robot's environment~\citep{tellex_robots_2020}.

\subsection{From Pretrained Knowledge to Robot Behavior}

Pretrained representations can enter robot behavior at different levels.
Vision-language-action models connect language and visual knowledge to control learned from robot experience.
RT-2 studies this transfer by incorporating robot actions into a vision-language model's output representation~\citep{brohan2023rt}.
OpenVLA develops policies from diverse robot demonstrations and supports adaptation to additional robot setups~\citep{kim2024openvlaopensourcevisionlanguageactionmodel}, while $\pi_0$ combines pretrained vision-language representations with continuous action generation~\citep{black2024pi0}.
In each case, robot data connects semantic knowledge to the actions available to a particular embodiment.

The same knowledge can also support explicit intermediate reasoning.
SayCan combines language-model scores with learned skill value functions to select skills using estimates of their relevance to the instruction and their feasibility in the current state~\citep{ichter2023saycan}.
Code as Policies generates programs that call perception and control interfaces~\citep{liang_code_2023}.
Language-to-temporal-logic methods instead produce task specifications that can be passed to a planner~\citep{liu_grounding_2023}.
These approaches differ in what the learned model supplies: motor commands, skill choices, executable code, or formal requirements.
That choice determines which decisions remain available for separate examination.

A language-conditioned action policy can use task meaning without expressing it as a checkable specification.
Conversely, a generated specification can be checked for logical properties while still referring to the wrong object.
The value of an explicit interface is that it separates these questions and makes the relevant outputs accessible.
To understand what a planner can establish from such an interface, it is first necessary to describe the states, actions, and requirements over which planning operates.

\section{Symbolic State, Planning, and Task Specifications}
\label{sec:background:symbolic-planning}

To use knowledge about a task in planning, the robot needs a way to reason about what its actions would change.
For the book-transfer task, the relevant connection is that grasping the book makes it available to be carried, and carrying it to the sofa makes the requested placement possible.
Symbolic planning captures such relationships while leaving much of the detail of their physical realization to perception and control.
The resulting abstraction allows the planner to explore how familiar actions could accomplish a new goal.
Its construction begins with the conditions that distinguish one task-relevant state from another.

\subsection{Predicates and Action Models}

Let $\xi_t$ denote the physical state of the robot and its environment at time $t$.
It may include robot configuration, object poses, contact conditions, and quantities that are not directly observed.
A grounded predicate is a Boolean function of this state.
For example, $\mathrm{on}(b,T)$ describes whether book $b$ is on table $T$, and $\mathrm{holding}(b)$ describes whether the robot holds it.
The predicate name and its arguments define the condition of interest; an evaluator connects that condition to geometry, sensor data, or another state estimate.

Given a finite set $\mathcal{P}$ of grounded predicates, a symbolic state can be represented by the subset that is true:
\begin{equation}
 s_t=\{p\in\mathcal{P}:p(\xi_t)=1\}.
 \label{eq:bg:abstraction}
\end{equation}
This expression defines an abstraction of physical state, not a claim that the robot observes $\xi_t$ exactly.
In practice, the planner receives an estimate $\hat s_t$ constructed from available evidence.
Two physically different scenes can have the same symbolic description: the book may lie at several positions on the table while $\mathrm{on}(b,T)$ remains true.
Such abstraction is useful when the omitted differences do not invalidate the reasoning being performed~\citep{konidaris2018skillstosymbols}.

In a basic STRIPS-style representation, an action operator $A$ has a set of preconditions $\mathrm{Pre}(A)$, an add set $\mathrm{Add}(A)$, and a delete set $\mathrm{Del}(A)$~\citep{fikes_strips_1971}.
For the positive-precondition case, the operator is applicable when $\mathrm{Pre}(A)\subseteq s$ and predicts the successor
\begin{equation}
 \Gamma(s,A)=\bigl(s\setminus\mathrm{Del}(A)\bigr)\cup\mathrm{Add}(A).
 \label{eq:bg:strips}
\end{equation}
Conditions outside the add and delete sets retain their values.
An illustrative pick operator might require $\mathrm{on}(b,T)$, $\mathrm{near}(b)$, and $\mathrm{handempty}$; it adds $\mathrm{holding}(b)$ and deletes $\mathrm{on}(b,T)$ and $\mathrm{handempty}$.
This model describes the changes needed for task reasoning.
It leaves the choice of grasp and the detailed motion to the mechanisms that realize the operator.

A planning problem supplies an initial state $s_0$ and a goal condition $g$ in addition to the action model.
A sequential plan $(A_1,\ldots,A_n)$ is valid within this model if each operator is applicable in the state reached by its predecessors and the final state satisfies $g$.
Planning Domain Definition Language (PDDL) separates reusable domain descriptions from problem instances containing objects, initial conditions, and goals~\citep{ghallab1998pddl}.
This separation supports reuse: changing the requested arrangement need not change the operators that describe picking and placing.

The usual set-of-true-facts representation adopts a closed-world interpretation: a grounded fact absent from $s$ is false.
That convention is convenient for search, but it must be distinguished from incomplete perception.
Failure to detect a book is not evidence that no book exists.
Similarly, a fixed semantic map is not inherently closed-world; the assumption arises when its contents are treated as complete for the relevant reasoning.
The symbolic domain must be understood together with the procedure that supplies its facts.

\subsection{Requirements over Time}
\label{sec:background:temporal-logic}

A final-state goal specifies an outcome but does not describe every requirement on the way to it.
The robot might be required to visit one location before another, avoid a region while carrying an object, or repeat an activity indefinitely.
Temporal Logics, such as Linear Temporal Logic(LTL), express such requirements over sequences of propositional valuations~\citep{pnueli_temporal_1977,emerson_temporal_1990}.

Let $\mathrm{AP}$ be a set of atomic propositions and let $\sigma=\sigma_0\sigma_1\cdots$ be an infinite trace, with $\sigma_t\subseteq\mathrm{AP}$ containing the propositions true at step $t$.
An atomic proposition may be a grounded state predicate or an event associated with a skill, provided its meaning is specified.
LTL combines Boolean connectives with temporal operators.
The expression $\mathbf{X}\varphi$ requires $\varphi$ at the next step; $\mathbf{F}\varphi$ requires it eventually; and $\mathbf{G}\varphi$ requires it at every step.
The strong-until expression $\varphi\,\mathbf{U}\,\psi$ requires $\psi$ at some step and $\varphi$ at every preceding step.
For example,
\begin{equation}
 \begin{aligned}
 \sigma,t\models\varphi\,\mathbf{U}\,\psi
 \quad\Longleftrightarrow\quad
 &\exists j\geq t:\;\sigma,j\models\psi\ \text{and}\\
 &\forall i\in\{t,\ldots,j-1\},\ \sigma,i\models\varphi.
 \end{aligned}
 \label{eq:bg:until}
\end{equation}
Here $\models$ denotes satisfaction.
The operators describe logical ordering over trace positions; they do not by themselves specify durations in seconds.

For example, let $k$ mean that the robot is at the kitchen and $f$ that it is at the fridge.
The formula
\begin{equation}
 \varphi=\mathbf{F}\bigl(k\land\mathbf{F}f\bigr)
 \label{eq:bg:visit-order}
\end{equation}
requires a kitchen visit with a fridge visit at that time or later.
It permits earlier fridge visits and permits simultaneous satisfaction if both propositions can hold at once.
Replacing the inner $\mathbf{F}f$ with $\mathbf{X}\mathbf{F}f$ requires a later fridge visit, but still does not forbid an earlier one.
These differences illustrate why fluent language and a plausible subgoal list are insufficient to establish that a translation expresses the intended constraint.

\subsection{Automata, Synthesis, and the Scope of Verification}

Temporal specifications can be converted into automata that track progress as propositions become true~\citep{vardi_automata-theoretic_1996}.
For a co-safe specification, satisfaction is witnessed by a finite \emph{good prefix}: once that prefix has occurred, every continuation satisfies the specification~\citep{kupferman2001safety}.
Such requirements can be represented by a finite automaton accepting their good prefixes.
This differs from claiming that every LTL requirement is decidable from a completed finite task.
For example, observing a safe run so far cannot establish $\mathbf{G}\neg b$ for an unbounded future.
An avoidance requirement limited to reaching a goal can instead be written $(\neg b)\,\mathbf{U}\,(g\land\neg b)$, which requires eventual achievement of $g$ and excludes $b$ through that achievement.

\begin{figure}[htbp]
 \centering
 \includegraphics[width=\textwidth]{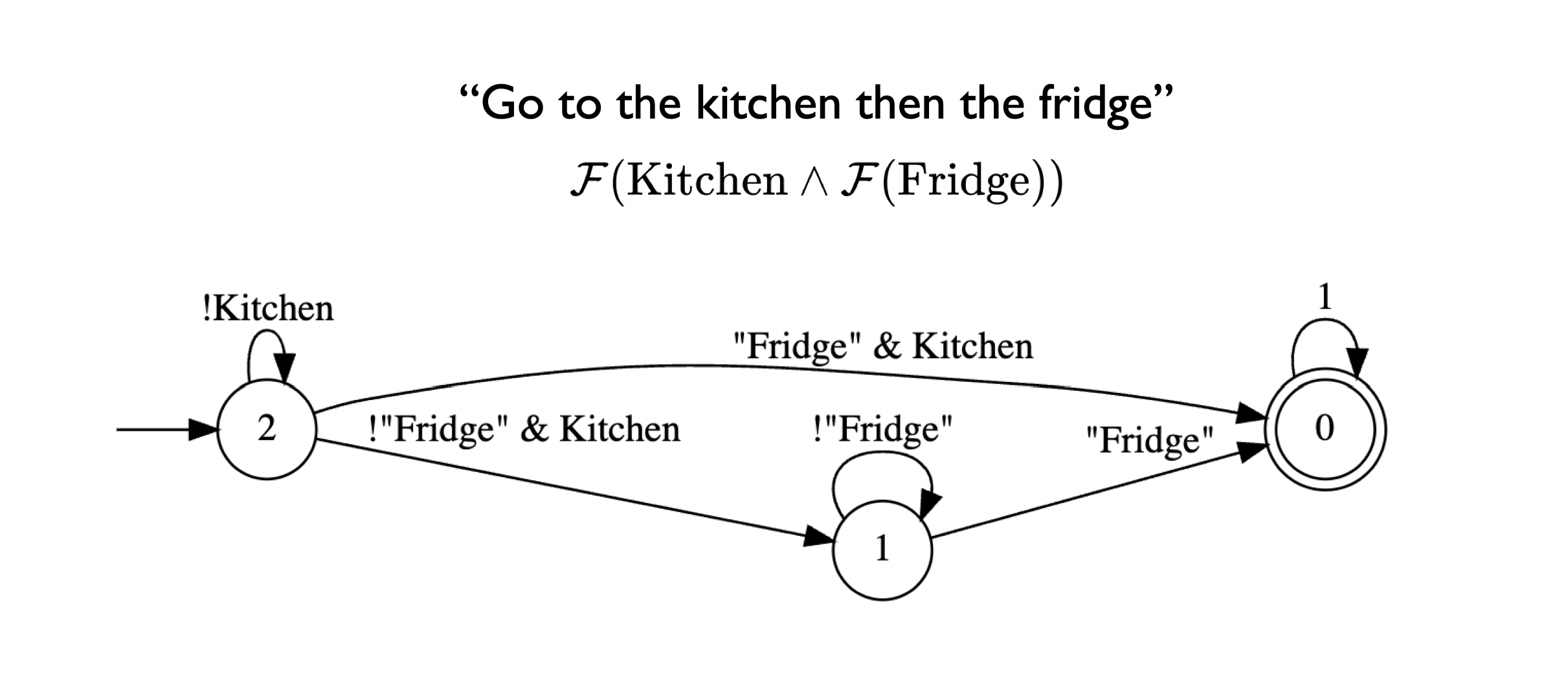}
 \caption[An automaton for an ordered-visitation requirement]{An automaton for $\mathbf{F}(k\land\mathbf{F}f)$. State 2 awaits a kitchen visit; state 1 records that the kitchen requirement has been met; state 0 accepts after a qualifying fridge visit. The direct transition from 2 to 0 permits simultaneous satisfaction. The self-loop at 2 also permits a fridge visit before the kitchen, so the formula does not prohibit such a visit. Edge labels describe proposition valuations, not physical actions.}
 \label{fig:bg:automaton}
\end{figure}

A deterministic automaton can be written as $\mathcal{A}_\varphi=(Q,q_0,2^{\mathrm{AP}},\delta,F)$, where $Q$ is its state set, $q_0$ its initial state, $\delta$ its transition function, and $F$ its accepting states.
After reading a valuation $\sigma_t$, its progress state changes according to
\begin{equation}
 q_{t+1}=\delta(q_t,\sigma_t).
 \label{eq:bg:automaton-update}
\end{equation}
Figure~\ref{fig:bg:automaton} shows how this state records which parts of a requirement have already been satisfied.
The automaton does not itself identify the fridge, compute a path, or cause the robot to move.
Those operations supply and act on the propositions that drive its transitions.

Verification and synthesis use the specification in different ways.
Model checking asks whether a supplied model satisfies a property; synthesis constructs behavior to satisfy the property under a model and stated assumptions~\citep{vardi_automata-theoretic_1996,kress-gazit_temporal-logic-based_2009}.
Combining a task model with an automaton allows a planner to reason about both the environment state and progress through the specification.
In robot applications, the resulting logical plan must also admit a physically feasible realization.
Thus a correct-by-construction plan is correct relative to its specification, grounding, and action model.
Whether that specification expresses the user's intent, and whether execution follows the modeled transitions, are additional questions.

\section{Task and Motion Planning with Reusable Skills}
\label{sec:background:tamp}

A symbolic plan can establish that picking precedes placing without determining how either action is carried out.
Physical execution introduces choices that affect the validity of the sequence: where the robot stands, which grasp it uses, how it approaches the object, and whether the object can pass through the available space.
Task and motion planning (TAMP) treats the discrete action sequence and these continuous choices as connected parts of one problem~\citep{garrett_integrated_2021}.

\subsection{Geometric Feasibility}

A robot configuration $q\in\mathcal{C}$ specifies the variables needed to determine its geometry, such as joint angles and mobile-base pose.
For a fixed environment and manipulation mode, the collision-free subset $\mathcal{C}_{\mathrm{free}}$ contains configurations that satisfy the chosen collision constraints.
A geometric path $\gamma:[0,1]\rightarrow\mathcal{C}_{\mathrm{free}}$ connects a start configuration to a goal configuration or goal set.
The path parameter specifies progression along the curve; a time-parameterized trajectory additionally specifies when configurations are reached.
Kinematic feasibility, collision avoidance, and dynamic feasibility impose different restrictions on such a trajectory.

Manipulation changes the geometry relevant to those restrictions.
An object attached to the gripper becomes part of the moving assembly, and a release changes its pose in the scene.
A task-level choice of grasp can therefore affect the feasibility of later transport and placement.
Methods such as PDDLStream connect symbolic search to procedures that generate and test continuous candidates, including grasps, poses, and motions~\citep{garrett_pddlstream_2020}.
The planner may need to revisit an earlier discrete or continuous choice when a later motion is infeasible.

Task requirements can also create geometric constraints beyond ordinary collision avoidance.
An instruction to avoid a particular object or region changes which paths are acceptable even when those paths are physically traversable.
Once a referent is localized, a proposition such as $\mathrm{near}(o)$ can label a region of space.
For a selected automaton transition, positions can then be distinguished according to whether they advance the task, violate its constraints, or leave its progress unchanged.
This is the role of the Task Progression Semantic Maps developed in Chapter~\ref{ch:limp}.
They connect logical progress to geometric search; they are not models of physical dynamics.

\subsection{Skills as Temporally Extended Actions}
\label{sec:background:options}

The planner need not select every motor command individually.
A skill can encapsulate a closed-loop behavior that runs over several control steps, such as approaching, grasping, or opening a door.
The options framework formalizes this temporal abstraction through
\begin{equation}
 O=(I_O,\pi_O,\beta_O),
 \label{eq:bg:option}
\end{equation}
where $I_O$ is the initiation set, $\pi_O$ the policy used during execution, and $\beta_O$ the termination rule~\citep{sutton_between_1999}.
In the stochastic formulation, $\beta_O(\xi)$ is the probability of terminating in state $\xi$.
The policy may itself use an observation or history rather than the full physical state.
A deterministic stopping condition is a special case of this formulation.

An option's duration can vary with its starting state and execution outcome.
At the level of skill selection, a transition model may therefore describe both the terminal state and elapsed duration, $p(\xi',d\mid\xi,O)$.
This is the source of the semi-Markov formulation associated with options: decisions occur after temporally extended behaviors, rather than necessarily after a single motor step.
The options framework can support reinforcement learning, but the abstraction does not require that every policy be learned through reward optimization.
An analytic controller, a motion planner, or a policy learned from demonstrations can each implement a skill.

An initiation set and a termination rule are not interchangeable with symbolic preconditions and successful effects.
A symbolic precondition summarizes when an operator is applicable; its physical realization requires the corresponding skill to be able to run from the actual state.
Likewise, a skill may stop because it succeeded, timed out, or encountered a failure.
Termination alone cannot establish its intended effects.
Composition requires a connection between the states one skill can produce and the states from which the next can operate~\citep{burridge1999seqcomp}.

This connection explains why state abstraction and skill design belong together.
The useful symbolic distinctions depend on the behaviors available to the robot and the questions the planner must answer about them~\citep{konidaris2018skillstosymbols}.
A representation that records only whether a book is on a table may suffice for one task model but omit the accessibility information needed by a particular grasping skill.
The planner's symbols must therefore be grounded at a level that supports the intended composition, rather than chosen solely because they are convenient labels.

\section{Semantic Mapping and Spatial Grounding}
\label{sec:background:grounding}

The ability to reuse a skill depends on recognizing where it can be applied.
A model of picking up a book can serve tasks in different rooms, but each use requires the robot to identify a particular book, locate it, and relate its geometry to the motion being considered.
The same abstraction that makes a planning model reusable must therefore be connected to the details of the present scene.
Semantic mapping and grounding establish this connection, allowing task reasoning to draw on what the robot perceives.

\subsection{From Images to a Metric Scene}

Observations from different viewpoints must first be brought into a common spatial frame so that detections and planned motions refer to the same environment.
Simultaneous localization and mapping (SLAM) addresses this problem by estimating the robot's state while constructing a representation of the environment from its observations~\citep{cadena2016slam}.
A metric map provides spatial relationships in that common coordinate system.
A semantic map adds associations with categories, instances, attributes, or relations.
These are complementary forms of information: a collision map may support navigation without identifying a book, while a text label alone does not provide the geometry needed to reach it.

Calibrated RGB-D observations connect visual detections to metric structure.
Let $K$ be a pinhole camera's intrinsic matrix, $(u,v)$ a pixel, and $D(u,v)$ its depth measured along the camera's optical axis.
For an undistorted image, backprojection gives the camera-frame point
\begin{equation}
 \mathbf{p}_C=D(u,v)K^{-1}[u,v,1]^\top.
 \label{eq:bg:backprojection}
\end{equation}
If $(R_{WC},\mathbf{t}_{WC})$ maps camera coordinates into the world frame, then
\begin{equation}
 \mathbf{p}_W=R_{WC}\mathbf{p}_C+\mathbf{t}_{WC}.
 \label{eq:bg:camera-world}
\end{equation}
The calibration and depth convention determine the meaning of these equations~\citep{hartley2004multiple}.
Applying them to pixels inside a segmentation mask produces candidate three-dimensional support for an object.
Combining observations from several viewpoints can improve coverage, but also requires deciding which detections refer to the same instance.

An object record can contain a persistent identifier, pose, spatial extent, appearance features, and supporting observations.
Scene graphs additionally represent relations among such entities.
Hydra and ConceptGraphs illustrate ways to organize metric and semantic information into structures that support higher-level reasoning~\citep{hughes2022hydra,gu2023conceptgraphs}.
For this thesis, the useful distinction is between storing the environment's geometry and deciding which parts of that geometry matter to an instruction.
Task-relevant semantics can be constructed when needed from retained observations or a suitable scene representation, rather than exhaustively enumerated at mapping time.

\subsection{Resolving Objects, Relations, and Identity}

Referent resolution must distinguish an object's description from its identity.
``The red book'' is a description that may match several objects; ``the red book on the table'' adds a relation that can restrict the candidates.
An instruction can nest these relations, as in a book on a table beside a sofa.
Resolving it requires evaluating candidate combinations, not simply detecting each noun independently.
The result should bind a task expression to an object instance that can be used consistently by the planner.

Spatial predicates also depend on their reference frame and geometric interpretation.
``Left of'' can refer to the observer's viewpoint or an object's intrinsic orientation, whereas ``above'' may be defined relative to gravity.
Different languages and contexts make different use of spatial frames~\citep{majid_can_2004}.
A numerical comparison between object centers is meaningful only after the relevant frame has been fixed.
Other relations, such as support or containment, may require shape, contact, or volume information rather than a comparison of centers alone.

The same distinction matters over time.
If the robot moves a book, its description may change while its identity persists.
Repeated detection provides new candidates but does not automatically establish their correspondence to earlier objects.
Similarly, a semantic identification does not determine which geometric elements should move with the object.
An instance record and its geometric support answer different questions: which object the task concerns, and which part of the scene represents that object.
Maintaining both becomes necessary when the representation must account for the robot's interventions.

\section{World Models and State Change}
\label{sec:background:world-models}

Grounding makes a representation useful for action, but action can change the very conditions on which that grounding depends.
Once the robot has moved the book, later instructions and plans must refer to the scene that results.
Continued autonomy therefore requires a way to carry knowledge of the environment forward, preserving what remains useful while accounting for what has changed.

A world model supports this reasoning by describing how aspects of the scene evolve.
For manipulation, this thesis adopts a functional view: the model should be grounded in the environment, predict task-relevant changes, and account for the actions that cause them.
The appropriate mechanism depends on what the robot needs to predict.
An analytic transformation, a learned transition distribution, or a combination of the two can supply the required information without modeling every physical process.

Let $W_t$ be an internal representation at a task-relevant decision time, and let $A_t$ describe an intervention.
A deterministic transition model produces
\begin{equation}
 \widehat W_{t+1}=F(W_t,A_t),
 \label{eq:bg:world-prediction}
\end{equation}
while a probabilistic model may represent $p(W_{t+1}\mid W_t,A_t)$.
Both differ from a renderer, which predicts an image from a scene and camera pose without necessarily representing any change to the scene.
Predicting a new view of a fixed object is not the same operation as predicting the consequences of moving it.

The appropriate prediction depends on the decision being supported.
A task planner may need only whether a door becomes open; a motion planner needs the resulting geometry; a controller may require velocities and contact forces.
Dynamic Scene Representation, for example, studies object permanence and continuity in action-conditioned manipulation prediction~\citep{xu2020dynamic}.
Such work makes explicit that a useful state must retain information about objects even when they are temporarily hidden.
The level of detail can vary, but it must support the downstream question.

For a rigid object, a predicted change of pose can preserve its shape while relocating it in the scene.
If the prediction comes from the endpoint of a manipulation plan, its meaning is conditional: this is where the object should be if the planned action succeeds.
A failed grasp or slip can produce a different physical outcome.
Observation-based correction therefore plays a different role from prediction.
In a general state-estimation loop, new evidence $o_{t+1}$ can revise the predicted representation through an update $W_{t+1}=U(\widehat W_{t+1},o_{t+1})$.
The prediction provides an expectation; the observation constrains whether that expectation describes the environment.

Not every image used to update a scene is an independent observation of the world.
A rendered image depicts the representation, and a generated image supplies an inferred appearance consistent with its conditioning.
Such images can support prediction or reconstruction, but agreement with them measures a different property from agreement with a physical sensor observation.
Preserving the source of the evidence is necessary to interpret what an updated model establishes.

Continued planning also requires agreement among the representations derived from the world state.
If an object moves, its semantic location, collision geometry, and rendered appearance should describe the same revision of the scene.
A correct update to one representation cannot compensate for another component continuing to use an old one.
Chapter~\ref{ch:realm} develops this problem using an explicit, renderable spatial representation.
The properties of that representation explain both how local updates are possible and why visual reconstruction alone does not complete a world model.

\section{Gaussian Scene Representations}
\label{sec:background:gaussians}

For a world model to remain useful after manipulation, its scene representation must support the changes the robot needs to predict.
When an object moves, the robot needs a way to change its represented position while retaining the surrounding scene.
When a person inspects a proposed action, the representation should make the relevant geometry and spatial relationships visible.
Local editing and rendered inspection place different demands on the information stored about the world.

A point cloud records samples of three-dimensional geometry; a mesh connects surface elements; and an occupancy representation describes whether regions are occupied.
A radiance representation also models appearance as a function of viewpoint, allowing images to be synthesized from camera poses other than those used during capture.
Combining this ability to render new views with an explicit spatial structure offers a basis for inspecting a scene and updating the parts affected by manipulation.

\subsection{Gaussian Primitives and Image Formation}

Three-dimensional Gaussian Splatting represents scene appearance using explicit primitives with position, shape, opacity, and color parameters~\citep{kerbl2023gaussians}.
A primitive centered at $\boldsymbol\mu_i\in\mathbb{R}^3$ has spatial profile
\begin{equation}
 G_i(\mathbf{p})=
 \exp\!\left[-\tfrac{1}{2}(\mathbf{p}-\boldsymbol\mu_i)^\top
 \Sigma_i^{-1}(\mathbf{p}-\boldsymbol\mu_i)\right],
 \label{eq:bg:gaussian}
\end{equation}
where the positive-definite covariance $\Sigma_i$ controls its extent and orientation.
These roles can be separated through
\begin{equation}
 \Sigma_i=R_i\operatorname{diag}(s_i^2)R_i^\top,
 \label{eq:bg:covariance}
\end{equation}
where the vector $s_i$ contains three positive scales, squared componentwise, and $R_i$ is a rotation matrix.

To render an image, the primitives are projected into the camera and their contributions are combined in visibility order.
For a pixel $\mathbf{u}$, a simplified front-to-back compositing expression is
\begin{equation}
 C(\mathbf{u})=
 \sum_i T_i(\mathbf{u})\,\alpha_i(\mathbf{u})\,c_i(\mathbf{d}),
 \qquad
 T_i(\mathbf{u})=\prod_{j<i}\bigl(1-\alpha_j(\mathbf{u})\bigr),
 \label{eq:bg:compositing}
\end{equation}
where $\alpha_i(\mathbf{u})$ combines opacity with the projected footprint, $T_i$ is the remaining transmittance, and $c_i(\mathbf{d})$ is color for viewing direction $\mathbf{d}$.
Spherical-harmonic coefficients provide one way to represent this directional appearance.
The parameters can be optimized by comparing rendered and captured images, with the original method also adjusting the density of primitives during reconstruction~\citep{kerbl2023gaussians}.

The resulting representation is explicit in space, but its elements need not align with object boundaries.
A primitive may contribute to several pixels, and a collection that looks like one object from a particular view may include support belonging to nearby surfaces.
Good image reconstruction and correct object ownership are consequently different requirements.
This distinction matters for manipulation: moving every primitive in a broad image crop can inadvertently move background content along with the target object.

\subsection{Object Motion, Local Updates, and Inspection}

Once the primitives belonging to an object have been identified, their explicit geometry makes it possible to move them together.
For a rigid change of coordinates $(R,\mathbf{t})$, a Gaussian's geometry transforms as
\begin{equation}
 \boldsymbol\mu_i'=R\boldsymbol\mu_i+\mathbf{t},
 \qquad
 \Sigma_i'=R\Sigma_iR^\top.
 \label{eq:bg:gaussian-transform}
\end{equation}
Applying the same transformation to an object's correctly identified primitives moves its represented geometry while preserving its internal arrangement.
Other scene elements can remain fixed.
When appearance is directional, its representation must also be handled consistently with the chosen rotation and lighting assumptions.
The mathematical convenience of a rigid transformation does not establish that the physical object underwent that transformation; the distinction between a planned and an observed pose still applies.

Object motion also exposes parts of the scene that may not have been visible before.
Relocating the book reveals the surface beneath it, and a new viewpoint may reveal an unobserved side of the book itself.
Geometry alone cannot recover all of this appearance.
Local reconstruction methods such as CL-Splats update changed regions while retaining the rest of an existing Gaussian scene~\citep{ackermann2025clsplats}.
This provides a basis for maintaining scene continuity without fitting the entire environment after each change.
Which regions may change, which quantities remain constrained by the action model, and which observations support the update are additional design choices.

Rendered inspection can expose consequences of these choices.
A user can view a proposed path relative to objects and obstacles, or examine whether an updated scene places an object where expected.
The image makes a representation accessible, but its realism does not certify collision freedom, object identity, or physical correctness.
A Gaussian scene therefore becomes useful as a world model through its transition and update mechanisms and through the geometric and semantic abstractions it supplies to later decisions.

\section{Learning Visuomotor Skills from Demonstrations}
\label{sec:background:skill-learning}

A current world representation allows the planner to reconsider what can be done as the scene changes.
The range of behaviors it can construct, however, still depends on the skills available to it.
Extending that range requires a way to acquire control policies whose detailed movements would be difficult to engineer for every object and situation.
Learning from demonstration approaches this problem through examples of successful behavior.

For manipulation, those examples can contain images, robot state, and the commands issued during execution.
The policy learns how observations relate to actions without requiring the designer to specify every movement.
The quality and coverage of the demonstration data, the observation representation, and the policy architecture all affect the resulting behavior~\citep{robomimic2021}.

\subsection{Behavior Cloning and Closed-Loop Execution}

Let $\mathcal{D}$ contain demonstrated observation--action pairs $(o_t,a_t)$.
Behavior cloning trains a conditional policy to reproduce those actions; a probabilistic formulation minimizes
\begin{equation}
 \mathcal{L}_{\mathrm{BC}}(\theta)
 =-\mathbb{E}_{(o_t,a_t)\sim\mathcal{D}}
       \bigl[\log\pi_\theta(a_t\mid o_t)\bigr].
 \label{eq:bg:behavior-cloning}
\end{equation}
The conditioning input may include a recent observation history and a task description.
At deployment, a closed-loop policy receives new observations as the robot acts.
It therefore responds to the scene rather than replaying a stored sequence of commands.
Its responsiveness nevertheless depends on whether its inputs distinguish the states that require different actions.

Supervised action prediction is evaluated on a distribution that execution can itself change.
An early error may bring the robot to a state that the demonstrator rarely visited, where another error is more likely.
This sequential distribution shift motivates approaches that collect corrective behavior on states visited by the learned policy~\citep{ross2011reduction}.
Feedback helps the robot respond to deviations, but does not ensure that every unfamiliar situation is covered by its demonstrations.
Task-level recovery may also require replacing the current skill, rather than adjusting its next movement.

Several different action sequences may be appropriate for the same observation.
For example, an object can be approached from either side, yet averaging the two approaches may produce an undesirable path between them.
Generative policies address this multimodality by learning a distribution over action trajectories.
Let
\begin{equation}
 \mathbf{x}=(a_t,\ldots,a_{t+H-1})\in\mathbb{R}^{H\times d_a}
 \label{eq:bg:action-chunk}
\end{equation}
be a length-$H$ action chunk, with $d_a$ control dimensions per step.
A conditional generator samples $\mathbf{x}\sim p_\theta(\mathbf{x}\mid o_t)$.
The robot can execute a short prefix, acquire another observation, and generate a new chunk.
Diffusion Policy uses this receding-horizon organization to combine coordinated action generation with visual feedback~\citep{chi2024diffusionpolicy}.

\subsection{Denoising Diffusion}
\label{sec:background:diffusion}

Sampling a whole action trajectory requires a model that can preserve the relationships among its movements while accommodating several possible ways to act.
Denoising diffusion learns such a distribution by learning to remove noise from corrupted training examples~\citep{ho2020denoising}.
For a clean action trajectory $\mathbf{x}^0$, the forward corruption process can be sampled directly at noise level $k$ as
\begin{equation}
 \mathbf{x}^k=\sqrt{\bar\alpha_k}\,\mathbf{x}^0
       +\sqrt{1-\bar\alpha_k}\,\epsilon,
 \qquad \epsilon\sim\mathcal{N}(0,I),
 \label{eq:bg:diffusion-noise}
\end{equation}
where $\bar\alpha_k=\prod_{j=1}^{k}\alpha_j$ and the schedule $\alpha_j\in(0,1)$ determines how much signal is retained.
A conditional noise-prediction network is trained with the objective
\begin{equation}
 \mathcal{L}_{\mathrm{DDPM}}
 =\mathbb{E}\!\left[
 \left\|\epsilon_\theta(o_t,\mathbf{x}^k,k)-\epsilon\right\|_2^2
 \right].
 \label{eq:bg:diffusion-loss}
\end{equation}
The expectation samples demonstrated trajectories, noise levels, and Gaussian noise.
At inference, a noisy sample is iteratively refined into an action trajectory.
Conditioning on the current observation guides the generated behavior without requiring the model to generate future images~\citep{chi2024diffusionpolicy}.

The diffusion index $k$ and the physical action index $t$ have different meanings.
One trajectory contains actions for several future control steps; denoising repeatedly refines that entire trajectory before its selected prefix is executed.
Neither a denoising step nor the reduction of noise in a sample is an event in the physical task.
This distinction is essential when interpreting diagrams of generated actions and predicate trajectories in Chapter~\ref{ch:pacts}.

\subsection{Conditional Flow Matching}
\label{sec:background:flow-matching}

The trajectory-generation problem can also be approached by learning how to move samples continuously from noise toward demonstrated behavior.
Flow matching learns a vector field that transports samples between these distributions~\citep{lipman2022flow,tong2023improving}.
To match the convention used in Chapter~\ref{ch:pacts}, let $\mathbf{x}^0$ be a data trajectory, $\mathbf{x}^1\sim\mathcal{N}(0,I)$ a noise sample, and define a linear interpolation
\begin{equation}
 \mathbf{x}_\tau=(1-\tau)\mathbf{x}^0+\tau\mathbf{x}^1,
 \qquad \tau\in[0,1].
 \label{eq:bg:flow-path}
\end{equation}
For this sampled pair, the target velocity is $\mathbf{x}^1-\mathbf{x}^0$.
A conditional vector field $v_\theta$ can be trained by
\begin{equation}
 \mathcal{L}_{\mathrm{CFM}}
 =\mathbb{E}\!\left[
 \left\|v_\theta(o_t,\mathbf{x}_\tau,\tau)
            -(\mathbf{x}^1-\mathbf{x}^0)\right\|_2^2
 \right].
 \label{eq:bg:flow-loss}
\end{equation}
The expectation samples trajectory--noise pairs and interpolation values $\tau$ uniformly from $[0,1]$.
Generation integrates the learned field from noise at $\tau=1$ toward data at $\tau=0$.
The interpolation parameter is again a sampling coordinate, not physical time.
Independent data--noise pairing defines one conditional flow-matching construction; optimal-transport variants change how the pairs are selected.
The straight-line form alone does not establish an optimal transport coupling between the complete distributions.

Diffusion and flow matching are alternative ways to train and sample a conditional trajectory model.
They do not specify what information the trajectory must contain.
A temporal convolutional network can model local structure along the horizon, while a Transformer can model interactions among time-step representations through attention.
Either architecture can condition on an observation encoding and a noise or flow parameter.
The modeling target, training objective, and network architecture are therefore separate choices.
For compositional autonomy, the central question is whether the generated behavior also supplies the state information needed to reason about its use.

\section{Symbolic Interfaces for Composition and Monitoring}
\label{sec:background:composition}

The value of an acquired skill extends beyond the movements it has learned to produce.
To use a grasping skill as part of a longer activity, the planner needs to reason about whether the object will be held and whether subsequent transport or placement is appropriate.
These are the conditions represented by the action models in Section~\ref{sec:background:symbolic-planning}.
Connecting the learned policy to those conditions gives the planner a basis for composing the skill with others and assessing progress through the task.
The remaining question is how the skill's state reports relate to the behavior it produces.

\subsection{State Estimates and Predicted Outcomes}

A predicate evaluator can estimate current symbolic conditions from an observation, for example through scores $b_j(o_t)\in[0,1]$ for predicates $j=1,\ldots,J$.
A threshold converts those scores into a discrete valuation for a classical planner.
This conversion is an interface decision: it does not make an uncertain estimate certain.
Nor is a score between zero and one automatically a calibrated probability.
Calibration concerns whether stated confidence agrees with empirical correctness frequencies, a property that must be assessed separately from classification accuracy~\citep{guo2017calibration}.

A predicted predicate trajectory answers a different question from a current-state evaluator.
For an action chunk $\mathbf{x}$, let
\begin{equation}
 \mathbf{z}=(\mathbf{z}_{t+1},\ldots,\mathbf{z}_{t+H})
       \in[0,1]^{H\times J}
 \label{eq:bg:predicate-trajectory}
\end{equation}
describe the conditions expected at the corresponding future steps.
An observation-conditioned classifier and a trajectory predictor may use similar inputs, but their targets refer to different times.
The vector $\mathbf{z}_{t+1}$ sampled before an action is taken is a prediction of its next state; it is not an independent observation confirming success after execution.
A later rollout can incorporate a new observation without changing the provenance of the earlier prediction.

There is also a distinction between predicting action and outcome marginals and predicting their relationship.
Independent samples from $p(\mathbf{x}\mid o_t)$ and $p(\mathbf{z}\mid o_t)$ may select incompatible alternatives when different action modes lead to different outcomes.
A joint model $p(\mathbf{x},\mathbf{z}\mid o_t)$ can represent this dependence.
Sharing an observation encoder does not by itself require two independently sampled outputs to agree.
Chapter~\ref{ch:pacts} studies joint generation as a way to acquire this symbolic interface together with the control policy.

The distinction also determines what must be evaluated.
Predicate accuracy compares a prediction with labeled conditions; action--predicate coherence compares predicted conditions with those produced by executing the associated actions.
Accurate marginals do not establish that correspondence, and joint generation is a modeling choice rather than a proof of physical agreement.
Evaluating paired rollouts test the relationship on which the planner relies.

\subsection{Composition, Progress, and Recovery}

With state estimates and action models available, a planner can select a skill, monitor its intended effects, and plan again from updated conditions.
This differs from replaying a fixed order of skills: the next choice depends on the state presented to the planner.
If a door closes, the original sequence may no longer be applicable even when the current controller continues to produce valid motor commands.
Recovery requires recognizing the changed condition and selecting behavior that can restore progress.

The policy and planner can operate at different timescales.
The policy uses frequent observations to adjust movement; the planner reasons over task-relevant conditions to decide which skill should run.
This separation permits reuse across goals, provided the skill's initiation conditions, effects, and state reports remain meaningful in the new composition.
It does not remove the need for a suitable symbolic vocabulary or an action model.
Learning a controller, learning a predicate evaluator, discovering useful predicates, and learning operator preconditions and effects are distinct problems.

Similarly, zero-shot composition concerns what is new at the level of the requested behavior.
The constituent skills can be trained from demonstrations while their requested combination has no end-to-end demonstration.
This is different from performing a skill that the robot has never learned.
The distinction connects the two sources of generality introduced earlier: learning broadens the available behavior, while planning can arrange that behavior into additional tasks.

Learned knowledge and behavior become useful to planning when the robot can relate them to explicit requirements, conditions, and changes in the world.
The same connections make it possible to examine whether the planner's reasoning applies to the task and environment at hand.
Constructing these connections is therefore central to advancing capability and assurance together.

The first such connection is needed before a plan can be constructed.
A person's instruction must become a task that the planner can reason about, with its referents identified and its temporal and spatial requirements made explicit.
Chapter~\ref{ch:limp} develops that connection without assuming a complete semantic map in advance.
Its language-to-plan framework then provides the basis for examining how the world representation and the skill library can themselves be extended in Chapters~\ref{ch:realm} and~\ref{ch:pacts}.
\par
\endgroup


\chapter{Grounding Language in the Real World}
\label{ch:limp}

\begin{quote}
\small
 This chapter is adapted from \textit{Verifiably Following Complex Robot Instructions with Foundation Models}, coauthored by Benedict Quartey, Eric Rosen, Stefanie Tellex, and George Konidaris~\citep{quartey2025verifiably}.
The authorial ``we'' refers to this collaborative research.
\end{quote}
\vspace{0.75\baselineskip}

\noindent
The usefulness of a general-purpose robot depends in part on the range of tasks a person can ask it to perform.
An instruction may identify an object through its relationship to another, require several actions in a particular order, or prohibit behavior that would otherwise be feasible.
The preceding background chapter established how symbolic planning can express and reason about such requirements.
The difficulty is to obtain the right requirements from the person's words and connect them to the environment in which the robot must act.
A plan can satisfy a specification while pursuing the wrong object or the wrong interpretation of the task.

This chapter develops the first connection in the thesis argument: \emph{grounded language understanding} at the interface of intent.
LIMP brings the breadth of foundation models into a language-to-plan framework in which task requirements and their spatial grounding can be inspected.
The instruction determines which semantic information the robot constructs from its observations, allowing the planner to work beyond a prespecified set of named landmarks.
This establishes a basis for expressive instruction following while retaining two assumptions: the mapped environment remains static, and suitable navigation and manipulation skills are already available.
Chapters~\ref{ch:realm} and~\ref{ch:pacts} take up these assumptions in turn.

\section{From an Instruction to a Grounded Plan}
\label{sec:limp:introduction}

\begin{figure}
    \centering
    \includegraphics[width=1\columnwidth,trim={0cm 0cm 0cm 0cm},clip]{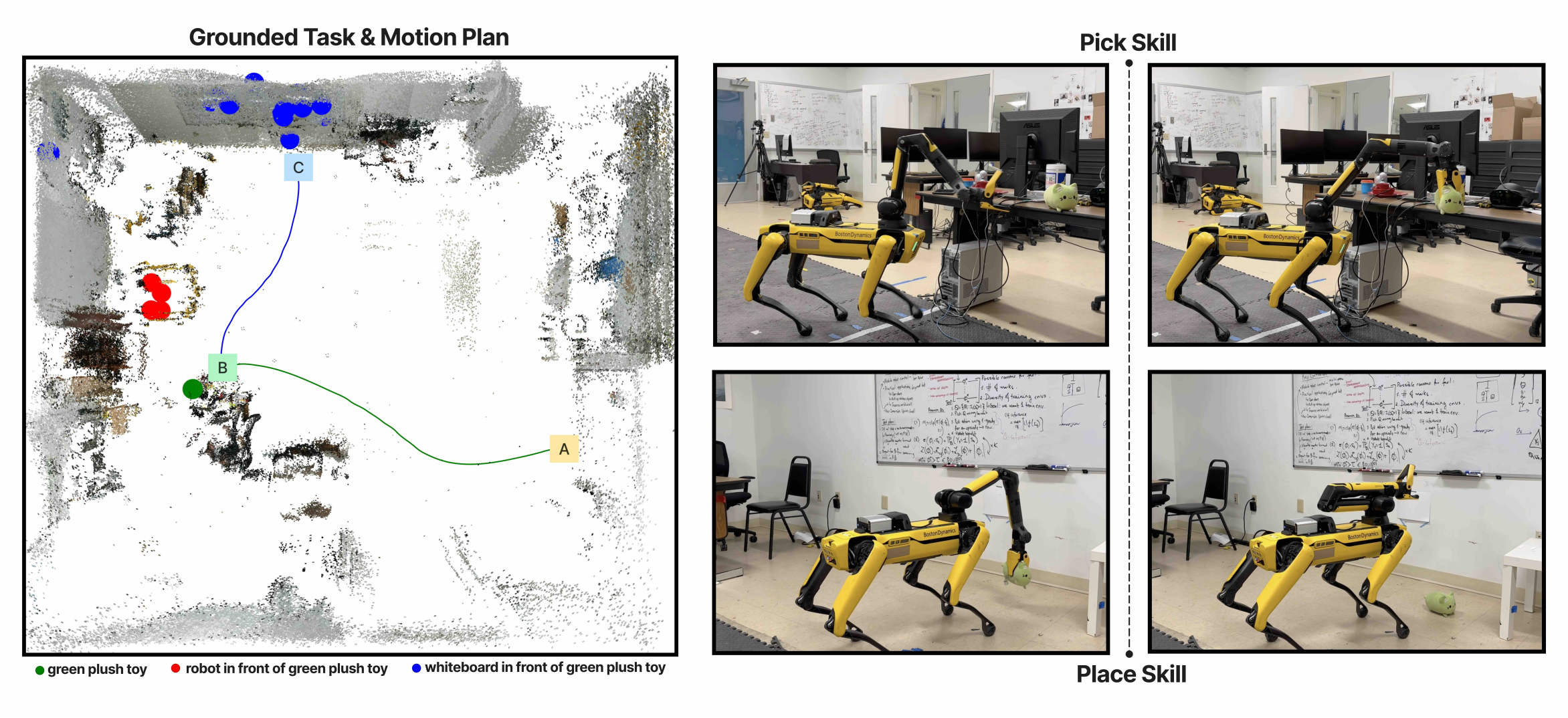}
    \vspace{-0.5cm}
    \caption[Grounded instruction following with LIMP]{LIMP executing the instruction \textit{``Bring the green plush toy to the whiteboard in front of it, watch out for the robot in front of the toy''}. The robot dynamically detects and grounds open-vocabulary referents with spatial constraints to construct an instruction-specific semantic map, then synthesizes a task and motion plan to solve the task. In this example, the robot navigates from its start location (yellow, A), to the green plush toy (green, B), executes a pick skill then navigates to the whiteboard (blue, C), and executes a place skill. Note that the robot has no prior semantic knowledge of the environment.}
\label{fig:limp:splash}
\end{figure}

The instruction in Figure~\ref{fig:limp:splash} illustrates why language understanding and spatial grounding must be developed together: \textit{``Bring the green plush toy to the whiteboard in front of it, watch out for the robot in front of the toy.''}
Recognizing a plush toy or a whiteboard is insufficient.
The robot must distinguish the intended instances through their spatial relationships, determine the order of navigation and manipulation, and interpret the other robot as a constraint on the route.
The meaning of the request therefore depends on both the structure of the instruction and the arrangement of the scene.

Language-guided navigation systems connect descriptions to object goals and landmarks~\citep{gadre_cows_2023,shah_lm-nav_2023}, while mobile-manipulation systems extend this connection to picking and placing~\citep{yenamandra_homerobot_2023,chen_open-vocabulary_2023,liu_demonstrating_2024,jaafar2025lambda}.
LIMP addresses the additional problem of resolving descriptive spatial references together with temporal constraints.
Directly generating a sequence of subgoals or a control program does not, by itself, establish that these constraints are respected~\citep{liang_code_2023}.
Translating the request into temporal logic makes the requirements available to a planner, but the symbols must still refer to the right things.
Earlier language-to-logic systems ground those symbols in prebuilt semantic maps or landmark descriptions~\citep{liu_grounding_2023,berg_grounding_2020}.
LIMP instead constructs the required semantics from the instruction and the robot's retained observations.

\textit{Language Instruction grounding for Motion Planning (LIMP)} combines language translation, spatial grounding, and task and motion planning.
An LLM translates the instruction into a temporal-logic specification with composable descriptions of its referents.
Vision-language models and spatial reasoning locate those referents in a metric map, and the planner uses the resulting specification and semantic information to construct navigation and manipulation behavior.
The three contributions are a modular language-to-plan framework, a spatial grounding method for open-vocabulary objects with relational descriptions, and a planning procedure that turns task-progress requirements into geometric goal and avoidance regions.
Together, they make the interpretation and planned realization of an instruction available for inspection before execution.


\section{Related Work}
\label{sec:related-works}
NLMap~\citep{chen_open-vocabulary_2023} uses pretrained vision-language models to make a scene representation queryable through open-vocabulary descriptions. Object availability and location then ground an LLM-based planner. LIMP connects such grounding to an explicit representation of spatial relations and temporal constraints, allowing these requirements to guide task and motion planning.

Lang2LTL~\citep{liu_grounding_2023} uses entity extraction and replacement to translate language instructions into temporal logic. Lang2LTL-2~\citep{liu_lang2ltl-2_2024} extends this approach with visual and textual landmark descriptions and spatial-relation reasoning. Both ground symbols using a supplied semantic map or landmark database. LIMP instead constructs instruction-specific semantic information from the robot's retained observations.

Action-Oriented Semantic Maps (AOSMs)~\citep{rosen_synthesizing_2023} construct navigation abstractions that connect portable manipulation skills in new environments. Their representation is organized around the supplied manipulation skills and their symbolic abstractions. LIMP develops a complementary connection: an instruction determines which open-vocabulary referents and spatial relationships must be grounded for planning.
\subsection{Temporal Specifications for Instruction Following}
\label{sec:background}
The temporal-logic semantics and automata construction in Section~\ref{sec:background:temporal-logic} provide the basis for LIMP's task representation.
The choice of LTL allows ordering, eventual achievement, and avoidance requirements to be expressed together~\citep{pnueli_temporal_1977,menghi_specification_2021}.
Recent approaches translate natural language into such specifications~\citep{liu_grounding_2023,berg_grounding_2020,pan_data-efficient_2023,cosler_nl2spec_2023,fuggitti_nl2ltl_2023,chen_nl2tl_2023}; LIMP couples this translation to the resolution of open-vocabulary spatial descriptions.

Temporal task structure can also support learning.
\citet{quartey_exploiting_2023} combine temporal-logic task representations with language-model-derived object embeddings to generate contextually related auxiliary tasks that can be learned from shared experience.

The planner uses the specification to synthesize behavior rather than checking an independently generated plan afterward~\citep{vardi_automata-theoretic_1996,kress-gazit_temporal-logic-based_2009,colledanchise_synthesis_2017}.
The resulting correctness claim is relative to the translated specification, its grounding, and the modeled actions.
Inspection of the translation and the referent map addresses whether these inputs express the user's intended task; successful physical execution is a further requirement.

\section{Problem Definition}
\label{sec:problem_definition}
Given a natural language instruction $l$, our goal is to synthesize and sequence navigation and manipulation behaviors to produce a policy that satisfies the temporal and spatial constraints in $l$. Spatial constraints determine task success based on the sequence of robot poses traversed during execution; temporal constraints determine the sequencing of these spatial constraints as a function of task progression. We assume a robot with an RGB-D camera has already navigated a space, capturing images and camera poses. From these observations, the robot builds a metric map $m$ (e.g., point cloud, 3D voxel grid) of the environment, defining the space of possible $SE(3)$ poses $P$ and enabling robot localization (i.e., estimating $p_{\text{robot}} \in P$). Unlike previous work leveraging temporal logic~\citep{liu_grounding_2023}, we do not assume access to a semantic map with prespecified object locations or predicates. LIMP uses the foundation-model interfaces introduced in Section~\ref{sec:background:foundation-models}: a task-agnostic vision-language model $\sigma$ for text-conditioned detection or segmentation, and an LLM $\psi$ for instruction translation.
\vspace{1mm}

\noindent\textbf{Navigation}: Navigation is formalized as an object-goal oriented continuous path planning problem, where the goal is to generate paths to a goal pose set $P_{goals} \subset P$ while staying in feasible regions ($P_{feasible} \subset P$) and avoiding infeasible regions ($P_{infeasible} =  P_{feasible}^{C}$). Infeasible regions include environment obstacles as well as dynamically determined semantic regions that violate constraints in the instruction $l$. \looseness=-1
\vspace{1mm}

\noindent\textbf{Manipulation}: Manipulation behaviors use the option formulation from Section~\ref{sec:background:options}, with an object parameter $\theta$. For $O_{\theta}=(I_{\theta},\pi_{\theta},\beta_{\theta})$, initiation and termination depend on the robot pose and object attributes, and the policy is conditioned on the target object. The initiation set $I_{\theta}$ denotes the global reference frame robot positions and object-centric attributes––such as object size––that determine if the option policy $\pi_{\theta}$ can be executed on the object $\theta$. To execute a manipulation skill on an object, an object-goal navigation behavior must first be executed to bring the robot into proximity with the object. We assume access to a library of these manipulation skills and demonstrate our approach on multi-object goal navigation and open-vocabulary mobile pick-and-place~\citep{yenamandra_homerobot_2023,yokoyama_asc_2024}. \looseness=-1


\section{Language Instruction Grounding for Motion Planning}
\begin{figure*}
\centering
\includegraphics[width=1\textwidth]{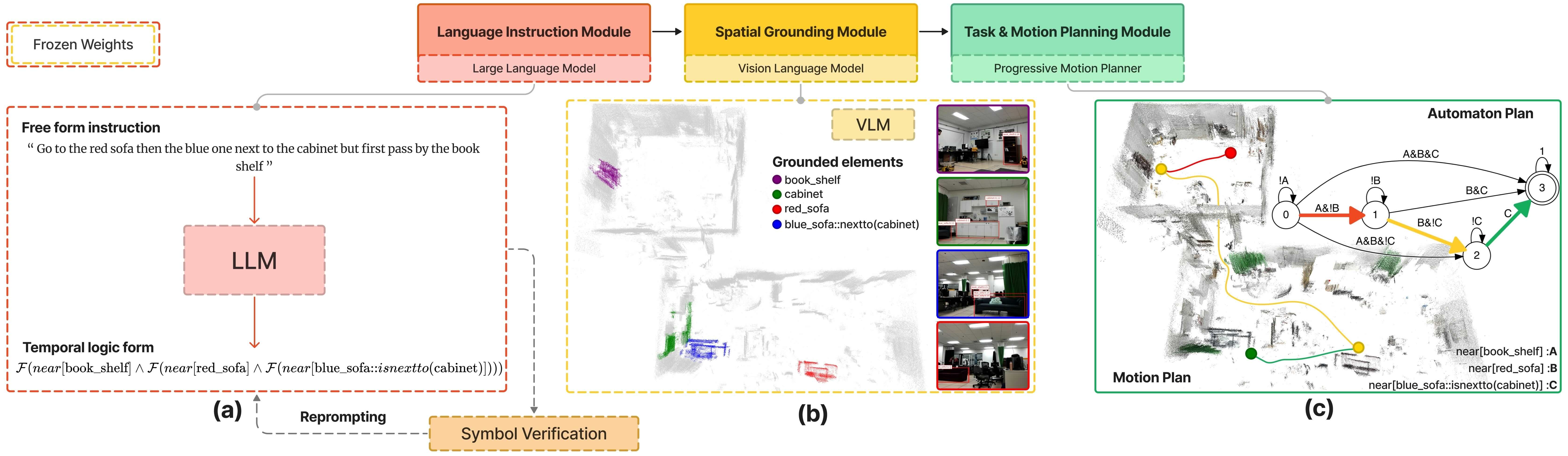}
\captionof{figure}[The LIMP language-to-plan framework]{\textbf{[A]} LIMP translates natural language instructions into temporal logic expressions, where open-vocabulary referents are applied to predicates that correspond to robot skills––note the context-aware resolution of the phrase ``blue one'' to the referent ``blue\_sofa''. \textbf{[B]} Vision-language models detect referents, while spatial reasoning disambiguates referent instances to generate a 3D semantic map that localizes instruction-specific referents. \textbf{[C]} Finally, the temporal logic expression is compiled into a finite-state automaton, which a task and motion planner uses with dynamically-generated task progression semantic maps to progressively identify goals and constraints in the environment, and generate a plan that satisfies the high-level task specification.}
\label{fig:approach}
\end{figure*}

The framework in Figure~\ref{fig:approach} separates instruction translation, referent grounding, and plan synthesis so that each representation can be inspected at the point where it enters the next computation.
The following sections describe these mechanisms. Appendix~\ref{app:limp} provides the prompts, spatial comparators, planning algorithms, skill implementations, and evaluation details accompanying the study.

\subsection{Language Instruction Module}
\label{sec:language-instruction-section} 
The language instruction module uses an LLM $\psi$ to translate a natural language instruction $l$ into a linear temporal logic specification $\varphi_{l}$ with a novel composable syntax for referent disambiguation. Translation proceeds through two stages of in-context learning. The first stage prompts $\psi$ to translate $l$ into a conventional LTL formula $\phi_{l}$ where propositions refer to open-vocabulary objects. The second stage takes $l$ and $\phi_{l}$ as input and prompts $\psi$ to generate a new formula $\varphi_{l}$ with predicate functions corresponding to parameterized robot skills.

We define three predicate functions—\textbf{near}, \textbf{pick}, and \textbf{release}—for the primitive navigation and manipulation skills required for multi-object goal navigation and mobile pick-and-place. Predicate functions in $\varphi_l$ are parameterized by \textit{Composable Referent Descriptors} (CRDs), our novel propositional expressions representing specific referent instances by chaining comparators that encode descriptive spatial information. For example, the instruction \textit{``the yellow cabinet above the fridge that is next to the stove’’} can be represented with the CRD:
\begin{equation}
\text{yellow\_cabinet}::\text{isabove}( \text{fridge}::\text{isnextto}(\text{stove})).
\end{equation}
This specifies that there is a fridge \textit{next to} a stove, and the desired yellow cabinet is \textit{above} that fridge. CRDs are constructed from a set of 3D spatial comparators~\citep{jatavallabhula_conceptfusion_2023} defined in our prompting strategy.

\begin{figure}
\centering
\fbox{\includegraphics[scale=0.19]{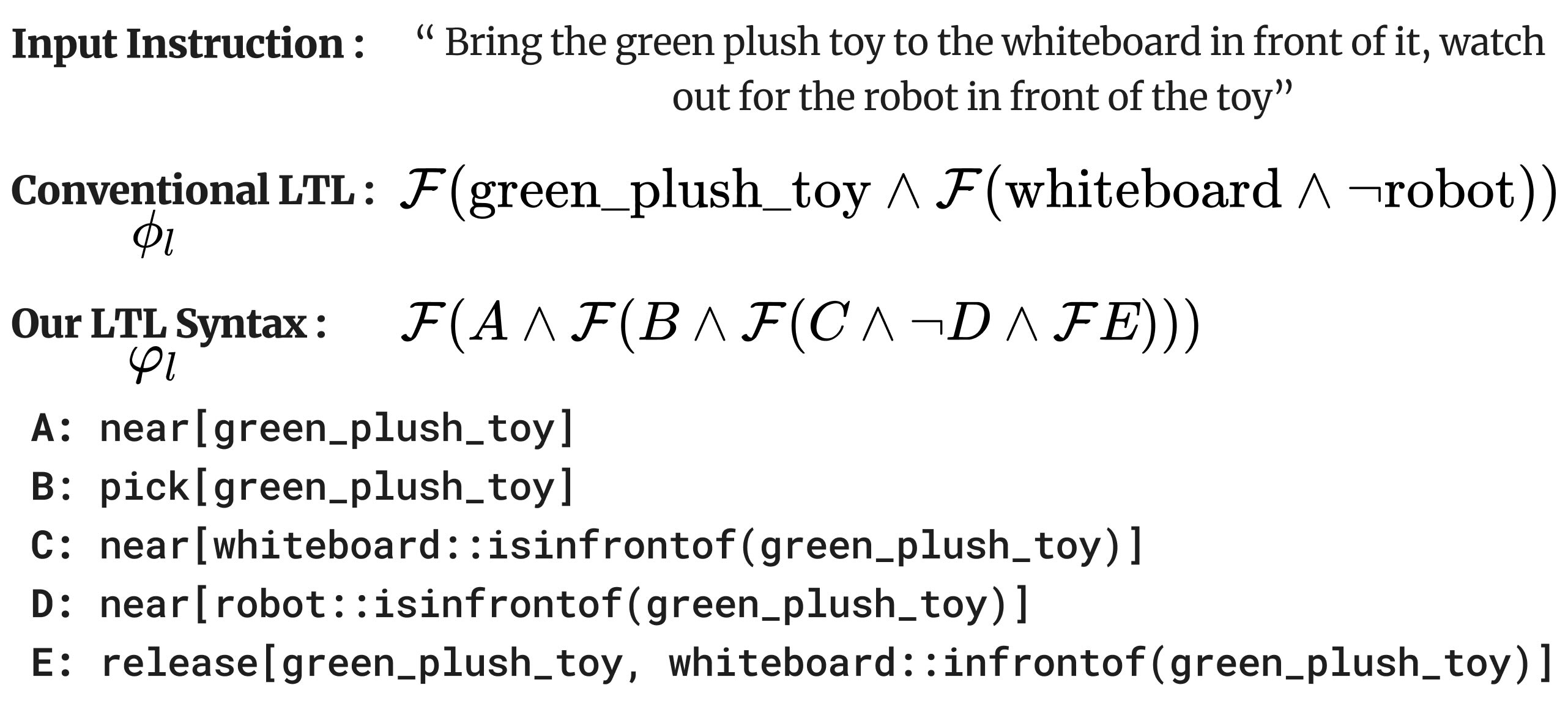}}
\caption[Two-stage translation from language to task specifications]{An instruction is first translated into a conventional LTL formula $\phi_l$ that loosely captures the desired temporal occurrence of referent objects, then into our LTL syntax $\varphi_l$ with predicate functions that temporally chain required robot skills parameterized by composable referent descriptors.}
\label{fig:language_module}
\end{figure}

LIMP accepts varied phrasing and composable spatial descriptions rather than requiring a fixed instruction template~\citep{yenamandra_homerobot_2023,liu_demonstrating_2024}. Its ability to resolve a particular instruction nevertheless depends on the translation and grounding models. The LLM $\psi$ directly samples the entire LTL formula $\varphi_{l}$ with predicate functions parameterized by CRDs using appropriate spatial comparators based on the instruction’s context. Figure~\ref{fig:language_module} illustrates the result of our two-stage prompting strategy.

\vspace{1mm}

\noindent\textbf{Inspecting the Language Translation}: Errors in referent extraction or temporal structure can change the task presented to the planner.
LIMP's symbol verification node (Figure~\ref{fig:approach}) presents the extracted CRDs and the implied task structure to the user, then reprompts the LLM when the user supplies a correction.
A deterministic translation of the specification's automaton into English statements supports this inspection without requiring the user to read the formula directly.
Human review of translated constraints also appears in related approaches~\citep{yang_plug_2024}; here, it provides an opportunity to correct the interpretation before planning.
The evaluation in Tables~\ref{tab:instruction-spatial} and~\ref{tab:temporal-tamp} measures the initial translations without this human correction step.
Appendix~\ref{sec:eval-details} reports additional results with interactive reprompting.

\subsection{Spatial Grounding Module}
\label{sec:spatial-grounding-section}
\begin{figure*}
\centering

\includegraphics[width=1\textwidth]{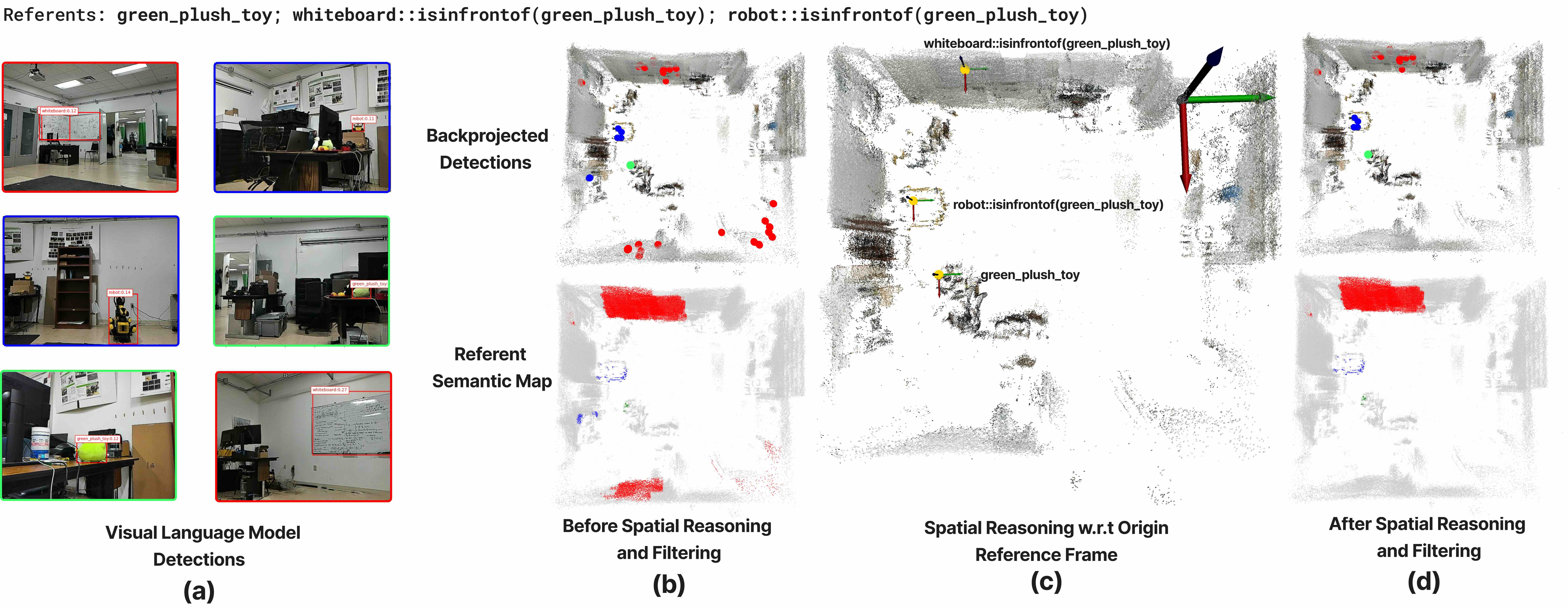}

\captionof{figure}[Spatial grounding of instruction referents]{\textbf{[A]} Our spatial grounding module leverages a VLM to detect all referent occurrences from prior observations of the environment. \textbf{[B]} An initial semantic map with all detected referent instances is generated by backprojecting pixels in segmented referent masks unto the 3D map. \textbf{[C]} Each referent’s spatial comparators is resolved with respect to the origin coordinate frame of reference. [D] Failing instances are filtered out to obtain a Referent Semantic Map (RSM) that localizes the exact referent instances described in the instruction.}
\label{fig:spatial}
\end{figure*}

This module detects and localizes specific object instances referenced in an instruction. The spatial grounding module extracts CRDs from the translated formulas and uses the vision-language models OWL-ViT~\citep{minderer_simple_2022} and SAM~\citep{kirillov_segment_2023} to detect and segment all referent occurrences from the robot's prior observations of the environment. It backprojects pixels in these segmentation masks into the 3D map using the calibrated geometry described in Section~\ref{sec:background:grounding}, creating an initial semantic map of the detected instruction referents. From the example in Figure~\ref{fig:language_module}, occurrences of \textit{green\_plush\_toy}, \textit{whiteboard}, and \textit{robot} are detected, segmented, and backprojected onto the map (Figure~\ref{fig:spatial}[a\&b]).

To obtain the specific object instances described in the instruction, the module evaluates the 3D spatial comparators in each referent’s CRD––recall that CRDs are propositional expressions and can be evaluated as true or false. We define eight spatial comparators (\textit{isbetween}, \textit{isabove}, \textit{isbelow}, \textit{isleftof}, \textit{isrightof}, \textit{isnextto}, \textit{isinfrontof}, \textit{isbehind}) to reason about spatial relationships based on backprojected 3D positions. Since all backprojected positions are relative to an origin coordinate system, our spatial comparators are resolved from the perspective of this origin position as shown in Figure \ref{fig:spatial}[c]. This type of relative frame of reference (FoR) when describing spatial relationships between objects, in contrast to an absolute or intrinsic FoR, is dominant in English \citep{majid_can_2004}, and is a logical choice for our work.

Using the 3D position of each referent’s center mask pixel as its representative position, the module resolves a referent with a spatial description by applying the appropriate spatial comparator to all detected pairs of the desired referent and comparison landmark objects. This filtering process yields a Referent Semantic Map (RSM) that localizes specific object instances described in the instruction as shown in Figure \ref{fig:spatial}[d].

\vspace{1mm}

\noindent\textbf{VLM Verification}: Object-detector misclassifications are the main source of error in this module. LIMP provides 3D visualizations through which users can inspect referent localization; the reported system does not implement interactive correction of these misclassifications.

\subsection{Task and Motion Planning Module}
\label{sec:task-motion-section}
\begin{figure*}
\centering
\includegraphics[width=1\textwidth]{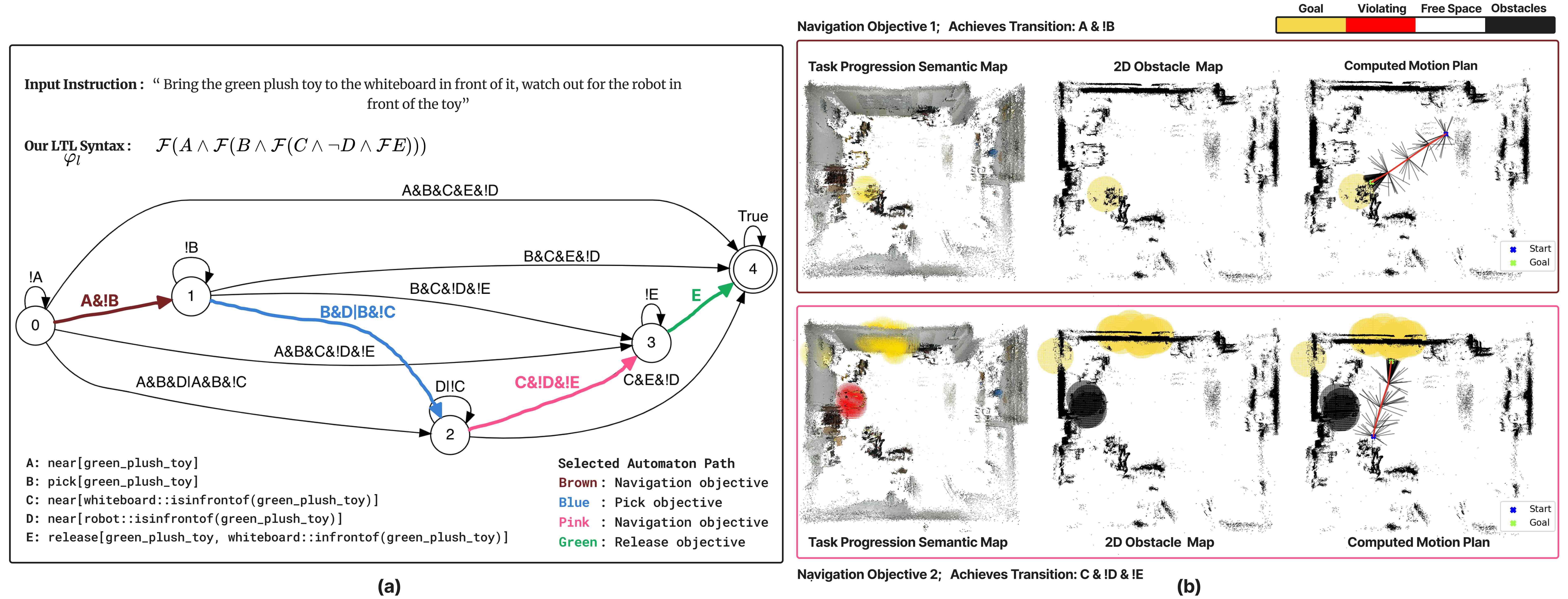}
\vspace{-0.4cm}
\captionof{figure}[Task and motion planning with task progression semantic maps]{\textbf{[A]} A given instruction translated into our LTL syntax $\varphi_l$ can be compiled into an equivalent finite-state automaton that captures the temporal constraints of the task. A path through this automaton is selected with a strategy that incrementally picks the next progression state from the initial state to the accepting state. The robot then executes the manipulation options and navigation behaviors dictated by this high-level task plan. \textbf{[B]} To execute navigation objectives our approach generates a task progression semantic map (TPSM) that augments the environment with state transition constraints, localizing goal (yellow) and avoidance (red) regions. Generated TPSMs are converted into 2D obstacle maps for constraint-aware continuous path planning.}
\label{fig:tamp}
\end{figure*}

Finally, our TAMP module synthesizes and sequences navigation and manipulation behaviors to produce a plan that satisfies the temporal and spatial constraints expressed in the given instruction.
\vspace{1mm}

\noindent\textbf{Progressive Motion Planner (PMP)}:
The TAMP module compiles the supported co-safe LTL specification with parameterized robot skills into a finite-state automaton (Figure~\ref{fig:tamp}[a]) to guide task and motion planning. A path from the initial to the accepting state in this automaton is a high-level task plan that interleaves navigation and manipulation objectives required to satisfy the instruction. The planner selects such a path with a strategy that incrementally selects the next progression state until the accepting state is reached, ensuring the plan obeys all temporal subgoal objectives.  As shown in Figure~\ref{fig:tamp}[a], automaton states are connected by transition edges representing the logical expressions required for transitions. For each transition, our algorithm executes the necessary low-level behaviors: for manipulation subgoals, it executes the appropriate parameterized skill; for navigation subgoals, it dynamically generates Task Progression Semantic Maps (TPSMs) to localize goal and constraint regions and performs continuous path planning using the Fast Marching Tree algorithm (FMT$^*$)~\citep{janson_fast_2015}.

\vspace{1mm}

\noindent\textbf{Task Progression Semantic Maps (TPSM)}: A TPSM augments a 3D scene with navigation constraints specified by logical state transition expressions, enabling goal-directed, constraint-aware navigation. Regions of interest in a TPSM are defined using a nearness threshold specifying proximity to an object. This threshold can be set globally or included in the language instruction module’s prompting strategy, allowing an LLM to infer its value based on the instruction. Like our spatial grounding module, TPSMs are agnostic to temporal logic representations and can be used with various planning approaches for semantic constraint-aware motion planning. For the ground mobile robot used in the evaluation, the module converts 3D TPSMs into 2D geometric obstacle maps, where constraint regions are treated as obstacles (Figure~\ref{fig:tamp}[b]). The spatial representation also permits direct three-dimensional search, although the reported evaluation concerns ground mobile manipulation. \looseness=-1


\section{Evaluation} 
\label{sec:experiments}
Our evaluations test the hypothesis that translating natural language instructions into LTL expressions and dynamically generating semantic maps enables robots to accurately interpret and execute instructions in large-scale environments without task-specific training. We focus on three key questions: (1) Can our language instruction module interpret complex, ambiguous instructions? (2) Can our spatial grounding module resolve specific object instances described in instructions? (3) Can our TAMP algorithm generate constraint-satisfying plans?

To answer these questions, we compare LIMP with two baselines: an LLM task planner (NLMap-Saycan~\citep{chen_open-vocabulary_2023}) and an LLM code-writing planner (Code-as-Policies~\citep{liang_code_2023}), representing language-model task sequencing and code generation for robot instruction following. Both baselines use the same LLM (GPT-4-0613), prompting structure, and in-context learning examples as our language instruction module. In Code-as-Policies, in-context examples are converted into language model-generated program (LMP) snippets~\citep{liang_code_2023}. To ensure competitive performance, we integrate our CRD syntax, spatial grounding module, and low-level robot control into these baselines, allowing them to query object positions, use our FMT$^*$ path planner, and execute manipulation skills.

We also evaluate ablations of our two-stage language instruction module due to its importance in instruction following. In our full approach, the first stage prompts an LLM to generate a conventional LTL formula $\phi_l$ from instruction $l$ by dynamically selecting relevant in-context examples from a standard dataset \citep{pan_data-efficient_2023} based on cosine similarity. Our first ablation selects in-context examples randomly; and the second ablation skips this stage entirely, directly sampling our LTL syntax with parameterized robot skills $\varphi_l$ from $l$.

We conduct a large-scale evaluation across five real-world environments on a diverse task set of 150 instructions from multiple prior works~\citep{chen_open-vocabulary_2023, liu_demonstrating_2024, liu_grounding_2023}. This task set consists of 24 tasks with fine-grained object descriptions (NLMD), 25 tasks with complex language (NLMC), 25 tasks with simple structured phrasing (OKRB), 37 tasks with complex temporal structures (CT) and 39 tasks with descriptive spatial constraints and temporal structures (CST). Below are examples from each task category illustrating the variety in complexity:

\vspace{1mm}

\begin{lstlisting}[language=instructions]
NLMD: Put the brown multigrain chip bag in the woven basket
NLMC: I like fruits, can you put something I would like on the yellow sofa for me
OKRB: Move the soda can to the box
CT: Visit the purple door elevator, then go to the front desk and then go to the kitchen table, in addition you can never go to the elevator once you have seen the front desk
CST: I have a white cabinet, a green toy, a bookshelf and a red chair around here somewhere. Take the second item I mentioned from between the first item and the third. Bring it the cabinet but avoid the last item at all costs.
\end{lstlisting}

\vspace{1mm}


To evaluate instruction understanding, we introduce performance metrics: \textbf{referent resolution accuracy}, \textbf{avoidance constraint resolution accuracy}, and \textbf{spatial relationship resolution accuracy}. These metrics utilize the word error rate (WER), widely used in speech recognition to quantify the difference between a reference and a hypothesis transcription by computing the minimal number of substitutions, deletions, and insertions needed to transform the hypothesis into the reference. WER is calculated as $WER = \frac{S + D + I}{N}$, where  $S$ is substitutions, $D$ deletions,  $I$ insertions, and  $N$ is the total number of words in the reference.  In our work:
\begin{itemize}
\item \textbf{Referent resolution accuracy} compares extracted referents in the generated LTL formula to ground truth referents.
\item \textbf{Avoidance constraint resolution accuracy} compares referents to avoid in the LTL formula (denoted by the unary negation operator) against ground truth avoidance referents.
\item \textbf{Spatial relationship resolution accuracy} compares generated Composable Referent Descriptors (CRDs) in the LTL specification with ground truth CRD expressions.
\end{itemize}

We also define \textbf{temporal alignment accuracy} and \textbf{planning success rate}. A plan is temporally aligned if the sequence of subgoals matches the instructor’s intention, and successful if it satisfies all spatial and temporal constraints specified in the instruction. Achieving a high plan success rate is challenging, requiring accurate referent and avoidance constraint resolution, spatial grounding, and temporal alignment. We report average word error rates for each baseline in Table~\ref{tab:instruction-spatial} and the percentage of successful and temporally accurate plans in Table~\ref{tab:temporal-tamp}. \looseness=-1
\begin{table*}[t]
\centering
\caption{Performance comparison of one-shot instruction understanding and spatial resolution.}
\label{tab:instruction-spatial}
\resizebox{\textwidth}{!}{
\begin{tabular}{@{}lccc@{}}
\toprule
\textbf{Approach} & 
\makecell{Referent Resolution \\ Accuracy \\ (Average WER)  $\downarrow$} & 
\makecell{Avoidance Constraint Resolution \\  Accuracy \\ (Average WER)  $\downarrow$} & 
\makecell{Spatial Relationship Resolution \\  Accuracy \\ (Average WER)  $\downarrow$} \\
\midrule
NLMap-Saycan & 0.09 & 0.12 & 0.05 \\
Code-as-Policies & 0.22 & 0.24 & 0.06 \\
Limp Single Stage Prompting & 0.09 & 0.11 & 0.05 \\
Limp Two Stage Prompting [Random Embedding] & 0.08 & 0.11 & 0.03 \\
\textbf{Limp Two Stage Prompting [Similar Embedding]} &  \textbf{0.07} &  \textbf{0.04} &  \textbf{0.03} \\
\bottomrule
\end{tabular}
}

\end{table*}

\begin{table*}[t]
\centering
\caption{Performance comparison of one-shot temporal alignment and plan success rate.}
\label{tab:temporal-tamp}
\resizebox{\textwidth}{!}{
\begin{tabular}{@{}lcccccccccc@{}}
\toprule
\textbf{Approach} & 
\multicolumn{5}{c}{\makecell{Temporal Alignment Accuracy \\ (\% of instructions) $\uparrow$}} & 
\multicolumn{5}{c}{\makecell{Planning Success Rate \\ (\% of instructions) $\uparrow$}} \\
\cmidrule(lr){2-6} \cmidrule(lr){7-10}
& NLMD & NLMC & OKRB & CT & CST & NLMD & NLMC & OKRB & CT & CST \\  
\midrule
NLMap-Saycan & 88\% & \textbf{96\%} & 100\% & 32\% & 41\% & 75\% & \textbf{96\%} & 100\% & 35\% & 38\%\\
Code-as-Policies & 58\% & 68\% & 100\% & 35\% & 38\% & 46\% & 68\% &100\% & 38\% & 38\%\\
Limp Single Stage Prompting & 79\% & 64\% & 100\% & 68\% & 74\% & 63\% & 60\% & 100\% & 62\% & 62\%\\
Limp Two Stage Prompting [Random Embedding] & 83\% & 68\% & 100\% & 76\% & 85\% & 79\% & 68\% & 100\% & 57\% & 72\%\\
\textbf{Limp Two Stage Prompting [Similar Embedding]} & \textbf{88\%} & 80\% & \textbf{100\%} & \textbf{76\%} & \textbf{92\%} & \textbf{79\%} & 76\% & \textbf{100\%}  & \textbf{65\%} & \textbf{79\%} \\
\bottomrule
\end{tabular}
}

\end{table*}

\section{Discussion} 
\label{sec:limp:discussion}
Beyond the verification benefits of symbolic planning, LIMP outperforms baselines in most task sets, notably in complex temporal planning and constraint avoidance. While NLMap-Saycan and Code-as-Policies effectively generate sequential subgoals, they struggle with strict temporal constraints—for example, avoiding a specific referent while approaching another. LIMP incorporates the instruction's avoidance regions into geometric path planning, accounting for its stronger performance on the CT and CST task sets. As shown in Table~\ref{tab:temporal-tamp}, LIMP underperforms only against NLMap-Saycan in the NLMC task category. This task set, introduced in the same paper as the baseline \citep{chen_open-vocabulary_2023} (which outperforms LIMP), includes instructions with implicit details such as: “I like fruits, can you put something I would like on the yellow sofa for me.” NLMap-Saycan is better suited to infer and generate plans with possible fruit options, whereas our few-shot LTL translation process is not designed for this.\looseness=-1

\section{From Grounded Intent to Continued Autonomy}
\label{sec:limp:conclusion}

LIMP connects expressive instructions to explicit task constraints and the scene geometry needed for planning.
Instruction-conditioned semantic maps let the robot work beyond prespecified landmarks, while inspection exposes the translated task and its referents to the user.

This result depends on the task and world models.
The planner handles co-safe specifications, whose satisfaction has a finite good-prefix witness (Section~\ref{sec:background:temporal-logic}).
Grounding remains sensitive to perception and spatial interpretation, and plan synthesis does not guarantee physical success or optimality.
Suitable manipulation skills and a static mapped environment are still assumed.

These assumptions motivate the next contributions.
Successful manipulation can invalidate the representation used to plan it; later tasks must therefore be grounded in a world that accounts for earlier actions.
Chapter~\ref{ch:realm} develops this evolving representation.
Chapter~\ref{ch:pacts} then addresses how additional behaviors can be learned together with the symbolic information needed to compose them.


\chapter{Evolving World Representations for Planning}
\label{ch:realm}
\begin{quote}
\small
This chapter presents joint work by Benedict Quartey and George Konidaris, developed in \textit{REALM: Action-Conditioned Spatial World Models for Long-Horizon Mobile Manipulation}.
The authorial ``we'' refers to this collaboration.
\end{quote}
\vspace{0.75\baselineskip}

\noindent
Understanding an instruction establishes what the robot should accomplish; it does not keep the world used to interpret that instruction current.
Chapter~\ref{ch:limp} connected language to task constraints and grounded those constraints in a captured environment.
That connection becomes vulnerable as soon as manipulation changes the scene.
An object can keep its identity while acquiring a new location, and the next request may depend on both facts.
For autonomy to extend across successive tasks, the representation supporting one plan must account for its consequences before it supports another.

This chapter develops \emph{evolving representations of the world}, addressing the world interface of trust.
REALM extends the language-to-plan framework with a persistent spatial representation from which grounding, planning geometry, and visual inspection draw their information.
A planned manipulation predicts a change to that representation; local visual reconciliation produces a successor for the next instruction.
The central contribution is continuity between tasks: object identity, geometry, and the information supplied to the planner advance together.
A sequential study examines this continuity in a captured scene using simulated manipulation plans and synthetic post-action imagery, distinguishing the predicted world from evidence of what a physical robot has actually changed.

\section{Planning Across World Changes}
\label{sec:intro}

\begin{figure}
    \centering
    \includegraphics[width=\textwidth]{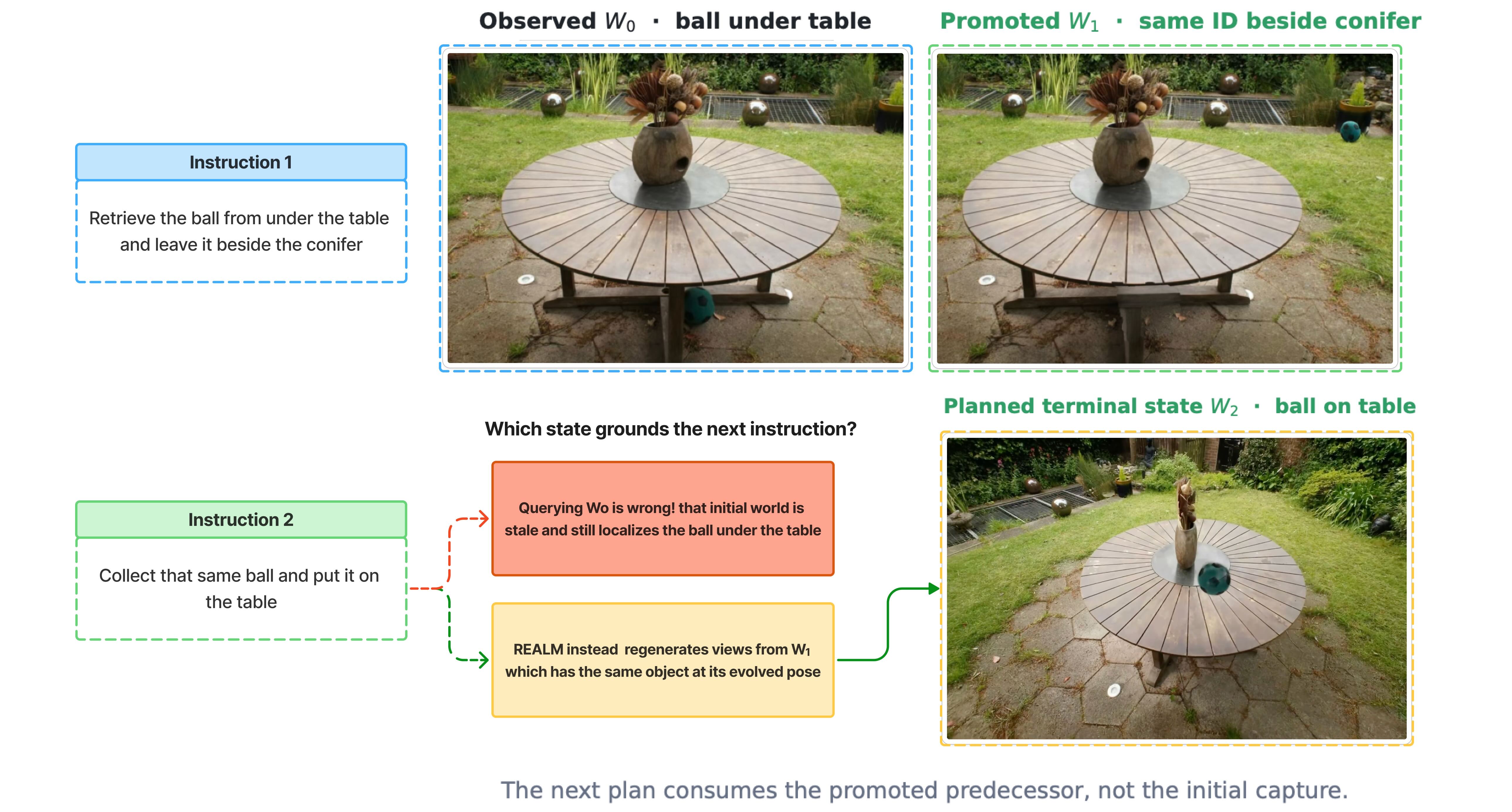}
  
    \caption[A robot can invalidate the map it just used]{\textbf{A robot can invalidate the map it just used.} Instruction 1 moves the ball from under the table to beside the conifer. If instruction 2 is grounded against the original world $W_0$, the ball is still represented under the table. \method instead promotes the evolved world $W_1$, preserves the ball's identity, regenerates the representations used for grounding and planning, and carries the task forward to $W_2$. The images are Gaussian renders from the simulated garden sequence; the final panel illustrates the planned terminal configuration. The displayed instructions are abbreviated; Section~\ref{sec:evaluation} gives the evaluated wording.}
\label{fig:realm:splash}
\end{figure}

Consider the sequence in Fig.~\ref{fig:realm:splash}. The robot first retrieves a blue ball from beneath a table and leaves it beside a conifer. A later instruction refers to that same ball and asks the robot to place it on the table. The first plan may be perfectly grounded and feasible, but its success invalidates the map from which it was produced. If the next instruction is grounded against the original reconstruction, the robot searches beneath the table for an object that its own action moved. This is not principally a failure of language understanding or motion planning, but a failure to represent the world as something that evolves under robot intervention. Reconstructing the scene from scratch does not completely solve the problem: the moved object may be occluded, its former location may expose unseen background, and a new detection need not preserve the identity used by the preceding plan. The robot must connect what it intended to change, what it subsequently observes, and what its next task is allowed to believe.

The problem is to preserve this continuity without reconstructing the entire scene from scratch after each action.
The functional view of world models in Section~\ref{sec:background:world-models} identifies the capabilities required: grounding in the environment, prediction of task-relevant change, and an account of the intervention that causes it.
REALM supplies these capabilities through an explicit Gaussian scene and mechanisms for changing it under planned manipulation.
The Gaussian representation provides editable geometry and rendered views; the transition and update mechanisms make it useful as a spatial world model.

\method maintains an explicit Gaussian world as the persistent base representation and constructs the semantic, geometric, and visual abstractions needed for task-and-motion planning at runtime from the current world state. Grounded objects, task-progression semantics, digital-twin geometry, collision clouds, and rendered observations are instruction-specific abstractions of that state. For a planned manipulation, \method first predicts its geometric consequence using an analytic forward model, then reconciles its visual consequences by constructing registered post-action evidence, and finally promotes a successor from which the next instruction is grounded and planned.

This sequence separates geometry predicted from the plan, appearance inferred from visual evidence, and the successor accepted for reuse after validation. \emph{Predicting physical change} applies a forward model based on the object's terminal pose in the validated plan. \emph{Reconciling visual consequences} renders the prediction from calibrated viewpoints and uses a pretrained image-generation foundation model to synthesize appearance changes within geometry-derived masks--physical observations can complement or replace these synthetic re-observations through the same interface. \emph{Committing the updated world} uses the resulting RGB-D evidence to locally update the affected Gaussians, validates the result, and promotes the accepted successor. Stable ownership binds grounded objects to the Gaussian rows that move, while immutable history preserves the worlds produced over time.

The sequential study in Figure~\ref{fig:realm:splash} follows one object through this process in a captured real scene, using synthetic post-action RGB-D. The ball begins under the table in $W_0$. Instruction 1 derives its grounding and planning abstractions from $W_0$ and predicts the ball beside the conifer. After reconciliation and promotion, instruction 2 derives a new grounding, twin, collision cloud, and plan from $W_1$, associates the ball with the same identity at its evolved location, and predicts $W_2$ on the table. Across both transitions, \method transforms the same 254 object-owned Gaussians, preserves every non-moved row exactly during prediction, improves agreement with synthetic targets on held-out views after reconciliation, and recovers every ancestor exactly. The chapter makes four contributions:
\begin{enumerate}
\item an action-conditioned spatial world model that evolves one persistent, renderable 3D state authority from which instruction-specific abstractions used for language grounding and planning can be generated;
\item stable object ownership that gives robot actions explicit support inside a monolithic Gaussian scene and preserves object identity across tasks;
\item a prediction, reconciliation, and promotion process that distinguishes geometric predictions based on planned actions from appearance inferred through visual evidence while retaining an immutable world history; and
\item a sequential study with simulated mobile-manipulation plans that regenerates the next instruction's abstractions from the world produced by the preceding plan rather than from the initial capture.
\end{enumerate}

\section{Related Work}
\label{sec:related}

\paragraph{Semantic scenes and dynamic memory.}
Hydra builds hierarchical metric-semantic scene graphs online~\citep{hughes2022hydra}, and ConceptGraphs constructs open-vocabulary object-centric graphs for perception and planning~\citep{gu2023conceptgraphs}. F3RM distills language-aligned features into 3D fields for few-shot manipulation~\citep{shen2023f3rm}; D$^3$Fields combines dynamic geometry, semantic descriptors, and instance structure for zero-shot rearrangement~\citep{wang2023d3fields}. Dynamic Scene Representation explicitly studies permanence, amodal completion, and spatiotemporal continuity for predictive manipulation~\citep{xu2020dynamic}. DynaMem updates an online point-cloud memory as objects move, appear, and disappear~\citep{liu2024dynamem}, and DovSG updates an open-vocabulary scene graph for long-term mobile manipulation~\citep{yan2024dovsg}. \method shares the aim of persistent semantic state, but couples each object identity to the rows transformed by a specified planned action and to the history consumed by later planning.

\paragraph{Gaussian worlds and continual reconstruction.}
3D Gaussian Splatting provides an explicit scene representation that is both differentiable and efficiently renderable~\citep{kerbl2023gaussians}. CL-Splats identifies changed regions, optimizes them locally, and retains temporal reconstruction history~\citep{ackermann2025clsplats}. Gaussian digital twins can also connect semantic reconstruction to collision geometry and physical pick-and-place planning~\citep{sun2026digitaltwin}. GaussMemory learns task-dependent retention and update in a persistent Gaussian memory~\citep{hu2026gaussmemory}. \method builds on these representational capabilities, placing them inside a robot-facing state lifecycle in which a plan predicts a successor and promotion determines what the next task sees.

\paragraph{Action-conditioned world models.}
Predictive world models for manipulation take many forms. Physically Embodied Gaussian Splatting couples particles and Gaussians to predict and visually correct physical dynamics~\citep{abouchakra2024pegs}. DreMa constructs compositional Gaussian and physics worlds for imagined robot data~\citep{barcellona2024drema}. GWM and MRO-GWM learn action-conditioned future Gaussian states for policy learning and model-predictive control~\citep{lu2025gwm,kreber2026mrogwm}; GaussianDream decodes future Gaussians as dense supervision for vision-language-action learning~\citep{zhang2026gaussiandream}; GAF represents motion and interaction through a 4D Gaussian action field~\citep{chai2025gaf}; and ParticleFormer predicts multi-object, multi-material point-cloud dynamics~\citep{huang2025particleformer}. \method occupies a complementary regime: the task planner supplies a validated rigid-object outcome, and the world model turns that outcome into a persistent, inspectable, evidence-reconciled state for future language-guided behavior.

Across these literatures, rich spatial representations, action prediction, and language-grounded planning are often developed as separate capabilities. The central role of \method is to close the loop among them: the representation used to interpret an instruction is updated under its planned action and becomes the representation used to interpret what comes next.

\section{One World, Many Instruction-Specific Views}
\label{sec:formulation}

Using the Gaussian representation introduced in Section~\ref{sec:background:gaussians}, let a scene contain rows
\begin{equation}
g_i=(\mu_i,s_i,q_i,\alpha_i,c_i),\qquad i\in\{1,\ldots,N_t\},
\label{eq:gaussian}
\end{equation}
where $\mu_i\in\mathbb R^3$ is the mean, $s_i\in\mathbb R^3$ the componentwise log-scale (rather than the positive scale used in Chapter~\ref{ch:background}), $q_i$ a unit quaternion specifying orientation, $\alpha_i\in[0,1]$ the opacity, and $c_i$ the spherical-harmonic appearance coefficients. Gaussian parameters are indexed within a particular revision; their time index is suppressed for readability. A stable object registry contains identities $o\in\objects_t$, poses $X_t^o\in\SE(3)$, and an ownership function
\begin{equation}
\omega_t(i)\in\{0\}\cup\mathbb{N}^{+},
\label{eq:owner}
\end{equation}
where zero denotes static or unowned scene support and a positive value identifies one movable object. The registry $\objects_t$ stores these ownership assignments together with each object record. A pose $X_t^o=(R_t^o,t_t^o)$ maps object-local coordinates into the common world frame: $x^{\mathrm{world}}=R_t^o x^{\mathrm{object}}+t_t^o$.

Let $t$ index planned manipulation transitions. Prediction and reconciliation create distinct revisions at the same logical time; their parent links distinguish them. We represent a persistent world revision as
\begin{equation}
W_t=(\gaussians_t,\objects_t,H_t,M_t).
\label{eq:state}
\end{equation}
$H_t$ links the state to its immutable parent, while $M_t$ records the action, evidence provenance, and validation information that explain the transition. This state is the persistent authority. It is not itself a fixed planner input: an instruction $\ell_t$ asks \method to construct the abstractions needed at that moment,
\begin{equation}
\mathcal{A}(W_t,\ell_t)=(G_t,S_t,D_t,C_t),
\label{eq:abstraction}
\end{equation}
where $G_t$ contains grounded objects, $S_t$ is the task-progression semantic state, $D_t$ is task-relevant digital-twin geometry, and $C_t$ is the planning collision representation. Grounding views rendered from $W_t$ support the construction of $G_t$; they are not a second world model. Planning and subsequent visual inspection are
\begin{align}
\pi_t&=\Pi\!\left(\ell_t;G_t,S_t,D_t,C_t\right),
\label{eq:plan}\\
V_t^{\mathrm{insp}}&=\mathcal R_{\mathrm{plan}}(W_t,\pi_t).
\label{eq:plan-inspection}
\end{align}
Thus inspection renders are produced from a candidate plan rather than used to create it. The instruction determines which abstractions are materialized, but every abstraction, plan, and inspection view remains bound to the same $W_t$.

Suppose the validated plan $\pi_t$ yields a manipulation transition $a_t$ with released object pose $\widehat X_{t+1}^{o}$. The transition $a_t$ includes the planned endpoint. For an accepted reconciled successor, \method evolves the world in three stages:
\begin{align}
\widehat{W}_{t+1}&=\operatorname{Predict}(W_t,a_t),\nonumber\\
B_{t+1}&=\operatorname{Reconcile}(W_t,\widehat{W}_{t+1}),\nonumber\\
W_{t+1}&=\operatorname{Commit}(\widehat{W}_{t+1},B_{t+1}).
\label{eq:loop}
\end{align}
$\operatorname{Predict}$ transforms the known object support according to the validated endpoint; $\operatorname{Reconcile}$ renders that prediction and constructs registered post-action evidence $B_{t+1}$; and $\operatorname{Commit}$ uses that evidence to update only affected Gaussian support, validates the result, and grants the successor current authority. The labels \emph{known}, \emph{inferred}, and \emph{trusted} in the method diagram refer to these roles: an explicit planned endpoint, visually inferred consequences, and a successor accepted by the representation checks. They do not establish that the predicted endpoint was physically realized.

For calibrated view $k$, let $\kappa_k$ contain the camera intrinsics and pose in the world frame. The renderer produces predicted RGB $\widehat I$ and depth $\widehat Z$; the image-editing model $E$ produces RGB only:
\begin{align}
(\widehat I_{t+1}^{(k)},\widehat Z_{t+1}^{(k)})
  &=\mathcal R(\widehat W_{t+1},\kappa_k),
  \label{eq:predicted-render}\\
I_{t+1,\mathrm{syn}}^{(k)}
  &=E\!\left(I_t^{(k)},\widehat I_{t+1}^{(k)},\mathcal M_{t\rightarrow t+1}^{(k)},r_o\right),
  \label{eq:synthetic-observation}\\
Z_{t+1,\mathrm{syn}}^{(k)}&=\widehat Z_{t+1}^{(k)}.
  \label{eq:synthetic-depth}
\end{align}
Here $I_t^{(k)}$ is the parent RGB used for editing, $\mathcal M_{t\rightarrow t+1}^{(k)}$ contains source-removal and target-insertion masks, and $r_o$ is an object-appearance reference. The registered bundle is
\begin{equation}
B_{t+1}=\bigl\{(I_{t+1}^{(k)},Z_{t+1}^{(k)},\mathcal M_{t\rightarrow t+1}^{(k)},\kappa_k)\bigr\}_{k=1}^{n_v},
\label{eq:observation-bundle}
\end{equation}
where $n_v$ counts views, separately from the $N_t$ Gaussian rows. Each view also records provenance and its fitting or validation role. Physical RGB-D can populate the same bundle, with change masks derived from observations. Equations~\ref{eq:synthetic-observation}--\ref{eq:synthetic-depth} describe the synthetic branch used here, pairing generated RGB with depth rendered from the geometric prediction.

Commit lifts the registered multi-view evidence into the Gaussian map. Let $\mathcal L(\widehat{W}_{t+1},B_{t+1})$ denote the existing row indices supported as changed by that evidence, $\mathcal J_{t+1}^{\mathrm{act}}$ the indices owned by objects changed by the validated action, and $\mathcal J_{t+1}^{\mathrm{new}}$ newly introduced support for disoccluded content. The initial active set for fitting is
\begin{equation}
\mathcal J_{t+1}
=\mathcal L(\widehat{W}_{t+1},B_{t+1})
\cup\mathcal J_{t+1}^{\mathrm{act}}
\cup\mathcal J_{t+1}^{\mathrm{new}}.
\label{eq:active-set}
\end{equation}
With $U_{\mathrm{CL}}$ denoting local fitting based on CL-Splats followed by restoration of action-owned geometry and opacity, and $P$ denoting validation and promotion, the final stage expands to
\begin{align}
\widetilde{W}_{t+1}
&=U_{\mathrm{CL}}(\widehat{W}_{t+1},B_{t+1};\mathcal J_{t+1}),\nonumber\\
W_{t+1}&=P(\widetilde{W}_{t+1}).
\label{eq:commit-update}
\end{align}
Together these two operations implement $\operatorname{Commit}$ in Eq.~\ref{eq:loop}.
Only support in $\mathcal J_{t+1}$ and any descendants introduced by local densification may be optimized; unrelated Gaussian support is frozen. Action-owned geometry and opacity are restored to their predicted values, preserving the planned object configuration through visual reconciliation. Equation~\ref{eq:commit-update} describes successful promotion. A validated prediction may first become current while awaiting reconciliation; if the reconciled candidate fails validation, the preceding current revision is retained (Appendix~\ref{app:lifecycle}). After successful promotion, the next instruction calls $\mathcal A(W_{t+1},\ell_{t+1})$ rather than reusing $\mathcal A(W_t,\ell_t)$. This regeneration closes the loop from one instruction to the next.

The evolution is governed by six design requirements (Table~\ref{tab:requirements}). Together they keep the persistent representation local, continuous, reversible, and usable by every task-specific consumer.

\begin{table}[t]
\centering
\caption{Design requirements for an evolving robot world.}
\label{tab:requirements}
\small
\begin{tabularx}{\columnwidth}{@{}p{0.27\columnwidth}X@{}}
\toprule
Requirement & Meaning \\
\midrule
Local change & An action changes the owned object without moving unrelated rows. \\
Stable identity & One object ID connects grounding, prediction, twin geometry, and later reuse. \\
Persistent history & A successor is added without overwriting the world from which it came. \\
Split authority & The planned action fixes predicted object geometry; evidence updates appearance and exposed scene content. \\
Validated promotion & A reconciled candidate replaces the current prediction only after spatial, visual, and lineage checks. \\
Synchronized views & Grounding, planning, rendering, and twin artifacts are generated from one named state. \\
\bottomrule
\end{tabularx}
\end{table}

\section{REALM}
\label{sec:realm:method}

\begin{figure*}[t]
  \centering
  \includegraphics[width=\textwidth]{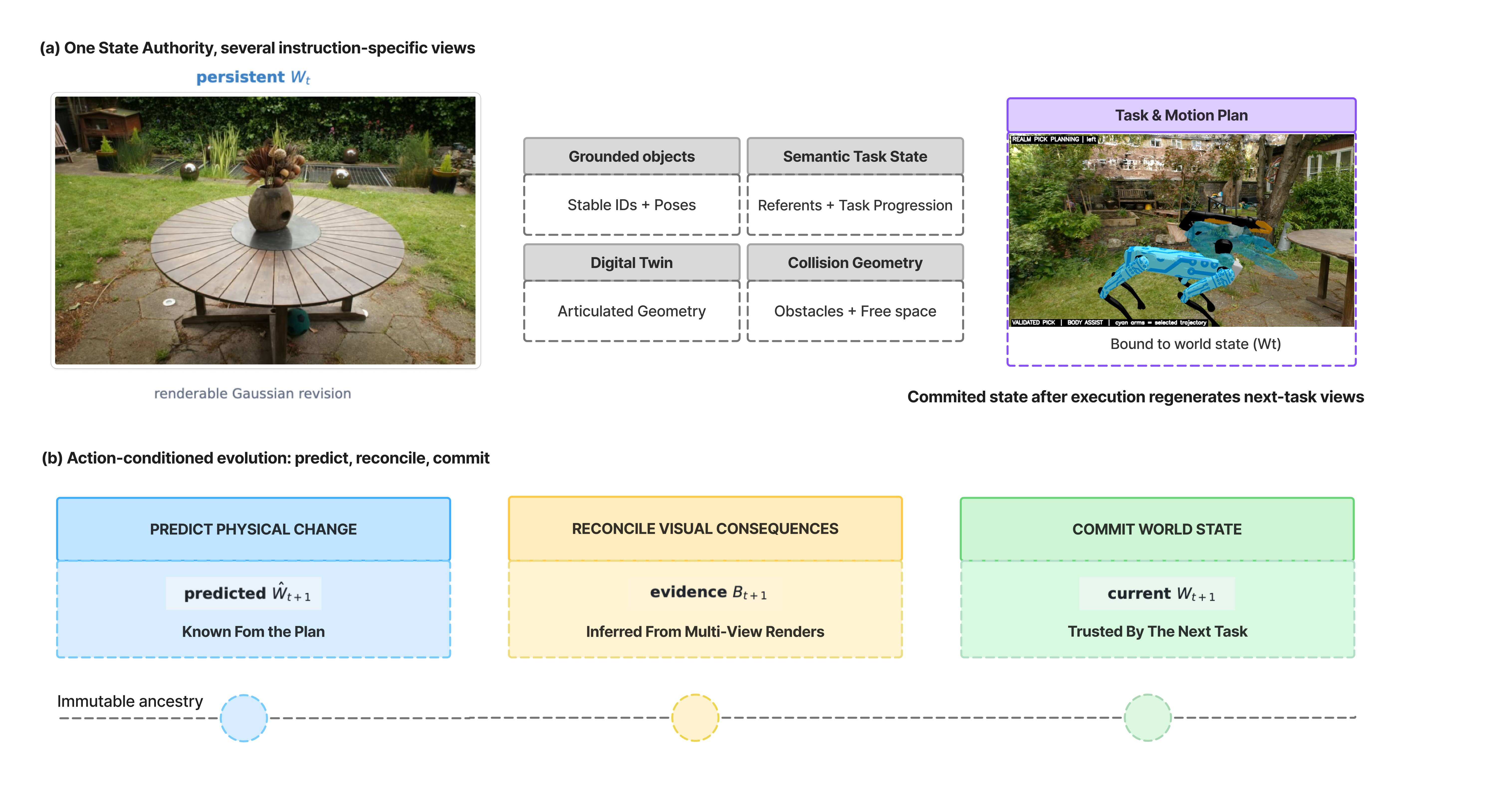}
  \caption[One persistent world, rebuilt for each instruction]{\textbf{One persistent world, rebuilt for each instruction.} At timestep $t$, \method renders grounding views from the current Gaussian world $W_t$ and constructs grounded objects, task-progression semantics, digital-twin geometry, and collision geometry. Together these instruction-specific abstractions produce a task-and-motion plan, which can then be rendered against the same state for inspection. Its validated endpoint predicts physical change (\emph{known}); rendering and re-observation construct the reconciled visual evidence (\emph{inferred}); and commit locally updates, validates, and grants $W_{t+1}$ authority for future use (\emph{trusted}). The next instruction regenerates its own abstractions from $W_{t+1}$, not from the original capture.}
  \label{fig:architecture}
\end{figure*}

At each timestep, \method plays two connected roles. First, it is the base representation from which the robot constructs whatever the current instruction requires. Second, after a plan is produced, it is the model that predicts and absorbs the resulting change. The recurring progression is therefore \emph{represent $\rightarrow$ derive task views $\rightarrow$ ground and plan $\rightarrow$ predict physical change $\rightarrow$ reconcile visual consequences $\rightarrow$ commit and regenerate}. Figure~\ref{fig:architecture} visualizes this progression.

\subsection{From the Persistent World to the Current Instruction}

At mapping time, \method starts from posed scene images and a renderable Gaussian reconstruction. In the garden study, the captured input is the Mip-NeRF~360 image dataset~\citep{barron2022mipnerf360}; depth used for subsequent view lifting is rendered from the reconstruction. The Gaussian scene is the persistent spatial representation, but it is not the task plan. Given an instruction such as ``get the blue ball from under the table and leave it beside the potted lime-green conifer,'' \method first renders views of $W_t$ in which the named referents can be detected. Multi-view detections are lifted into 3D, consolidated into stable object records, and inserted into a task-progression semantic state.

From the same $W_t$, \method constructs only the geometric abstractions required by that instruction. Task-relevant surfaces and objects become a digital twin for manipulation; the remaining geometry becomes a collision cloud for navigation and whole-body planning. LIMP translates the language into temporal constraints and produces navigation, pick, and release trajectories from these state-bound abstractions~\citep{quartey2025verifiably}. Because the underlying world is renderable, the robot and candidate trajectories can also be composed with Gaussian views from several cameras for visual plan inspection. Grounding views, semantic records, twin geometry, collision geometry, and plan renders serve different consumers, but they all depict the same named world.

Each state stores a checkpoint, object registry, digital twin, planning cloud, optional observation bundle, source plan, transition record, and validation report. Artifact digests bind these products to the revision from which they were derived. The scientific purpose of that binding is consistency: a grounded object cannot come from one world while the collision cloud or plan comes from another.

\subsection{Grounding Says What the Task Means; Ownership Says What Moves}
\label{sec:ownership}

Grounding identifies the semantic object referred to by the instruction. Ownership connects that object identity to the part of the persistent 3D state that may follow an action. A broad crop first proposes Gaussian rows. For candidate $i$, \method measures positive depth-consistent votes $p_i$, visible contradictions $n_i$, visible views $v_i$, footprint-supported views $f_i$, fused-surface distance $d_i^S$, oriented-box overflow $d_i^B$, maximum Gaussian scale $\sigma_i^{\max}$ with $\sigma_i=\exp(s_i)$ componentwise, and anisotropy $r_i$. The selected support is
\begin{align}
m_i={}&[p_i\!\ge\!\tau_p]
\land[n_i/\max(v_i,1)\!\le\!\tau_n]
\land[f_i\!\ge\!\tau_f]\nonumber\\
&\land[d_i^S\!\le\!\tau_S]
\land[d_i^B\!\le\!\tau_B]
\land[\sigma_i^{\max}\!\le\!\tau_s h_o]
\land[r_i\!\le\!\tau_r],
\label{eq:selector}
\end{align}
where each bracket is a Boolean test, $\land$ requires every test to pass, and $h_o$ is the maximum object half-extent. The distances $d_i^S,d_i^B$, scales $\sigma_i$, and half-extent $h_o$ use the same world-coordinate units; $\tau_s$ is dimensionless. The selected checkpoint indices and object-local means are stored with the same digital twin used for planning. Empty, duplicate, overlapping, and insufficient supports are rejected: if a background row is owned by the ball, every future action on the ball would incorrectly move that background as well.

\begin{figure*}[t]
  \centering
  \includegraphics[width=\textwidth]{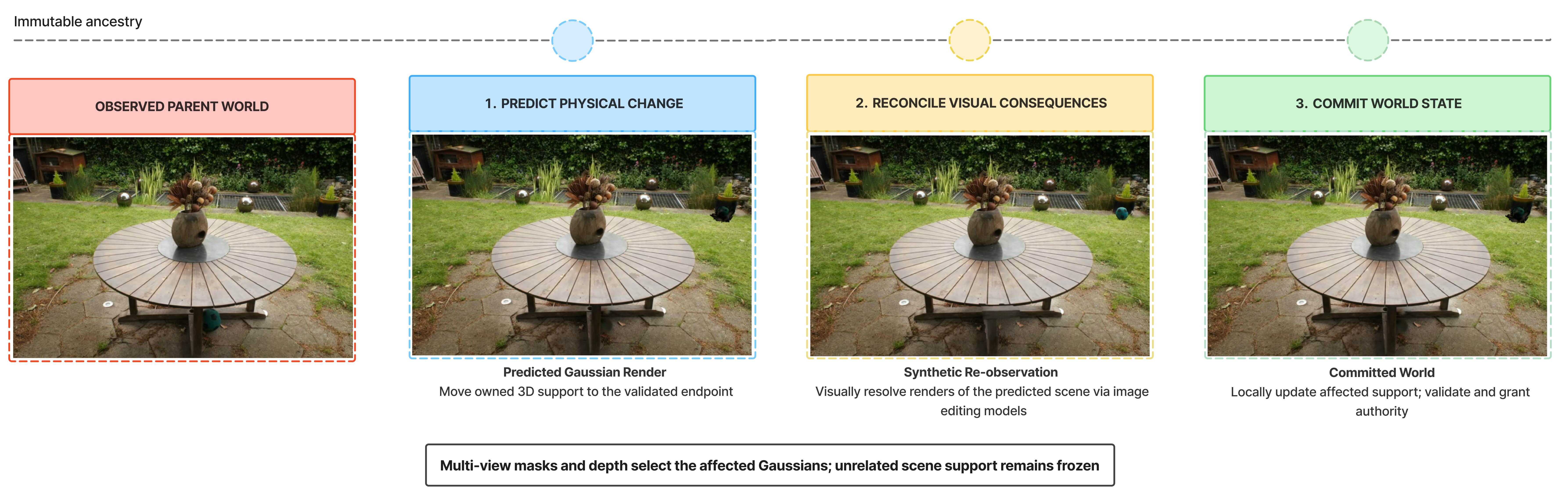}
  \caption[How the world evolves]{\textbf{How the world evolves.} \method predicts the spatial consequence specified by the plan, renders that prediction to reconcile visual consequences as registered multi-view evidence, then commits by locally updating affected Gaussian support, validating the result, and granting it authority for the next instruction. The four panels show one calibrated view from the clean garden run: parent evidence, the predicted Gaussian render, a geometry-constrained synthetic re-observation, and the committed Gaussian world.}
  \label{fig:world-evolution}
\end{figure*}

\subsection{Predict Physical Change (Known from the Plan)}

The task-and-motion plan is not merely an output rendered for inspection. Its validated terminal pose provides an intervention prior for the world model. \method composes the state-bound navigation and manipulation artifacts to obtain the released pose of every changed object. A prediction is accepted only when the plan, trajectories, grounding, checkpoint, and current state agree.

For owned Gaussian $i$, write its object-local mean as $\bar\mu_i=(R_t^o)^\top(\mu_i-t_t^o)$. Its predicted world-frame mean is
\begin{equation}
\mu_i'=\widehat R_{t+1}^{o}\bar\mu_i+\widehat t_{t+1}^{o}.
\label{eq:rigid}
\end{equation}
Its orientation is composed with the relative rotation $\Delta R=\widehat R_{t+1}^{o}(R_t^o)^\top$, and its real spherical-harmonic coefficients are rotated consistently through degree three. In covariance form, the geometric update is
\begin{equation}
\Sigma_i=R(q_i)\operatorname{diag}\!\left(\exp(2s_i)\right)R(q_i)^\top,
\qquad \Sigma_i'=\Delta R\Sigma_i\Delta R^\top.
\label{eq:realm:covariance-transport}
\end{equation}
This preserves the Gaussian's principal scales while rotating its axes. The spherical-harmonic rotation similarly carries its represented directional appearance into the new orientation. Scale and opacity remain unchanged. Every row outside the moved object's support is copied exactly during prediction. The transition simultaneously evolves the object registry, twin, and planning cloud and publishes a reversible delta linking the predicted world to its parent.

The prediction therefore places the planned spatial outcome inside the same renderable representation used by perception. The ball expected beside the conifer is already \texttt{blue\_ball\_obj0}, at the plan's released pose, before it is detected again. This identity-preserving expectation is what lets later evidence refine the represented consequence of a specified intervention rather than rediscover an unrelated object.

\subsection{Reconcile Visual Consequences (Inferred from Evidence)}

A rigid object transform is geometrically informative but visually incomplete. It does not reveal what lay behind the ball, determine appearance from every camera, or confirm that the successor is consistent across views. \method therefore renders the predicted checkpoint at calibrated cameras, including full-scene, static-scene, old-object, and new-object RGB, alpha, and metric-depth components. These renders state what the world is expected to look like if the plan reaches its endpoint.

Registered RGB-D evidence enters through the same observation bundle whether it comes from physical sensors or a pretrained image-generation foundation model. Every view carries camera transforms, frame role, metric depth, masks, and provenance. Together these views form $B_{t+1}$: a registered account of the visual consequences that the rigid prediction alone cannot supply. 

The reported study uses geometry-constrained synthetic post-action observations from an image-generation model for this stage. Newly exposed source regions are inpainted, and a canonical object appearance is inserted inside the predicted target silhouette. The Gaussian prediction remains the authority for masks and metric depth; generated color is confined to explicitly marked pixels, and original RGB is restored exactly outside each edit mask. This provides a controlled observation target for exercising the complete prediction--reconciliation loop.

\subsection{Commit (Trusted for Reuse)}

Commit turns the reconciled evidence into the next Gaussian world. Change masks are lifted through calibrated depth into the 3D scene, newly visible background may introduce unowned seed Gaussians, and \method uses CL-Splats~\citep{ackermann2025clsplats} to optimize only the affected region while unrelated content remains frozen.

Planning and observation have complementary authority. The validated action endpoint determines the transformed object's geometry and opacity; observation refines appearance and the background revealed by its motion:
\begin{equation}
\underbrace{(\mu,s,q,\alpha)_{\mathrm{owned}}}_{\text{action-conditioned prediction}}
\quad\big|\quad
\underbrace{(c_{\mathrm{owned}},\gaussians_{\mathrm{disocc.}})}_{\text{observation-dependent update}}.
\label{eq:authority}
\end{equation}
After localized optimization, \method restores the predicted means, scales, orientations, and opacities of owned rows while retaining reconciled appearance. The resulting candidate preserves the planned object configuration while incorporating appearance and exposed background inferred from the registered targets. This division of responsibility keeps the geometric transition and the visual update consistent within one persistent world.

The locally updated candidate is committed only after its artifacts are intact, tensors remain finite and aligned, ownership is consistent, predicted object geometry is preserved, held-out changed-region error meets its criterion, and the increase in outside-region reconstruction error remains bounded. Every parent edge is also replayed in reverse so that promotion never destroys the worlds that preceded it.

Promotion grants the successor current authority, carrying the world forward. For the next instruction, \method does not patch the old semantic map or reuse the former planning scene. It renders new grounding views from $W_{t+1}$, associates detections near the predicted pose with the stable object identity, rebuilds task-progression semantics, constructs a new task-relevant twin and collision cloud, and binds the next plan to those products. The prior trajectory endpoint becomes the next robot start. In this way, the terminal world of one instruction becomes the base representation from which the next instruction is understood.

\section{Following the World Across Two Instructions}
\label{sec:evaluation}
We evaluate \method through a study over a captured real scene that follows one blue ball across two spatially and temporally constrained mobile-manipulation instructions. The sequence traces the same progression used to explain \method: construct task views from a persistent state, produce a plan, evolve the state from that plan, and reconstruct the views needed for what comes next. The initial world $W_0$ is a 6,125,086-row Gaussian splat reconstruction of a captured outdoor garden with calibrated source cameras~\citep{barron2022mipnerf360}.

\subsection{Timestep 0: Build the First Task from \texorpdfstring{$W_0$}{W0}}

Instruction 1 asks: ``Get the blue ball from under the table and leave it beside the potted lime-green conifer.'' Rendered views of $W_0$ ground the ball (254 rows), table, and conifer; their consolidated 3D records populate the semantic task state. The $W_0$ digital twin and collision cloud support the navigation, pick, and release plan, while Gaussian views provide several perspectives from which that plan can be inspected. Language grounding and temporal-logic planning follow LIMP~\citep{quartey2025verifiably}; Appendix~\ref{app:tasks} gives the full specification.

The plan contains a 323-frame pick trajectory and a 249-frame release trajectory. Its validated terminal pose places \texttt{blue\_ball\_obj0} beside the conifer. \method applies that pose to 254 Gaussian rows owned by the grounded ball to produce the predicted successor $\widehat W_1$.

\subsection{From the First Plan to the Next World}

To exercise the visual half of the transition, the study uses registered, geometry-constrained synthetic post-action RGB-D. Newly exposed source regions are inpainted and a canonical object appearance is inserted inside the predicted target silhouette. The Gaussian prediction remains authoritative for masks and metric depth, generated color is confined to marked pixels, and original RGB is restored outside each edit mask. Stage 1 contains 12 source-disocclusion and 9 target-insertion edits. Reconciliation produces a candidate world, and promotion makes the validated result $W_1$ current.

At this point the first instruction is over, but the representational process is not. $W_1$---not $W_0$---is now the source from which later grounding, semantics, geometry, and plans are constructed.

\subsection{Timestep 1: Rebuild the Second Task from \texorpdfstring{$W_1$}{W1}}

Instruction 2 asks: ``Stop by the small garden ornament, collect the blue ball beside the conifer, and put it on the table.'' \method renders new views from $W_1$, re-associates the detected ball with \texttt{blue\_ball\_obj0}, and rebuilds the semantic registry, digital twin, and collision cloud around its evolved pose. The new plan begins at instruction 1's trajectory endpoint, contains a 443-frame pick and a 308-frame place trajectory, and predicts the ball on the table in $\widehat W_2$. Stage 2 reconciliation uses 12 source-disocclusion and 12 target-insertion edits before the successor is promoted as $W_2$.

The trajectories are planned and simulated directly in the representation rather than executed on hardware. Validation checks kinematics, continuity, collision and contact policy, attachment, release, and artifact integrity.

\subsection{What the Sequence Must Preserve}

We examine four links in this progression. \textbf{Grounding-to-action support} asks whether ownership attaches the semantic ball to its own 3D content rather than nearby background. \textbf{Local world evolution} measures moved and inactive rows, inverse recovery, and action-owned drift. \textbf{Visual reconciliation} measures held-out L1 inside and outside the changed region. \textbf{Instruction-to-instruction continuity} asks whether the stable identity, grounding, checkpoint, twin, planning cloud, and robot start for instruction 2 all derive from promoted $W_1$.

\section{Results}
\label{sec:realm:results}

\subsection{Grounding Becomes Action Support Through Ownership}

\begin{figure*}[t]
  \centering
  \includegraphics[width=\textwidth]{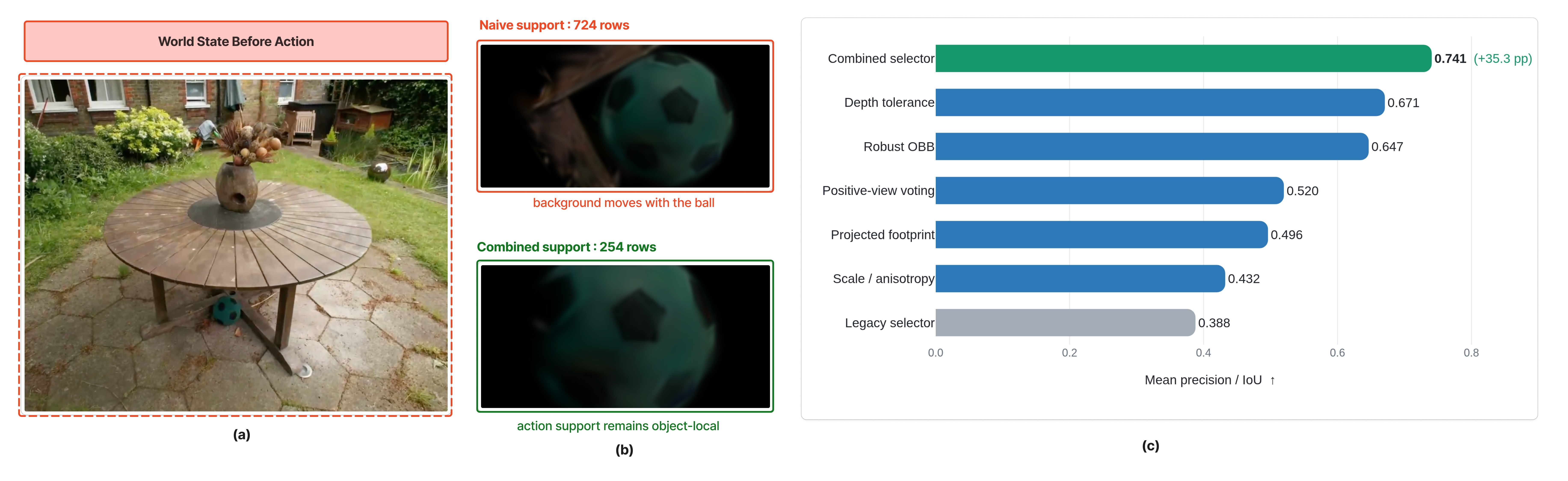}
  \caption[Ownership defines what the action moves]{\textbf{Ownership defines what the action moves.} The legacy 724-row selector attaches background support to the ball. The combined 254-row selector remains object-local. Across the seven selector variants, the combined rule improves mask precision/mean IoU from 0.388 to 0.741 while preserving measured recall on the garden ball.}
  \label{fig:ownership}
\end{figure*}

Rendered views first identify the ball as a semantic referent; ownership must then turn that referent into valid action support inside the Gaussian world. The original positive-voting selector assigned 724 rows to the ball, including 470 centers outside its fitted oriented bounding box. These rows contain table and ground support that would move whenever the ball moves. The combined selector reduces the support to 254 rows and raises mask precision/mean IoU from 0.388 to 0.741 while retaining measured recall of 1.0 (Table~\ref{tab:ownership}). Figure~\ref{fig:ownership} shows the consequence: stable ownership makes the rigid transition object-local rather than transporting part of the scene with the object.

\begin{table}[t]
\centering
\caption{Ownership selector analysis for the garden ball.}
\label{tab:ownership}
\small
\begin{tabular}{@{}lrrr@{}}
\toprule
Selector & Rows & Prec./IoU & Recall \\
\midrule
Legacy positive voting & 724 & 0.387891 & 1.000 \\
Stricter positive voting & 521 & 0.519845 & 1.000 \\
Lower depth tolerance & 315 & 0.670623 & 1.000 \\
OBB filter & 281 & 0.646528 & 1.000 \\
Footprint filter & 648 & 0.496164 & 1.000 \\
Scale/anisotropy filter & 721 & 0.432294 & 1.000 \\
\textbf{Combined selector} & \textbf{254} & \textbf{0.740647} & \textbf{1.000} \\
\bottomrule
\end{tabular}
\end{table}

\subsection{Known, Inferred, and Trusted States Remain Local and Reversible}

Both predicted transitions move exactly the same 254 owned rows. At stage 1, all 6,124,832 inactive rows remain exactly equal to $W_0$. Reconciliation introduces 12,170 unowned disocclusion seeds; the next prediction then preserves all 6,137,002 inactive rows exactly. The action-owned means, scales, orientations, and opacities have zero drift after reconciliation because the update restores them to their predicted values. Stored inverse deltas recover every parent exactly.

Reconciliation improves agreement with synthetic RGB targets on views held out from fitting. Masked L1 decreases from 0.138485 to 0.121646 at stage 1, a 12.2\% reduction, and from 0.134815 to 0.131212 at stage 2, a 2.7\% reduction. Outside-region L1 increases by approximately $2.2\times10^{-5}$ and $2.0\times10^{-5}$, respectively, within the $10^{-4}$ promotion criterion (Table~\ref{tab:transitions}). These held-out views provide the visual criteria for promotion: the reconciled state improves agreement with the synthetic targets in the edited region while keeping the increase in outside-region reconstruction error within the prescribed tolerance.

\begin{table}[t]
\centering
\caption[Spatial and visual continuity across the two world transitions]{Spatial and visual continuity across the two world transitions. Row counts describe rigid prediction; image errors compare the predicted and reconciled renders against synthetic targets on validation views. Geometry and opacity drift is measured after restoring the action-owned parameters.}
\label{tab:transitions}
\small
\begin{tabular}{@{}lrr@{}}
\toprule
Quantity & Stage 1 & Stage 2 \\
\midrule
Owner rows transformed in prediction & 254 & 254 \\
Non-moved rows exact in prediction & 6,124,832 & 6,137,002 \\
Disocclusion seeds & 12,170 & 3,551 \\
Masked L1 before & 0.138485 & 0.134815 \\
Masked L1 after & \textbf{0.121646} & \textbf{0.131212} \\
Outside L1 before & 0.022277 & 0.029291 \\
Outside L1 after & 0.022299 & 0.029311 \\
Max owned geometry/opacity drift & 0 & 0 \\
\bottomrule
\end{tabular}
\end{table}

\begin{figure*}[t]
  \centering
  \includegraphics[width=\textwidth]{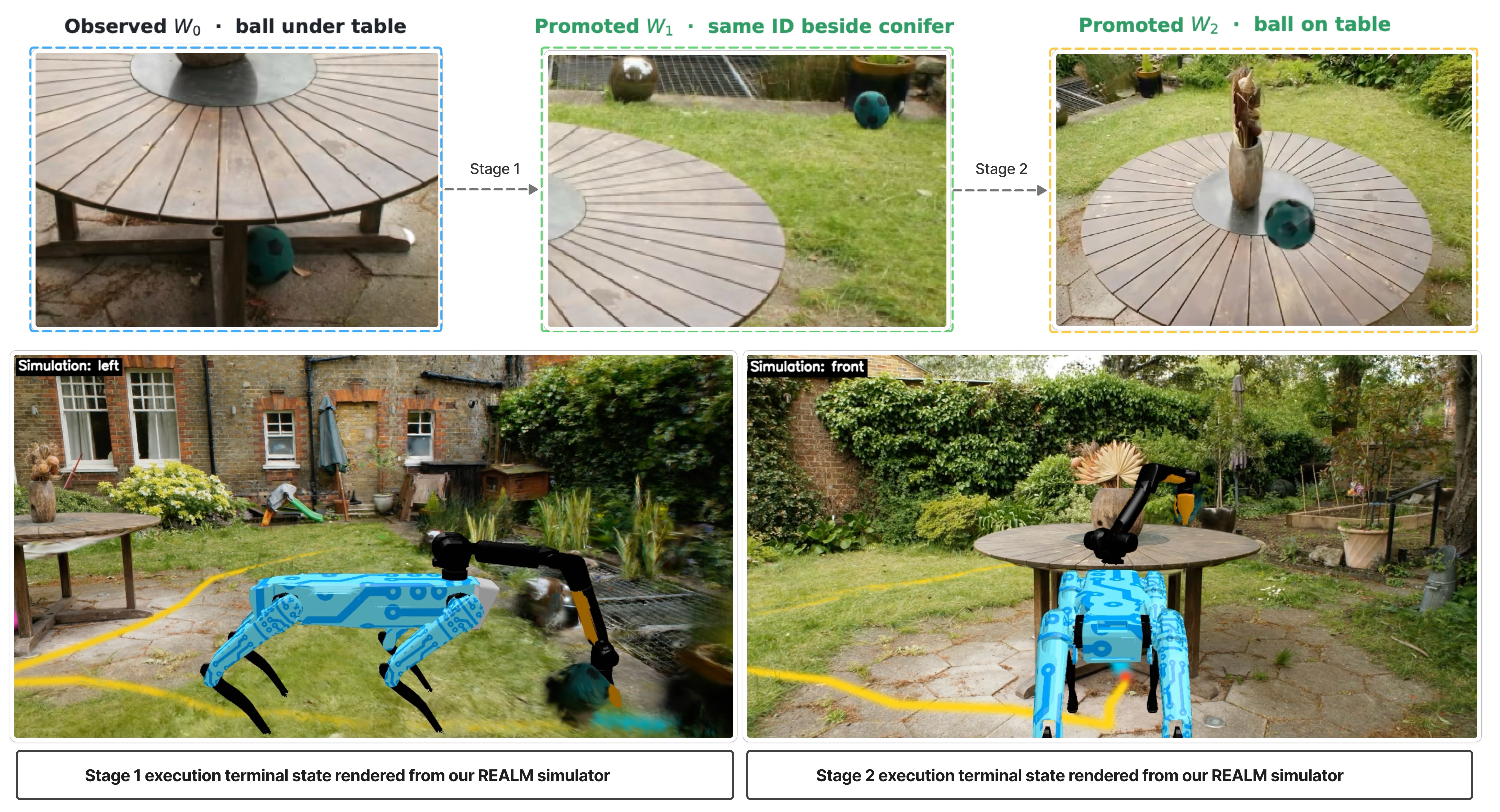}
  \caption[Sequential world evolution across two instructions]{\textbf{Sequential world evolution across two instructions.} The ball moves from its observed pose beneath the table to promoted $W_1$ beside the conifer and then to promoted $W_2$ on the table. Reconciliation improves held-out changed-region consistency at both stages. Each prediction moves the same 254 owned rows and preserves all other rows exactly; stored history recovers the ancestors. Fifteen detections across rendered views of $W_1$ associate to one stable object ID.}
  \label{fig:evidence}
\end{figure*}

\subsection{The Next Instruction Rebuilds Its World from \texorpdfstring{$W_1$}{W1}}

Promotion of $W_1$ triggers the same abstraction-building process that began at $W_0$, now over an evolved world. New rendered views produce 15 ball detections that associate with \texttt{blue\_ball\_obj0}. The selected pose lies 0.01879\,m from the action prediction and 3.7198\,m from the original under-table pose. From this $W_1$ pose, \method rebuilds the grounding record, semantic task state, digital twin, and planning cloud for instruction 2. The pick target is therefore beside the conifer, and the robot start is instruction 1's terminal state.

After instruction 2, the $W_2$ checkpoint, twin, semantic registry, and planning cloud agree on the ball's final table pose. The important continuity is not only that one coordinate changed. The persistent representation and every instruction-specific abstraction derived from it advance together: the object keeps its identity, grounding searches the updated location, planning receives geometry from that same revision, and history explains how the represented world arrived there.

\subsection{History Preserves How the Current World Was Reached}

The complete sequence contains five distinct states:
\begin{equation*}
W_0^{\mathrm{obs}}\rightarrow \widehat W_1\rightarrow \widetilde W_1^{\mathrm{syn}}
\rightarrow \widehat W_2\rightarrow \widetilde W_2^{\mathrm{syn}}.
\end{equation*}
Final validation walks all four parent edges, checks the state-bound artifacts and twin ownership, and reconstructs every ancestor exactly. The archived integration suite records 98/98 passing tests; a later targeted audit records 49/49 passing tests across state, ownership, reconciliation, re-grounding, and orchestration. The history makes prediction, observation, and promotion visible as different events in the robot's evolving belief.

\section{Discussion}
\label{sec:realm:discussion}

\subsection{From Verifiable Plans to an Evolving World}

LIMP establishes a path from expressive language to a grounded, verifiable plan. \method extends that path through time. The plan no longer terminates the representational process; it becomes an intervention that predicts a new world. That successor is reconciled, promoted, and used to interpret the next instruction. Together, the two systems connect an interpreted instruction to a planned intervention and to the world representation used for the next task.

The world interface concerns the premises of the planner's reasoning. Even when an instruction has been interpreted correctly, a plan grounded in an obsolete scene can concern the wrong location or geometry. Carrying the representation forward preserves the connection established by language grounding across successive tasks.

\subsection{One Representation, Four Connected Uses}

The Gaussian scene supports grounding, planning geometry, visual inspection, and action-conditioned updating through one persistent representation.
These uses depend on different information, but must agree about which world they describe.
The sequential study makes that dependency concrete: updating the ball's rendered appearance would be insufficient if grounding or collision checking continued to use its original location.

That distinction explains why regeneration matters. The semantic task state, digital twin, collision cloud, and plan views are useful precisely because they describe the current world. After promotion, rebuilding them from $W_{t+1}$ carries object identity and geometry forward together. Reusing any one of them from $W_t$ would reintroduce the stale-map failure through a different interface.

\subsection{From a Renderable Scene to a Spatial World Model}
The distinction developed in Section~\ref{sec:background:world-models} is now visible operationally.
A Gaussian render provides a view of a scene, whereas REALM's action-conditioned update produces a successor on which later grounding and planning can operate.
Its world-modeling role comes from representing task-relevant change while preserving object identity.

World models are often associated with learned latent dynamics or generated future video. \method instead models a task-relevant spatial consequence in an explicit 3D representation. The action transition is analytic; visual reconciliation can contain learned components; and the promoted state remains directly renderable and consumable by later grounding and motion planning. 

The known--inferred--trusted decomposition makes this hybrid construction explicit and is useful precisely because the planner already knows an important part of the future: where a successfully manipulated rigid object should be released. Prediction records what the plan says should change, reconciliation records what subsequent evidence adds, and promotion records which world later reasoning is allowed to consume.

\subsection{From an Evolving World to Composable Skills}

The garden study shows how the same object can remain available to later grounding and planning after its represented location changes.
Its evidence concerns continuity and local reconstruction under planned rigid-object motion.
Because the manipulation is simulated and the post-action imagery is synthetic, agreement with the updated scene does not establish successful physical execution.
In a physical deployment, new observations would need to correct predictions when a grasp fails, an object slips, or the environment changes independently of the robot.

An evolving representation still leaves the planner dependent on supplied skills.
Learning additional policies broadens what the robot can do, but composition also requires information about when those policies can be used and what conditions their actions are expected to produce.
Chapter~\ref{ch:pacts} develops this symbolic interface through joint learning of actions and predicted predicate trajectories.

\chapter{Learning Composable Skills}
\label{ch:pacts}

\begin{quote}
\small
This chapter is adapted from \textit{Jointly Learning Predicates and Actions Enables Zero-Shot Skill Composition}, coauthored by Benedict Quartey, Sebastian Castro, Eric Rosen, Wil Thomason, George Konidaris, and Stefanie Tellex~\citep{quartey2026pacts}.
The authorial ``we'' refers to this collaborative research.
\end{quote}
\vspace{0.75\baselineskip}

\noindent
A planner gains new possibilities when it can compose behaviors the robot has learned, rather than depend entirely on a supplied library of controllers.
The preceding chapters established how expressive instructions can become grounded plans and how the representation used for planning can evolve across tasks.
Both developments still depend on suitable skills being available.
Learning those skills from demonstrations offers a way to extend the robot's repertoire, but a policy that produces movements supplies only part of what composition requires.
The planner must also reason about the conditions those movements are expected to produce.

This chapter develops \emph{composable state-reporting skills}, addressing the action interface of trust.
PACTS learns a skill's control policy together with a symbolic trace of its expected effects, connecting continuous behavior to the predicates used in task planning.
The resulting interface supports new compositions of learned skills and replanning from updated predicate predictions.
The contribution complements LIMP's grounding of intent and REALM's evolving world representation by examining how learned behavior can participate in symbolic reasoning.
The experiments study this skill interface with a supplied predicate vocabulary and symbolic domain; their relationship to the preceding systems motivates the synthesis in Chapter~\ref{ch:scaling}.

\section{Learning Actions Together with Their Symbolic Outcomes}
\label{sec:pacts:introduction}

The generative policies introduced in Section~\ref{sec:background:skill-learning} can represent several ways to carry out a behavior.
For composition, however, plausible movements are insufficient: the symbolic conditions associated with a particular action sequence matter to the next skill.
Different ways to manipulate an object can lead to different outcomes, even when both begin from the same observation.
A policy and a separately sampled outcome predictor may therefore choose incompatible alternatives.
The distinction between their individual accuracy and their agreement under execution, developed in Section~\ref{sec:background:composition}, motivates learning the two trajectories together.

\begin{figure}
    \centering
    \includegraphics[width=1\columnwidth,trim={0cm 0cm 0cm 0cm},clip]{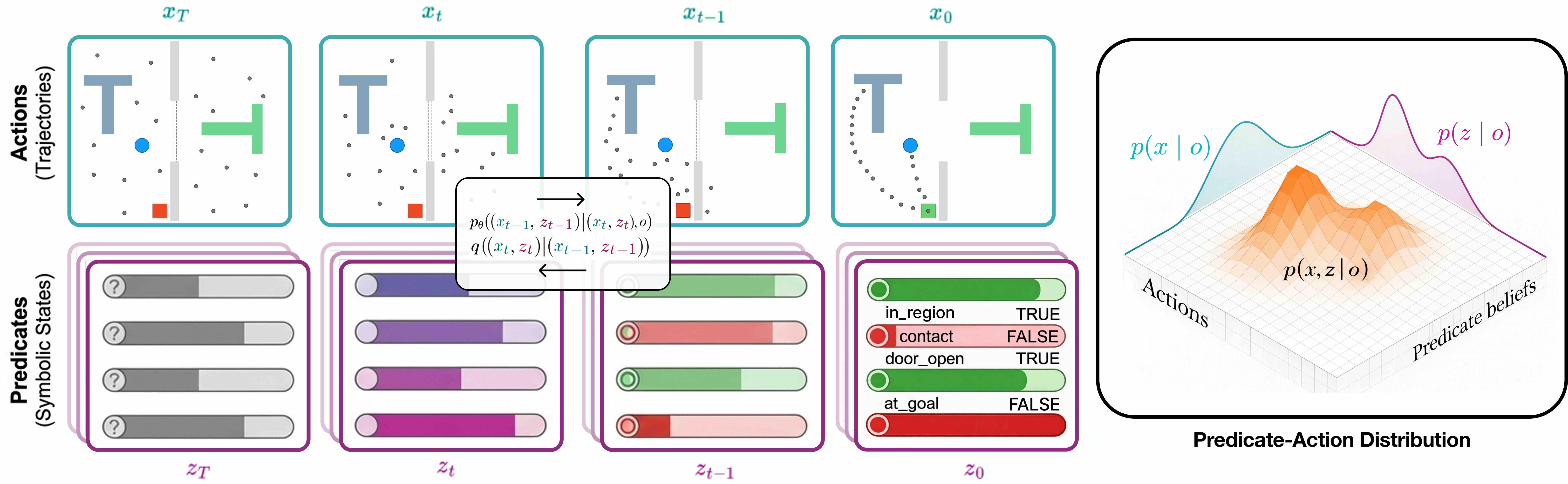}

    \caption[Predicate-Action Skills]{\textbf{Predicate-Action Skills}. Conditioned on current observations $\mathbf{o}$, we model skills as a joint generative process over an action trajectory $\mathbf{x}$ and a predicate-belief trajectory $\mathbf{z}$ by learning the coupled distribution $p(\mathbf{x},\mathbf{z}\mid \mathbf{o})$. Starting from noise ($\mathbf{x}_T$,$\mathbf{z}_T$), our model iteratively refines both modalities to produce paired action--outcome rollouts $(\mathbf{x}_0,\mathbf{z}_0)$.
    The resulting predicate-belief trajectory $\mathbf{z}_0$ provides an online symbolic interface for monitoring skill execution and planning-based skill composition using off-the-shelf planners.}
\label{fig:pacts:splash}
\end{figure}

\textbf{Predicate-Action Skills} (\approachname{}) represent a skill through a joint distribution over actions and predicted predicate values.
A sample contains both a control trajectory and the symbolic trace expected to accompany it (Figure~\ref{fig:pacts:splash}).
This formulation allows the model to represent dependence between the chosen action mode and its expected effects.
It also provides a practical interface through which a planner can assess progress, select subsequent skills, and reconsider a sequence as new observations condition later predictions.
Whether the predicted effects agree with execution is an empirical question, evaluated separately from predicate classification accuracy.

We instantiate this joint model with diffusion and flow-matching objectives and with UNet and Transformer architectures.
The evaluation examines action performance, predicate prediction, and action--predicate coherence on a compositional 2D benchmark and two 3D RoboMimic tasks~\citep{robomimic2021,mandlekar2023mimicgen}.
Planning-based composition is demonstrated in simulation and in a real-world cube-packing task.
The contributions are the joint generative skill formulation, a predicate-based interface for composition and monitoring, and a segmentation and labeling toolkit that prepares skill-specific demonstrations for joint learning.

\section{Related Work}
\label{sec:related_work}

\paragraph{Generative visuomotor policies for imitation learning}
Behavior cloning learns skills from demonstrated observation--action pairs~\citep{robomimic2021}, and recent work models \emph{distributions} over action trajectories to capture multiple ways to perform a behavior. Diffusion-based policies~\citep{chi2024diffusionpolicy} and conditional flow matching~\citep{chisari2024flowmatching} provide two such formulations. OpenVLA and $\pi_0$ connect pretrained vision-language representations to robot control~\citep{kim2024openvlaopensourcevisionlanguageactionmodel,black2024pi0}. Other approaches structure the learning problem through equivariance~\citep{wang2024equivariantdiffusionpolicy}, object-centric representations~\citep{yuan2022sornet}, or affordance-centered task frames~\citep{rana2024affordancecentricpolicylearningsample}. Diffusion has also been used for planning: Diffuser~\citep{janner2022planning} denoises state--action trajectories to enable flexible test-time conditioning, whereas Diffusion Policy~\citep{chi2024diffusionpolicy} treats control as learning the distribution over action sequences, conditioning on observations without denoising them for efficient real-time inference.
We adopt this conditional-policy perspective, but depart from action-only modeling by learning a coupled generator $p(\mathbf{x},\mathbf{z}\mid\mathbf{o})$ that jointly samples actions and a low-dimensional predicate-belief rollout $\mathbf{z}$ aligned with PDDL preconditions/effects, yielding an outcome-aware interface for execution-time monitoring and planning-based composition.
Vision--language--action models similarly couple control with language semantics, but their semantics are typically part of the conditioning interface rather than an explicit planner-consumable outcome trajectory.

\paragraph{Skills, abstraction, and planning-based composition}
The symbolic planning framework in Section~\ref{sec:background:symbolic-planning} requires both a physical implementation of each operator and a grounding for its predicates.
In robotic systems, these abstract actions are typically instantiated as hand-designed controllers, including statecharts~\citep{harel1987statecharts} and behavior trees~\citep{colledanchise2018bt}, and predicates are grounded through engineered perception and state estimation modules \citep{migimatsu2022symbolicstateestimationpredicates}. Several frameworks support such symbolic planning architectures \citep{diab2020skillman,martin2021plansys2,mayr2023skiros2,paulius2025bootstrapping, quartey2025verifiably}.

Recent approaches reduce this manual engineering by learning the components that connect behavior to symbolic planning. BLADE learns compositional behaviors and their preconditions and effects from demonstrations and language~\citep{liu2024BLADE}; Pixels-to-Predicates uses pretrained vision--language models to ground predicates for symbolic world-model learning~\citep{athalye2026pixels}. These approaches establish symbolic interfaces for planning, but do not jointly sample an action trajectory and its associated predicate trajectory. In long-horizon manipulation, actions and symbolic outcomes are tightly coupled: different action \emph{modes} correspond to different predicate transitions, and treating them as separate predictions can lead to inference-time inconsistency. \approachname{} addresses this coupling directly by modeling a skill as a \emph{joint} generative process over action trajectories and predicate-belief trajectories. Joint sampling allows the model to represent the dependence between action modes and symbolic outcomes, which independently sampled heads do not explicitly represent. \looseness=-1

\section{Jointly Modeling Predicates and Actions}
\label{sec:pacts:method}

\begin{figure*}   
  \centering
  \includegraphics[width=1\textwidth]{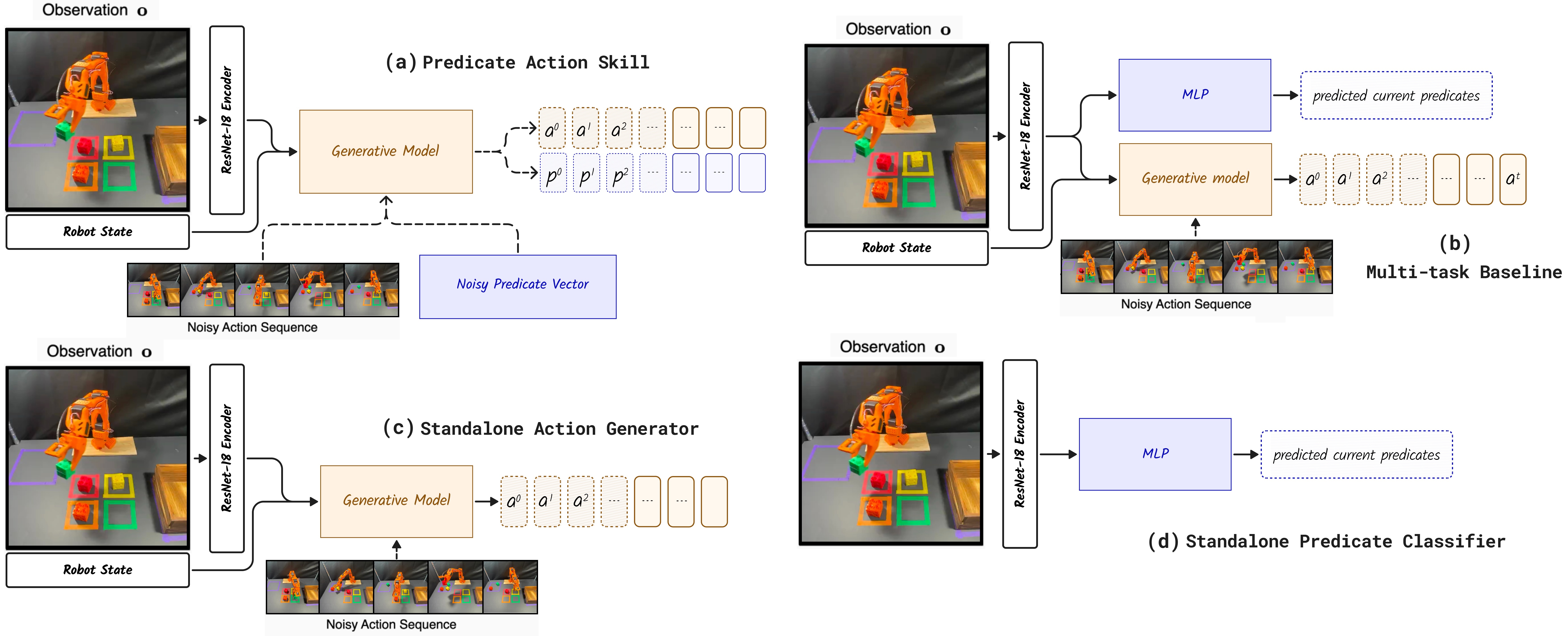}

\caption[Architectures and baselines]{\textbf{Architectures and baselines.}
\textbf{(a)} \approachname{} uses a \emph{single} conditional generative model to sample \emph{joint} action--predicate rollouts, refined together via DDPM denoising or flow matching.
To \emph{disentangle} joint modeling from multi-task learning, we include a multi-task baseline \textbf{(b)} that shares the same observation encoder but predicts predicates and actions with \emph{separate} heads trained with a combined loss. Standalone baselines isolate each component: \textbf{(c)} a generative \emph{action-only} generator that generates actions, and \textbf{(d)} a \emph{predicate-only} model predicts predicates from observations.}
\label{fig:architectures}
\end{figure*}

\subsection{Problem definition: joint action--outcome skill model}

Using the trajectory conventions from Sections~\ref{sec:background:skill-learning} and~\ref{sec:background:composition}, we model a closed-loop skill as a conditional distribution over a horizon-$H$ action trajectory
$\mathbf{x}=(\mathbf{a}_t,\dots,\mathbf{a}_{t+H-1})\in\mathbb{R}^{H\times d_a}$, where each $\mathbf{a}_{t+h}\in\mathbb{R}^{d_a}$ is a single control action. To support planning-based skill composition, we additionally model the symbolic outcomes induced by actions using a
\emph{predicate-belief trajectory}
$\mathbf{z}=(\mathbf{z}_{t+1},\dots,\mathbf{z}_{t+H})\in[0,1]^{H\times J}$,
where $\mathbf{z}_{t+h,j}$ is a soft prediction that predicate $j\in\{1,\dots,J\}$ will hold at future step $t+h$. These values are called predicate beliefs in this chapter; their range does not establish probabilistic calibration.

Given the current observation $\mathbf{o}_t$, \approachname{} learns the coupled distribution $p_\theta(\mathbf{x},\mathbf{z}\mid \mathbf{o}_t)$ so that actions and symbolic outcomes form one generated sample. We implement this by defining the joint trajectory variable $\mathbf{y} \triangleq [\mathbf{x}\,|\,\mathbf{z}] \in \mathbb{R}^{H\times(d_a+J)}$ and learning a single conditional generative model $p_\theta(\mathbf{y}\mid \mathbf{o}_t)$. This joint modeling choice is the key distinction from decoupled or multi-task baselines: $\mathbf{z}$ is not produced by an independent classifier head, but is part of the generated trajectory sample.

\paragraph{Predicate-belief representation.}
In the dataset, predicates are provided as binary labels in $\{0,1\}^J$.
We treat predicate channels as continuous during generation by linearly normalizing labels to $[-1,1]$ and learning to generate them jointly with actions. At inference, the model maps the generated predicate channels back to $[0,1]$, yielding soft beliefs that can be thresholded when discrete predicate evaluations are required (e.g., checking preconditions/effects during planning).

\subsection{DDPM formulation}
\label{sec:ddpm_formulation}
The DDPM construction in Section~\ref{sec:background:diffusion} is applied here to the joint variable $\mathbf{y}=[\mathbf{x}\,|\,\mathbf{z}]$.
Following Diffusion Policy~\citep{chi2024diffusionpolicy}, the generator conditions on $\mathbf{o}_t$ without generating future images or a full physical state trajectory.
It does predict future symbolic conditions through the predicate channels.
Let $q(\mathbf{y}^k\mid \mathbf{y}^{k-1})$ denote the forward noising process with a standard noise schedule
$\{\alpha_k\}_{k=1}^{K}$. We train a noise-prediction network $\epsilon_\theta$, conditioned on $\mathbf{o}_t$, to predict the injected noise:
\begin{equation}
\mathcal{L}_{\textsc{ddpm}}
=
\mathbb{E}_{\mathbf{o}_t,\mathbf{y}^0,k,\epsilon}
\left[
\left\|
\epsilon_\theta(\mathbf{o}_t,\mathbf{y}^k,k)-\epsilon
\right\|^2
\right],
\end{equation}
where $\mathbf{y}^k$ is obtained by noising the clean joint trajectory $\mathbf{y}^0=[\mathbf{x}\,|\,\mathbf{z}]$ at \emph{denoising step} $k$.
At inference, the model samples $\mathbf{y}^K\sim\mathcal{N}(0,I)$ and iteratively denoises it to obtain $\mathbf{y}^0$.

\subsection{Conditional Flow Matching formulation}
\label{sec:flow_matching}
The joint object $\mathbf{y}=[\mathbf{x}\,|\,\mathbf{z}]$ can also be learned via \emph{conditional flow matching} (CFM)~\citep{tong2023improving,lipman2022flow,black2024pi0}, which targets the same conditional distribution $p_\theta(\mathbf{y}\mid \mathbf{o}_t)$ but uses a different objective and sampler.
Let $\mathbf{y}^1\sim\mathcal{N}(0,I)$ and use the linear interpolation from Section~\ref{sec:background:flow-matching} between data $\mathbf{y}^0$ and noise $\mathbf{y}^1$ as
\begin{equation}
\mathbf{y}_\tau = \tau\,\mathbf{y}^1 + (1-\tau)\,\mathbf{y}^0,\qquad \tau\in[0,1],
\end{equation}
with constant target velocity $\mathbf{u}_\tau=\frac{d}{d\tau}\mathbf{y}_\tau=\mathbf{y}^1-\mathbf{y}^0$.
A conditional vector field $v_\theta(\mathbf{o}_t,\mathbf{y}_\tau,\tau)$ is trained by regression:
\begin{equation}
\mathcal{L}_{\textsc{cfm}}
=
\mathbb{E}_{\mathbf{o}_t,\mathbf{y}^0,\mathbf{y}^1,\tau}
\left[
\left\|
v_\theta(\mathbf{o}_t,\mathbf{y}_\tau,\tau) - (\mathbf{y}^1-\mathbf{y}^0)
\right\|^2
\right].
\end{equation}
At inference, samples are obtained by integrating from noise ($\tau{=}1$) to data ($\tau{=}0$), yielding $\mathbf{y}^0$.

DDPM and flow matching instantiate the same modeling choice---learning the coupled action--outcome distribution
$p_\theta(\mathbf{x},\mathbf{z}\mid\mathbf{o}_t)$---but provide complementary training lenses with different sampling trade-offs.

\subsection{Architecture}
\label{sec:architectures}
As shown in Fig.~\ref{fig:architectures}(a), \approachname{} factorizes into (i) an observation encoder and (ii) a conditional trajectory generator. The observation encoder maps the current observation $\mathbf{o}_t$ into a latent embedding used to condition the generator. The generator implements either a DDPM noise-prediction network $\epsilon_\theta(\mathbf{o}_t,\mathbf{y}^k,k)$ (Sec.~\ref{sec:ddpm_formulation}) or a CFM velocity field $v_\theta(\mathbf{o}_t,\mathbf{y}_\tau,\tau)$ (Sec.~\ref{sec:flow_matching}), operating on the joint trajectory $\mathbf{y}=[\mathbf{x}\,|\,\mathbf{z}]$.

We instantiate the trajectory generator with two backbones:
(i) a temporal UNet that applies 1D convolutions over the horizon dimension, and
(ii) a Transformer that treats time steps as tokens and models long-range temporal dependencies via self-attention.
In both cases, conditioning enters through (a) the observation embedding and (b) the diffusion/flow time parameter (discrete $k$ for DDPM, continuous $\tau$ for CFM), injected via standard conditional modulation.

Like prior trajectory-generating visuomotor policies, \approachname{} can be executed in a receding-horizon manner: at each control cycle, the policy conditions on the current observation $\mathbf{o}_t$, samples a length-$H$ action chunk $\mathbf{x}$ (and its paired predicate-belief trajectory $\mathbf{z}$), executes a short prefix of $\mathbf{x}$, then re-observes and re-samples.

\begin{algorithm}[t]
\caption{Planning with Predicted Predicate Beliefs}
\label{alg:predicate_belief_planning}
\begin{algorithmic}[1]
\REQUIRE Skill library $\mathcal{S}$ with \approachname{} $\{p_\theta(\mathbf{x},\mathbf{z}\mid \mathbf{o}, s)\}_{s\in\mathcal{S}}$; planning domain $\mathcal{D}$; PDDL problem instance $\mathcal{P}$ with initial predicate state $\hat{\mathbf{p}}_0$ and goal $g$
\STATE $t \leftarrow 0$; \ $\hat{\mathbf{p}}_t \leftarrow \hat{\mathbf{p}}_0$; \ observe $\mathbf{o}_t$
\WHILE{goal $g$ not satisfied under $\hat{\mathbf{p}}_t$}
    \STATE $\pi \leftarrow \textsc{Plan}(\mathcal{D}, \hat{\mathbf{p}}_t, g)$ \COMMENT{symbolic plan in predicate space}
    \STATE $s \leftarrow \pi[0]$ \COMMENT{next planned skill}
    \WHILE{\textbf{not} \textsc{EffectsSatisfied}$(s,\hat{\mathbf{p}}_t)$ \textbf{and} \textbf{not} \textsc{TerminateSkill}$(s)$}
        \STATE Sample rollout $(\mathbf{x},\mathbf{z}) \sim p_\theta(\mathbf{x},\mathbf{z}\mid \mathbf{o}_t,s)$
        \STATE Execute first action from $\mathbf{x}$ and observe $\mathbf{o}_{t+1}$
        \STATE $\hat{\mathbf{p}}_{t+1} \leftarrow \textsc{Thresh}(\mathbf{z}_{t+1})$ \COMMENT{predicted next predicate state}
        \STATE $t \leftarrow t+1$
    \ENDWHILE
\ENDWHILE
\end{algorithmic}
\end{algorithm}

\section{Planning-based Skill Composition with Predicate Beliefs}
\label{sec:planning}
A key benefit of modeling skills as joint action--outcome rollouts is that the generated predicate-belief trajectory provides a symbolic interface for composition. This interface is compatible with formal robot planning pipelines that represent tasks using symbolic or temporal-logic structure~\citep{garrett_integrated_2021,paulius2025bootstrapping,quartey2025verifiably}. We assume a PDDL symbolic planning domain specifying skill-level operators with preconditions and effects over predicates, and a PDDL problem instance specifying an initial predicate state and a goal. Given a library of learned skills, an off-the-shelf planner synthesizes a sequence of skills to achieve a goal.
During execution, \approachname{} produces online predicate-belief estimates that serve as the planner state, enabling the system to (i) check whether a skill's intended effects are satisfied and (ii) replan from an updated symbolic state.

\paragraph{Predicate beliefs as an execution interface}
At each decision point, the robot observes $\mathbf{o}_t$ and (for the selected skill $s$) samples a joint rollout $(\mathbf{x},\mathbf{z})\sim p_\theta(\mathbf{x},\mathbf{z}\mid \mathbf{o}_t,s)$.
The belief trajectory $\mathbf{z}\in[0,1]^{H\times J}$ provides a compact prediction of how symbolic conditions are expected to evolve as the skill unfolds.
The system obtains a discrete predicate valuation by thresholding beliefs (e.g., $\mathbf{z}_{t+h,j}>0.5$), and uses the predicted next-step predicates $\hat{\mathbf{p}}_{t+1}=\textsc{Thresh}(\mathbf{z}_{t+1})$ as the symbolic state for effect checking and replanning.
Although $\hat{\mathbf{p}}_{t+1}$ is obtained from the model's prediction $\mathbf{z}_{t+1}$ (made when sampling at time $t$, before observing $\mathbf{o}_{t+1}$), beliefs are re-grounded at every step because the next rollout is conditioned on the latest observation $\mathbf{o}_{t+1}$.

\paragraph{Plan--execute--check--replan}
Algorithm~\ref{alg:predicate_belief_planning} summarizes our composition loop.
The planner replans after each executed skill, using the updated predicted predicate state.
It selects the next skill from this symbolic state, and execution proceeds until the predicted predicates satisfy the skill's effect conditions or another termination condition is met.

\subsection{Skill segmentation and labeling toolkit.}
To make real-world deployment tractable, we introduce a skill segmentation and labeling toolkit that converts monolithic demonstrations into training-ready, skill-centric datasets for joint $(\mathbf{x},\mathbf{z})$ learning.
The toolkit operates natively on LeRobot \citep{cadene2024lerobot} episodic datasets (state/action logs with video observations) and provides three core capabilities:
(i) sparse keyframe annotation of grounded predicates defined directly from the PDDL domain/problem files,
(ii) automatic conversion of sparse annotations into dense per-timestep predicate labels via forward-fill interpolation, and
(iii) skill-level segmentation by identifying goal-achieving predicate configurations and backward-filling executing phases to extract contiguous per-skill segments.

\section{Evaluation}

\begin{figure}
\centering
   \includegraphics[width=1\textwidth]{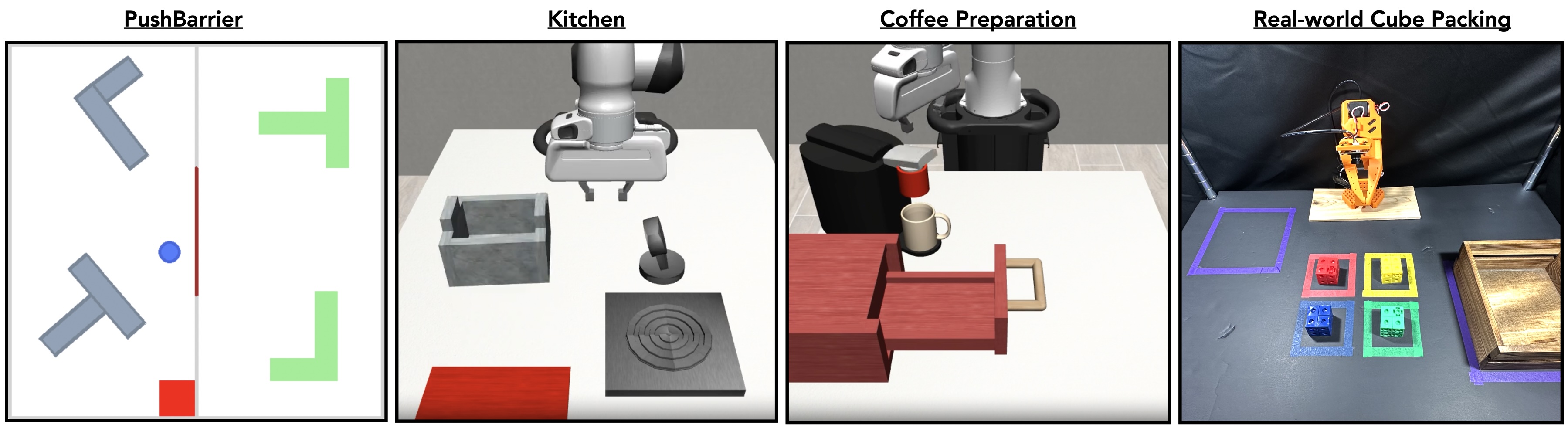}
\caption[Evaluation environments]{\textbf{Evaluation environments.}
We evaluate \approachname{} across (a) \textbf{PushBarrier}, a controlled 2D compositional environment for systematic ablations and planner-based skill sequencing; (b) \textbf{Kitchen} and (c) \textbf{Coffee Preparation} from RoboMimic/MimicGen, which test joint action--predicate modeling under realistic 3D visual manipulation with simulator predicate oracles; and (d) a \textbf{real-world cube packing} task, where a robot packs four colored cubes into a container.}
  \label{fig:environment}
\end{figure} 

We evaluate \approachname{} across three tasks in simulation and a real-world multi-step task. Our experiments are designed to answer three questions:
\begin{enumerate}
    \item How does joint predicate--action generation affect action performance and predicate classification accuracy?
    \item How do modeling choices--DDPM denoising versus conditional flow matching, and UNet versus Transformer backbones--impact performance?
    \item Can the learned predicate-belief interface support planning-based recomposition and replanning-based recovery on novel goals without retraining?
\end{enumerate}

\subsection{Tasks and datasets}
\label{sec:tasks_datasets}

We evaluate \approachname{} in three settings that trade off control, realism, and compositional complexity (see Fig. \ref{fig:environment}).

\vspace{2mm}

\noindent\textbf{2D PushBarrier (compositional benchmark).}
We introduce a multi-step 2D manipulation task adapted from the Push-T environment \citep{chi2024diffusionpolicy}.
The workspace contains two objects (a \texttt{T} block and an \texttt{L} block) that must each be pushed from their respective start regions to corresponding goal regions.
Successful completion requires not only reaching the goal regions but also achieving the correct final \emph{orientation/alignment} of each object with its goal marker.
A barrier separates the start and goal regions, and the robot must first press a button to open the barrier before either object can be transferred to the goal side.
This environment provides a controlled setting with clear compositional structure and is our primary benchmark for systematic evaluations and ablations.

We use PushBarrier in two complementary evaluation modes. 
To address Questions (1) and (2) in an end-to-end setting, we train \emph{single} long-horizon policies on the full task and compare task performance and predicate classification accuracy against our baselines.
To address Question (3) in a compositional setting, we construct a library of short-horizon skills and a task-relevant predicate set, segment demonstrations into per-skill datasets, and train a separate \approachname{} model for each skill. We then evaluate planning-based compositions that require sequencing skills, including \emph{novel skill sequences} not observed during training.

\vspace{2mm}

\noindent\textbf{RoboMimic (3D manipulation).}
To evaluate under realistic visual and geometric complexity, we additionally evaluate on two RoboMimic \citep{robomimic2021} tasks with MimicGen \citep{mandlekar2023mimicgen} demonstration data:
\textbf{Kitchen} and \textbf{Coffee Preparation}.
These tasks involve long-horizon manipulation under visually grounded observations and serve as a complementary test of whether joint modeling improves or maintains performance beyond the controlled 2D benchmark.
For each task, we define a predicate set aligned with task structure and implement predicate oracles inside the simulator that compute ground-truth predicate values from simulator state.
These predicate labels are used for both training and evaluation, enabling direct comparisons of predicate classification accuracy across methods. We use the base-difficulty (\texttt{d0}) variants, converted to absolute action space, with 200 demonstrations per task: \texttt{kitchen\_d0} and \texttt{coffee\_preparation\_d0}.

\vspace{2mm}

\noindent\textbf{Real-world demonstration}
We include a qualitative real-world demonstration on a cube-packing task. A tabletop robot must pick and place four colored cubes from fixed start regions into a container. We collect \emph{monolithic}, long-horizon demonstrations that complete the full task (packing all four cubes), then use our segmentation and labeling toolkit to decompose each episode into skill-centric sub-demonstrations corresponding to per-cube packing behaviors (e.g., \texttt{pack\_red}, \texttt{pack\_green}, etc.). We train a separate \approachname{} model for each subskill and compose them with an off-the-shelf PDDL planner. This enables recompositions beyond the original demonstrations, such as executing individual packing skills in isolation or packing only a specified subset of cubes (e.g., ``pack red and yellow''), providing a concrete test of zero-shot compositionality in the real world.

\subsection{Baselines and ablations}
\label{sec:baselines}

We design baselines to isolate the benefit of modeling the joint action--outcome distribution $p_\theta(\mathbf{x},\mathbf{z}\mid\mathbf{o}_t)$ and sampling paired action--predicate rollouts in \approachname{}. We begin with baselines that model each conditional in isolation (action-only and predicate-only). We then include a shared-encoder multi-task baseline that predicts actions and predicates with separate heads trained end-to-end, which isolates the effect of \emph{joint sampling} from standard multi-task representation sharing.

\paragraph{Action-only generative policy (Fig.~\ref{fig:architectures}(c)).}
A diffusion/flow policy that models only $p(\mathbf{x}\mid\mathbf{o}_t)$ by generating an action trajectory $\mathbf{x}$ without predicate channels. This represents the standard action-only generative-policy baseline and tests whether adding outcome modeling degrades (or improves) action quality.

\paragraph{Standalone predicate predictor (Fig.~\ref{fig:architectures}(d)).}
A predicate-only model that predicts predicate beliefs from observations, modeling $p(\mathbf{z}\mid\mathbf{o}_t)$. This quantifies predicate prediction performance in isolation and mirrors common pipelines where predicates are treated purely as a discriminative output.

\paragraph{Multi-task baseline (Fig.~\ref{fig:architectures}(b)).}
To disentangle joint modeling from multi-task learning, we include a baseline with a shared observation encoder and two output heads trained end-to-end with a combined loss.
This corresponds to learning a factorized conditional model
\begin{equation}
p(\mathbf{x},\mathbf{z}\mid \mathbf{o}_t) \approx p(\mathbf{x}\mid \mathbf{o}_t)\,p(\mathbf{z}\mid \mathbf{o}_t),
\end{equation}
where the action head is a conditional diffusion/flow policy trained to model $p(\mathbf{x}\mid\mathbf{o}_t)$ (using the same DDPM/CFM objectives as Sec.~\ref{sec:ddpm_formulation}--\ref{sec:flow_matching}),
and the predicate head is a discriminative predictor trained with binary cross-entropy on ground-truth predicates.
The combined objective is
\begin{equation}
\mathcal{L}_{\text{total}}=\mathcal{L}_{\text{action}}+\mathcal{L}_{\text{predicate}}.
\end{equation}
Although the encoder is shared, the two heads produce outputs independently at inference, so this baseline cannot encourage action--predicate consistency at sampling time.

\vspace{2mm}

\noindent\textbf{Objective and backbone sweeps.}
All methods (\approachname{} and baselines) are evaluated across both training objectives (DDPM and CFM) and both backbone choices (UNet and Transformer), so differences can be attributed to joint modeling rather than a particular generative objective or architecture.

\subsection{Metrics}
\label{sec:metrics}

We report the following metrics.
\paragraph{Policy evaluation}
We evaluate methods by rolling out policies in the environment and reporting task performance.
In PushBarrier we report \emph{Max Reward} ($\uparrow$) and \emph{Steps to Max Reward} ($\downarrow$). In RoboMimic, we follow the benchmark protocol and report rollout success/reward statistics. Results are mean $\pm$ standard error over seven seeded rollouts.

\paragraph{Predicate classification evaluation}
We compare predicted predicates to ground-truth oracle labels and report macro \emph{F1}, \emph{recall}, \emph{precision}, and \emph{accuracy}.  For belief outputs $\mathbf{z}\in[0,1]^{H\times J}$, we threshold at $0.5$ to obtain $\hat{\mathbf{p}}\in\{0,1\}^J$.  Predicate metrics are \emph{macro-averaged across predicates} so that each predicate contributes equally, preventing frequent predicates from dominating the score.

\paragraph{Predicate--action coherence}
To test whether predicted outcomes match the outcomes induced by executed actions, we compute horizon-indexed coherence curves over open-loop action chunks.
At each prediction cycle $\tau$, the policy predicts $\hat{\mathbf{z}}_{\tau+1:\tau+H}$ and executes its generated action chunk; we then compare $\hat{\mathbf{z}}_{\tau+h}$ to oracle predicates $\mathbf{p}_{\tau+h}$ for $h\in\{1,\dots,H\}$.
We report \textsc{CrossCons}$(h)$ (thresholded agreement, $\uparrow$), \textsc{CrossBrier}$(h)$ (MSE on beliefs, $\downarrow$), and \textsc{CrossNLL}$(h)$ (BCE on beliefs, $\downarrow$), along with majority/base-rate reference baselines.\looseness=-1

\begin{table*}[t]
\centering
\caption[Single-policy rollout evaluation on PushBarrier]{\textbf{Single-policy rollout evaluation on PushBarrier.}
We compare \approachname{} to action-only, predicate-only, and shared-encoder multi-task baselines when learning a \emph{single} policy for the full task. This isolates whether joint action--outcome modeling degrades action performance and whether it improves predicate prediction in the absence of any skill segmentation or planning. Policy metrics are mean$\pm$standard error over seven seeded rollouts}
\label{tab:rollout-eval-pushtlbarrier_blocked}
\resizebox{\textwidth}{!}{
\begin{tabular}{@{}llcccccc@{}}
\toprule
\textbf{Setting} & \textbf{Approach} &
\multicolumn{1}{c}{\makecell{Policy Evaluation}} &
\multicolumn{5}{c}{\makecell{Predicate Classification Evaluation}} \\
\cmidrule(lr){3-3} \cmidrule(lr){4-8}
(Objective + Backbone)& &
Max Reward $\uparrow$ &
F1 $\uparrow$ & Recall $\uparrow$ & Precision $\uparrow$ & Accuracy $\uparrow$ & Observations \\
\midrule
\midrule

\multirow{3}{*}{\makecell[l]{DDPM + UNet}}
&Action-Only Generator + Predicate-Only Predictor & 0.83 $\pm$ 0.09 & \SI{0.9476}{} & \SI{0.9547}{} & \SI{0.9471}{} & \SI{79.40}{\%} & 2403 \\
& Multi-task Baseline (Shared Encoder)  & 0.77 $\pm$ 0.13 & \SI{0.9764}{} & \SI{0.9712}{} & \SI{0.9834}{} & \SI{89.39}{\%} & 2705 \\
& \textbf{Predicate Action Skill (\approachname{})}            & \textbf{1.00 $\pm$ 0.00} & \textbf{\SI{0.9885}{}} & \textbf{\SI{0.9924}{}} & \textbf{\SI{0.9849}{}} & \textbf{\SI{93.78}{\%}} & 2234 \\
\midrule

\multirow{3}{*}{\makecell[l]{DDPM + Transformer}}
& Action-Only Generator + Predicate-Only Predictor & 0.45 $\pm$ 0.10 & \SI{0.9364}{} & \SI{0.9355}{} & \SI{0.9561}{} & \SI{84.18}{\%} & 2800 \\
& Multi-task Baseline (Shared Encoder)  & 0.33 $\pm$ 0.08 & \SI{0.8684}{} & \SI{0.8705}{} & \SI{0.8679}{} & \SI{85.43}{\%} & 2800 \\
& \textbf{Predicate Action Skill (\approachname{})}             & \textbf{0.58 $\pm$ 0.15} & \textbf{\SI{0.9872}{}} & \textbf{\SI{0.9925}{}} & \textbf{\SI{0.9822}{}} & \textbf{\SI{91.98}{\%}} & 2606 \\
\midrule

\multirow{3}{*}{\makecell[l]{CFM + UNet}}
& Action-Only Generator + Predicate-Only Predictor & 0.31 $\pm$ 0.11 & \SI{0.8297}{} & \SI{0.7985}{} & \SI{0.8775}{} & \SI{82.11}{\%} & 2800 \\
& Multi-task Baseline (Shared Encoder)  & 0.45 $\pm$ 0.11 & \SI{0.9551}{} & \SI{0.9595}{} & \SI{0.9559}{} & \SI{84.88}{\%} & 2784 \\
& \textbf{Predicate Action Skill (\approachname{})}            & \textbf{0.56 $\pm$ 0.12} & \textbf{\SI{0.9819}{}} & \textbf{\SI{0.9945}{}} & \textbf{\SI{0.9712}{}} & \textbf{\SI{92.18}{\%}} & 2800 \\
\midrule

\multirow{3}{*}{\makecell[l]{CFM + Transformer}}
& Action-Only Generator + Predicate-Only Predictor & 0.62 $\pm$ 0.12 & \SI{0.8969}{} & \SI{0.9194}{} & \SI{0.8869}{} & \SI{75.15}{\%} & 2612 \\
& Multi-task Baseline (Shared Encoder)  & 0.45 $\pm$ 0.12 & \SI{0.8914}{} & \SI{0.9387}{} & \SI{0.8573}{} & \SI{81.43}{\%} & 2800 \\
& \textbf{\textbf{Predicate Action Skill (\approachname{})}}            & \textbf{0.72 $\pm$ 0.12} & \textbf{\SI{0.9813}{}} & \textbf{\SI{0.9932}{}} & \textbf{\SI{0.9706}{}} & \textbf{\SI{92.21}{\%}} & 2759 \\

\bottomrule
\bottomrule
\end{tabular}
}
\end{table*}

\begin{table*}[t]
\centering
\caption[Single-policy rollout evaluation on RoboMimic]{\textbf{Single-policy rollout evaluation on RoboMimic (Kitchen\_d0 + Coffee\_d0).}
We evaluate \approachname{} and baselines on two RoboMimic manipulation tasks under realistic visual and geometric complexity, training a \emph{single} policy per task (no planning or composition). Results are averaged across Kitchen and Coffee Preparation, weighting each environment equally. Policy metrics report mean$\pm$SE over seven seeded rollouts, with standard errors propagated across environments. Predicate classification metrics are macro-averaged across predicates within each environment, then averaged across environments.}
\label{tab:rollout-eval-robomimic_blocked}
\resizebox{\textwidth}{!}{
\begin{tabular}{@{}llcccccc@{}}
\toprule
\textbf{Setting} & \textbf{Approach} &
\multicolumn{1}{c}{\makecell{Policy}} &
\multicolumn{5}{c}{\makecell{Predicate Classification}} \\
\cmidrule(lr){3-3} \cmidrule(lr){4-8}
(Objective + Backbone)& &
Max Reward $\uparrow$ &
F1 $\uparrow$ & Recall $\uparrow$ & Precision $\uparrow$ & Accuracy $\uparrow$ & Observations \\
\midrule
\midrule
\multirow{2}{*}{\makecell[l]{DDPM + UNet}}
& Action-Only Generator + Predicate-Only Predictor& 0.93 $\pm$ 0.02 & \SI{0.9194}{} & \SI{0.9095}{} & \SI{0.9479}{} & \SI{83.98}{\%} & 10476 \\
& \textbf{Predicate Action Skill (\approachname{})}               & \textbf{0.96 $\pm$ 0.02} & \textbf{\SI{0.9340}{}} & \textbf{\SI{0.9271}{}} & \textbf{\SI{0.9537}{}} & \textbf{\SI{90.74}{\%}} & 10246 \\
\midrule

\multirow{2}{*}{\makecell[l]{DDPM + Transformer}}
& \textbf{Action-Only Generator + Predicate-Only Predictor} & \textbf{0.97 $\pm$ 0.02} & \textbf{\SI{0.9451}{}} & \textbf{\SI{0.9360}{}} & \textbf{\SI{0.9643}{}} & \textbf{\SI{88.36}{\%}} & 10087 \\
& Predicate Action Skill (\approachname{})              & 0.93 $\pm$ 0.04 & \SI{0.9185}{} & \SI{0.9050}{} & \SI{0.9531}{} & \SI{88.03}{\%} & 10103 \\
\midrule

\multirow{2}{*}{\makecell[l]{CFM + UNet}}
& \textbf{Action-Only Generator + Predicate-Only Predictor} & \textbf{0.99 $\pm$ 0.01} & \textbf{\SI{0.9467}{}} & \textbf{\SI{0.9407}{}} & \textbf{\SI{0.9574}{}} & \textbf{\SI{88.26}{\%}} & 10243 \\
& Predicate Action Skill (\approachname{})              & 0.91 $\pm$ 0.03 & \SI{0.8884}{} & \SI{0.8824}{} & \SI{0.9296}{} & \SI{82.47}{\%} & 10543 \\
\midrule

\multirow{2}{*}{\makecell[l]{CFM + Transformer}}
& Action-Only Generator + Predicate-Only Predictor& \textbf{0.97 $\pm$ 0.02} & \SI{0.9324}{} & \SI{0.9206}{} & \textbf{\SI{0.9612}{}} & \SI{86.07}{\%} & 9927 \\
& Predicate Action Skill (\approachname{})              & 0.89 $\pm$ 0.03 & \textbf{\SI{0.9386}{}} & \textbf{\SI{0.9287}{}} & \SI{0.9587}{} & \textbf{\SI{89.78}{\%}} & 10369 \\

\bottomrule
\bottomrule
\end{tabular}
}
\end{table*}

\section{Results and Discussion}
\label{sec:pacts:results}
\begin{figure*}   
  \centering
  \includegraphics[width=1\textwidth]{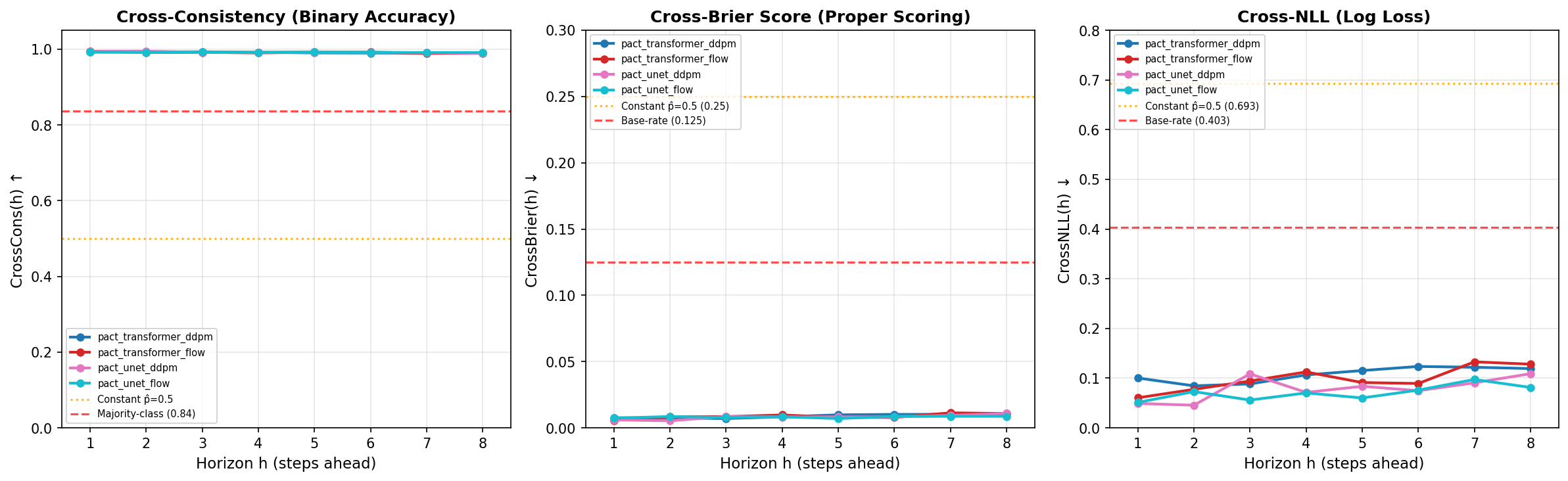}
\caption[Predicate--action coherence of predicted belief rollouts]{\textbf{Predicate--action coherence of predicted belief rollouts.}
We measure how well \approachname{}'s predicted predicate-belief rollout matches the predicate outcomes realized after executing its open-loop action chunks.  The x-axis reports lookahead $h$ steps into the chunk. We plot (left) \textsc{CrossCons}$(h)$ (thresholded agreement; higher is better), (middle) \textsc{CrossBrier}$(h)$ (Brier score; lower is better), and (right) \textsc{CrossNLL}$(h)$ (log loss; lower is better). Horizontal lines are dataset-statistics baselines: constant $\hat p{=}0.5$ (dotted) and a base-rate/majority predictor from label frequencies (dashed). These baselines ignore both observations and planned actions; outperforming them shows that the predicted rollouts carry information beyond fixed label frequencies. Across objectives and backbones, \approachname{} maintains high agreement and low losses over the horizon.}
\label{fig:coherence}
\end{figure*}

\begin{figure*}   
  \centering
  \includegraphics[width=1\textwidth]{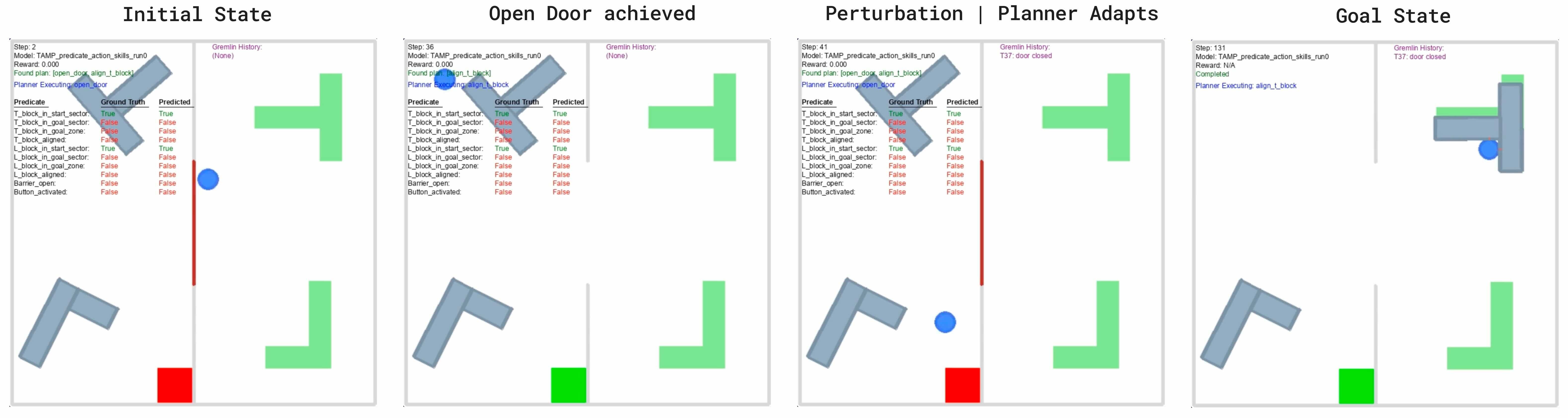}
\caption[Skill recomposition and adversarial recovery in PushBarrier]{\textbf{Skill recomposition and adversarial recovery in PushBarrier.}
Left to right: starting from the initial state, the planner synthesizes a sequence for a \emph{novel} subgoal (\texttt{open\_door} $\rightarrow$ \texttt{align\_t\_block}). After \texttt{open\_door} succeeds, an adversarial perturbation (\emph{Gremlin}) closes the door, which invalidates the current plan while aligning the $T$ block. Using online predicate-belief estimates to check effects, the system triggers replanning, re-invokes \texttt{open\_door}, and then resumes \texttt{align\_t\_block} to reach the goal.}
\label{fig:planning}
\end{figure*}

\noindent\textbf{Single-policy analysis.}
We first study \approachname{} in the simplest setting: training a \emph{single} long-horizon policy per task. This isolates the impact of jointly generating action--outcome rollouts on policy performance and predicate prediction independent of planning, or composition.
The study tests whether joint learning can \emph{add a planner-aligned outcome interface while preserving closed-loop control} and obtain an outcome trace that is (i) coherent under execution and (ii) suitable for downstream composition.

\paragraph{PushBarrier (2D compositional benchmark).}
Table~\ref{tab:rollout-eval-pushtlbarrier_blocked} reports single-policy performance and predicate classification on PushBarrier across objectives and backbones. Across all settings, \approachname{} achieves strong task reward while consistently improving predicate prediction and policy performance relative to baselines. Notably, the shared-encoder multi-task baseline often attains strong predicate scores, but its factorized heads yield weaker control; in contrast, \approachname{} samples \emph{paired} action--predicate rollouts from $p_\theta(\mathbf{x},\mathbf{z}\mid \mathbf{o}_t)$ and remains competitive (often best) on reward. This supports our primary modeling claim: adding outcome channels does \emph{not} require trading off control quality for symbolic prediction.
Instead, learning the coupled object $p_\theta(\mathbf{x},\mathbf{z}\mid \mathbf{o}_t)$ provides a useful inductive bias, encouraging the policy to represent \emph{which} action modes correspond to \emph{which} symbolic outcomes rather than learning actions and predicates as loosely related outputs.

Objective/backbone choices shift absolute numbers but not the conclusion: joint modeling remains strong across DDPM/CFM and UNet/Transformer. Consistent with prior work, UNet-style backbones are a reliable default, whereas Transformer variants can be more tuning-sensitive~\citep{chi2024diffusionpolicy}; in our evaluation, all models are trained for the same number of epochs without per-setting hyperparameter sweeps. We therefore view DDPM vs.\ CFM and UNet vs.\ Transformer primarily as implementation trade-offs (e.g., sampling speed and tuning sensitivity), with the primary differentiator being the \emph{joint} modeling target. \looseness=-1

\paragraph{RoboMimic (Kitchen\_d0 + Coffee\_d0).}
Table~\ref{tab:rollout-eval-robomimic_blocked} reports single-policy evaluation averaged across both RoboMimic tasks.
The action-only baseline remains strong on return across several objective/backbone choices, underscoring that explicit outcome modeling is not strictly necessary to achieve high reward when training end-to-end on a fixed task distribution.
Our objective, however, is planning-based composition, which requires a \emph{planner-aligned outcome interface} that stays coupled to the policy’s sampled actions and remains meaningful under execution---without sacrificing closed-loop control in realistic 3D manipulation.
In this setting, \approachname{} remains competitive in return and frequently improves predicate quality, with the strongest overall trade-off in the DDPM+UNet configuration.

\paragraph{Predicate--action coherence.}
An outcome interface is only useful for composition if it remains \emph{coupled} to what the policy actually does.
Per-step predicate metrics alone (Tables~\ref{tab:rollout-eval-pushtlbarrier_blocked}--\ref{tab:rollout-eval-robomimic_blocked}) do not test whether predicted outcomes match what the model’s \emph{own} generated actions induce.
Figure~\ref{fig:coherence} therefore evaluates predicate--action coherence: whether predicted belief rollouts agree with oracle predicate outcomes \emph{under the model’s executed actions} across lookahead horizons.
We report \textsc{CrossCons}$(h)$ and two proper scoring rules, \textsc{CrossBrier}$(h)$ and \textsc{CrossNLL}$(h)$, alongside constant and base-rate/majority reference baselines computed from dataset label statistics (i.e., priors).
Across objectives and backbones, \approachname{} achieves near-perfect \textsc{CrossCons} and substantially lower \textsc{CrossBrier}/\textsc{CrossNLL} than these reference baselines across the full horizon, indicating that predicted belief rollouts track outcomes realized under execution rather than reflecting label priors.
This provides empirical support for using $\mathbf{z}$ as an outcome trace associated with the sampled action chunk: its predictions agree with predicate outcomes under execution.

\noindent\textbf{Planning-based composition and recovery.}
We demonstrate planning-based recomposition of learned skills and replanning-based recovery under perturbations using the predicate-belief interface. Figure~\ref{fig:planning} shows a PushBarrier run with two key properties.
First, the goal is \emph{novel}: it specifies only a subgoal of the full demonstrated behavior (e.g., achieve \texttt{align\_t\_block} without completing the entire monolithic task).
Second, execution is perturbed adversarially: the door is closed mid-execution during \texttt{align\_t\_block}.
By checking symbolic effects using updated predicate beliefs, the system detects that the current plan is no longer applicable, triggers replanning, and recomposes a corrected sequence (\texttt{open\_door} $\rightarrow$ \texttt{align\_t\_block}) to reach the goal.

This illustrates two benefits of joint action--outcome skills:
(i) \emph{Recomposition}: skills expose planner-aligned predicates, allowing an off-the-shelf planner to synthesize sequences for goals not demonstrated end-to-end;
(ii) \emph{Recovery}: predicate beliefs make failures legible at the symbolic level, enabling replanning-based correction without retraining.

\section{Limitations}

\approachname{} inherits several limitations from predicate and skill learning.
First, performance depends on the choice and coverage of the predicate set: incomplete or mis-specified predicates that are misaligned with task structure can make planning brittle and limit the expressivity of the outcome interface.
Second, while joint modeling couples action and outcome prediction, it does not eliminate errors from perception and state aliasing---especially under occlusions and long horizons---and execution can still fail under stochasticity, partial observability, or distribution shift.
Finally, our composition results assume a manually defined symbolic domain (e.g., PDDL operators and goals); automatically discovering predicates, operators, or abstractions remains an open research direction.
A promising avenue is to leverage pretrained vision--language models to propose and evaluate candidate predicates directly from demonstrations \citep{athalye2026pixels}, then distill a compact planner-compatible set for downstream planning.

\section{From Composable Skills to Scaling Trusted Autonomy}
\label{sec:pacts:conclusion}

PACTS connect learned control to the symbolic conditions through which a planner composes behavior.
Joint generation makes the action trajectory and its predicted predicate trace parts of the same skill model.
The evaluations test control performance, agreement between predicted and executed outcomes, and composition for goals without end-to-end demonstrations of the requested sequence.
The contribution is a way to acquire control and state reporting together within an explicit symbolic planning domain.

Together, the technical chapters connect grounded intent, an evolving world representation, and learned behavior to symbolic reasoning.
The concluding Chapter~\ref{ch:scaling} develops their shared principle of constructing and maintaining task-relevant abstractions, and examines how these interfaces can support continued learning, observation, and recovery in a physical closed-loop system.
\par


\chapter{Scaling Trusted Autonomy}
\label{ch:scaling}
\label{ch:conclusion}
The ambition of generally capable robotics is to make useful behavior possible without anticipating every task or engineering every response in advance.
As the range of requests, environments, and learned abilities grows, so does the difficulty of establishing why a robot's behavior should be relied upon.
A system may understand more language while leaving its interpretation difficult to examine, acquire more skills while obscuring their effects, or construct a detailed scene whose contents become outdated through its own actions.
The central challenge is to develop the means of assessing behavior alongside the capability to produce it.

This thesis has approached that challenge through the connections between learned models and explicit reasoning.
The preceding chapters develop grounded task specifications, persistent spatial representations, and learned skills with symbolic state reports.
Together, they support a broader position: the structure required for planning need not all be fixed before the robot encounters a task.
Some of that structure can be constructed from language and observations when it becomes relevant; some can be learned with behavior; and some must be revised as the world changes.
Understanding how to construct and maintain these connections is central to scaling trusted autonomy.

\section{From Greater Capability to Trusted Autonomy}
\label{sec:conclusion:capability}

The introduction established two requirements for robots that operate in human environments.
People should be able to specify useful goals without prescribing each movement, and they should have a basis for assessing whether the resulting behavior satisfies their requests.
Foundation models make a much broader range of language and perceptual knowledge available to the robot.
Symbolic planning provides a means of expressing requirements and reasoning over combinations of actions.
Their complementary strengths become useful together when the robot can connect the particular request and environment to the representations on which planning depends.

The contributions develop this connection along three dimensions of capability.
LIMP expands the instructions that can be grounded into explicit temporal and spatial requirements without a complete prebuilt semantic map.
REALM carries a persistent spatial state across planned changes, allowing subsequent instructions to obtain their grounding and planning abstractions from an evolved world.
PACTS connect learned control to predicted symbolic conditions, enabling a planner to compose trained skills for requested goals without demonstrations of the entire sequence.
These are different forms of expansion: across descriptions, across successive world states, and across combinations of behavior.
They make the question of scaling concrete without equating it with a measured scaling law or unrestricted competence.

Each expansion also exposes something that the robot must be able to assess.
An expressive interpretation must identify the intended task and referents.
A skill sequence must have effects that support the planner's next decision.
A persistent representation must distinguish a predicted change from one supported by observation.
The scientific value of the three interfaces lies in making these dependencies explicit.
They identify what connects a plan that is correct within its model to a task that is correctly understood and carried out in the world.

This perspective changes where the burden of engineering lies.
A planning system needs more than an algorithm for searching a supplied domain; it needs a way to obtain and maintain the domain information that makes its reasoning applicable.
The thesis develops parts of that process through learning, grounding, and world evolution.
Its unifying question is therefore how a robot can acquire the structure needed for a particular decision while retaining enough information to reconsider that structure when circumstances change.

\section{Constructing the Abstractions a Task Requires}
\label{sec:conclusion:abstraction}

The world contains far more detail than a planner can use directly.
Even a familiar room admits many descriptions: objects have appearances, spatial relationships, possible uses, and properties that matter only for particular activities.
An abstraction makes planning tractable by retaining distinctions relevant to a decision and setting others aside.
The difficulty is deciding those distinctions before knowing what the robot will be asked to do.
A representation sufficient for one instruction may omit the very attribute or relationship needed to understand the next.

The approach developed in this thesis is to defer task-specific representational commitments until the task provides a reason to make them.
The robot retains observations or a richer spatial representation from which it can construct the semantic and geometric descriptions required for grounding and planning.
The resulting model is bounded and explicit, but it need not be the only model that the robot can construct from the available information.
This is the sense in which \emph{delaying closure} offers a path toward planning in open-world settings: the robot constructs a model for the current task without treating it as a complete description of everything it might later need to know.

LIMP makes this principle concrete at the interface between language and the mapped environment.
An instruction determines which referent descriptions and spatial relationships the system resolves using retained observations.
The semantic map is consequently shaped by the task rather than limited to an exhaustive catalogue of named landmarks prepared beforehand.
Foundation models supply the breadth of language and visual concepts used in this construction, while explicit task structure and spatial reasoning connect their outputs to the planner.
The important combination is semantic flexibility with a representation whose interpretation and consequences can be examined.

REALM extends the principle through time.
If the underlying scene is fixed, newly constructed abstractions can still describe an obsolete world.
REALM therefore maintains a persistent Gaussian scene with object identity and an explicit history of action-conditioned changes.
Grounded objects, planning geometry, collision representations, and rendered views are derived from the state associated with the current task.
Their continuity comes from that common state, while their form depends on the question they must answer.
The predictive transitions and subsequent use for decisions give the scene its world-model function; the Gaussian representation alone does not provide it.

PACTS addresses a complementary source of rigidity: the connection between a learned controller and the symbolic conditions used to compose it.
Its action trajectories and predicate-belief trajectories are learned jointly, so the skill exposes predicted state in a form the planner can use.
The predicate vocabulary and operator model are supplied, and the constituent skills are trained before their composition.
PACTS therefore does not demonstrate runtime discovery of the planning domain.
It shows how part of the interface that would otherwise require separate engineering can be acquired with behavior, making those learned abilities available for new combinations at runtime.

Deferring these commitments does not remove the need for prior structure.
LIMP depends on captured observations, spatial representations, and available skills; REALM depends on a reconstructed scene and specified transition mechanisms; PACTS depends on demonstrations and a symbolic domain.
Nor does constructing a model at runtime change the logical semantics of the planner that consumes it.
A finite or fixed map is not inherently closed-world; that assumption enters when what it contains is treated as a complete account of the relevant world.
The broader contribution concerns when and how the robot obtains such an account, and how it can replace it as new tasks or evidence make different distinctions necessary.

This principle also reveals a further requirement.
Information is relevant to a task when it can affect the task's interpretation, feasibility, or outcome, whether or not the instruction mentions it.
The robot may need to account for an obstacle, an unstable support, or a property that becomes apparent only during interaction.
Retaining richer observations creates an opportunity to recover omitted information, but does not ensure that it was observed or can be recovered correctly.
Trusted runtime abstraction therefore requires a way to assess whether the constructed model is adequate for the decision it supports.
The trust interfaces make that requirement more precise by locating the connections through which an inadequate model can mislead the robot.

\clearpage
\section{Trust Across Intent, Action, and World}
\label{sec:conclusion:interfaces}

\begin{figure}[htbp]
    \centering
    \includegraphics[width=\textwidth]{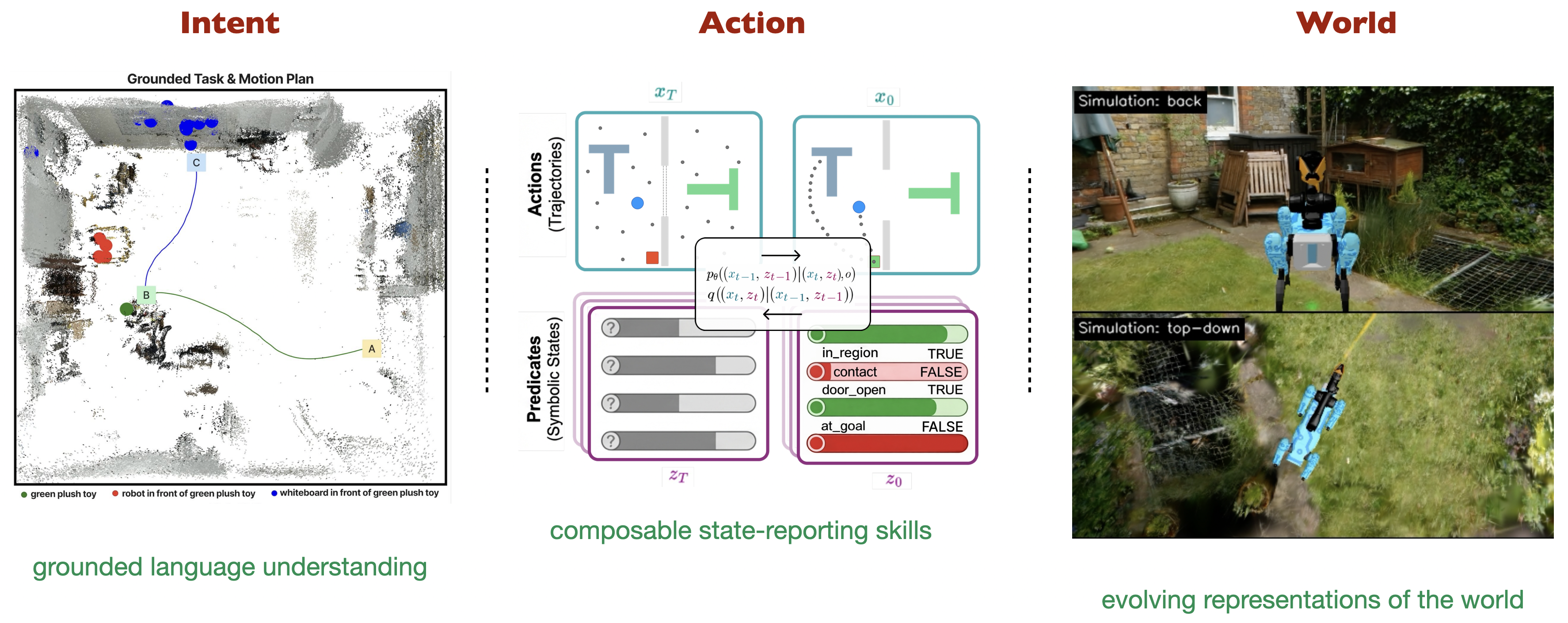}
    \caption[The three trust interfaces and their complementary contributions]{The three trust interfaces and their complementary contributions. LIMP exposes a grounded task and motion plan; PACTS jointly generates actions and predicted predicate trajectories; REALM supplies an evolving spatial representation for subsequent tasks.}
    \label{fig:conclusion:interfaces}
\end{figure}

The three questions introduced at the beginning of the dissertation can now be considered in light of the technical results.
Did the robot understand what the user meant?
Did its actions produce the intended effects?
Does its internal world still match reality?
Figure~\ref{fig:conclusion:interfaces} brings the corresponding contributions together in this conceptual order.
The chapter sequence followed the assumptions exposed by LIMP; returning to Intent, Action, and World shows how those contributions bear on the successive decisions involved in carrying out a request.

At the intent interface, assurance begins with making the interpreted request explicit.
LIMP represents temporal requirements in a co-safe task specification and connects its propositions to spatial referents and feasible motion.
Its evaluation across 150 instructions in five real-world environments tests the combination of language interpretation, grounding, and planning.
The results in Tables~\ref{tab:instruction-spatial} and~\ref{tab:temporal-tamp} show the benefit of this structure particularly for complex temporal requirements and avoidance constraints.
The weaker performance on instructions with implicit preferences is also informative: a formal specification can expose and preserve an interpretation without ensuring that it is the interpretation the user intended.

This distinction gives human inspection a clear role.
LIMP presents its interpreted task structure for review and supports reprompting after a user's correction; the reported translation evaluation measures the initial output without that correction.
Inspection offers a means of identifying errors before the planner acts on them.
It is distinct from checking whether a plan satisfies the resulting formal specification.
The latter establishes a relationship between a model and a plan, conditional on the interpretation, grounding, and action assumptions that made the model meaningful.

At the action interface, a planner needs to know how execution bears on its symbolic state.
PACTS learns to generate a proposed action trajectory together with the predicate conditions expected along it.
The composition and perturbation examples show how this interface supports new sequences of trained skills and replanning when updated beliefs invalidate a plan.
The action--predicate evaluation in Chapter~\ref{ch:pacts} goes beyond measuring predicate classification alone: it compares predicted beliefs with outcomes under executed action chunks.
This directly tests the connection that matters for composition, within the evaluated tasks and horizons.

The same distinction between representation and evidence applies here.
Joint sampling makes actions and predicate predictions parts of one learned model, but the architecture does not prove that they agree with physical outcomes.
The reported state is refreshed through observation-conditioned rollouts, yet predicted future predicates are not independent measurements of successful execution.
The contribution is an interface through which learned behavior can participate in symbolic reasoning and through which its predicted effects can be evaluated.
Reliable use depends on how well those predictions track the behavior and observations encountered during operation.

At the world interface, the question extends beyond the current skill to the state on which later tasks will depend.
REALM's two-instruction study shows that a persistent representation can carry object identity and planned change forward while regenerating the next task's grounding and planning geometry.
Object-local transitions preserve unaffected Gaussian content, and the recorded history permits exact recovery of ancestor states.
The representation and image-consistency results in Section~\ref{sec:realm:results} establish specific properties of that update process.

That boundary is essential to the world interface's role in continued autonomy.
A planned endpoint predicts where an object should be if the action succeeds.
A consistent rendering shows what the resulting representation implies.
An observation supplies evidence about what the robot actually changed.
Keeping these roles distinct allows an evolving world model to support scrutiny rather than merely accumulate plausible predictions.
REALM provides an explicit structure for carrying changes and their provenance forward; physical observations would have to test those changes during deployment.

Taken together, the results provide different kinds of evidence at complementary interfaces.
Specification satisfaction, empirical agreement between predicted and executed outcomes, and consistency of world-state updates answer different questions.
Their value comes from connecting them without allowing one to stand in for another.
A plan can satisfy the wrong specification, accurate skill-level predictions can be grounded in the wrong object, and an internally consistent world can depart from reality.
The thesis makes these dependencies available for examination and develops mechanisms for several of them.
Establishing assurance across their composition is the next scientific step.

\begin{samepage}
\section{Toward Continued Autonomy}
\label{sec:conclusion:research}

The contributions suggest a robot whose understanding, behavior, and world representation can develop together over successive tasks.
An instruction would determine the abstractions needed for planning; learned skills would expose the conditions relevant to their execution; and subsequent observations would inform the state used for the next request.
Realizing that loop poses questions about how the components should share evidence, recognize inadequacy, and change their decisions.
The following directions arise from those questions and from the assumptions of the demonstrated systems.
\par
\end{samepage}

\subsection{Recognizing When an Abstraction Is Insufficient}

Runtime construction makes it possible to build different models for different tasks.
The next challenge is to determine when the chosen model leaves out something consequential.
Suppose an instruction identifies a book to be moved from a table to a sofa.
Correctly identifying those objects may be sufficient for understanding the request, while planning and execution also depend on objects obstructing the grasp or the stability of the intended placement.
A description adequate at the intent interface may therefore be insufficient at the action interface.
The robot needs a way to discover and incorporate distinctions as they become relevant across these uses.

One research direction is to connect planning and execution failures to targeted expansion of the abstraction.
Repeatedly failing to find a feasible grasp could prompt the robot to inspect nearby geometry; an unexpected effect could reveal a property absent from its action model.
Foundation models could propose additional descriptions or relationships, while geometric checks and observations would test whether those proposals explain the difficulty.
The key question is how to select an informative revision without rebuilding an exhaustive model of the environment.
Success would involve both improved task decisions and a justified account of why the new information was required.

This question also applies to the meaning of absent information.
Failure to observe a relevant object or condition should be distinguished from evidence that it is absent.
A system using bounded classical planning models could manage that distinction around the planner, acquiring information or revising the domain before making a consequential commitment.
Evaluating this ability requires tasks with deliberately omitted but relevant entities, hidden conditions, and changed instructions.
Such tests would measure whether the robot recognizes the limits of its abstraction, rather than only whether it solves tasks already expressible within it.

\subsection{Connecting Predictions to Observations}

Continued autonomy depends on deciding which predicted changes deserve to become part of the robot's working account of the world.
The intended effect of a skill provides a useful expectation: it identifies what might have moved and where new evidence may be needed.
It also creates a risk if the system promotes that expectation without checking it.
A failed grasp, slip, or unmodeled interaction could otherwise affect every later grounding and planning decision.
The action and world interfaces must therefore connect predicted effects to observations that can confirm or contradict them.

PACTS and REALM provide complementary starting points for this connection.
Predicate beliefs describe changes in conditions used by the planner, while a persistent spatial representation records objects and their geometric relationships.
An integrated system could use a reported effect to direct re-observation and then compare the observed scene with the predicted transition.
Agreement would support reuse of the updated state; disagreement would require reconsidering the skill outcome, the object association, or the world update.
The challenge is to relate symbolic and geometric evidence without assuming that either is an infallible authority.

Observation should itself become a purposeful decision.
The robot need not reconstruct its entire environment after every action, but it needs evidence about changes on which later decisions depend.
Selecting viewpoints, inspecting an occluded object, or asking a person to resolve an ambiguous referent can each reduce a different source of uncertainty.
An account of the value of such information should consider both its cost and the consequences of acting without it.
This would extend runtime abstraction from deciding what to represent to deciding what must be observed before that representation can be relied upon.

\subsection{Expanding Skills and Their Symbolic Meaning}

Composition extends the usefulness of existing skills, but its reach is bounded by the abilities and conditions available to the planner.
PACTS assumes a supplied predicate vocabulary and symbolic operators.
Extending this work toward continued learning requires a way to introduce new behaviors together with the descriptions that make them usable alongside existing skills.
A controller alone cannot tell the planner when the skill applies, which conditions it may establish, or how those conditions relate to the rest of the domain.

Demonstrations and interactions could provide evidence for learning these relationships.
A pretrained model might propose candidate predicates or describe a skill's intended effect, but those descriptions would need to be grounded and tested against execution.
Different controllers that claim to establish the same condition should support a consistent interpretation of that condition.
Conversely, persistent failures may indicate that an operator omits a prerequisite or that one predicate conflates states requiring different behavior.
Learning the interface would therefore involve revising symbolic distinctions as well as fitting better action and state predictions.

The central evaluation question is whether each addition expands the planner's useful choices while preserving the meaning of existing ones.
This calls for testing both new compositions and previously achievable tasks after the domain changes.
It also requires relating predicate predictions to the actions actually executed, especially when new skills alter observation distributions or introduce unfamiliar interactions.
PACTS's coherence evaluation provides one starting point for this assessment; broader coverage, confidence calibration, and validation under distribution shift would be needed to support sustained expansion.

\subsection{Managing Uncertainty Across the Interfaces}

Uncertainty has different meanings at different points in the autonomy loop.
The robot may be unsure which object a phrase refers to, whether a skill will establish a condition, or whether a scene update reflects a physical change.
A single confidence score can obscure those differences.
For uncertainty to inform behavior, the system needs to associate it with a particular proposition or prediction and identify what decision depends on it.

The interfaces also share sources of error.
A mistaken object association can influence the task specification, skill input, and world update simultaneously.
Agreement among these components may then reflect their common mistake.
Multiplying independently reported confidences would be unjustified without accounting for such dependencies, and a predicate-belief score is not automatically a calibrated probability.
The research problem is to preserve enough information about the origin and meaning of uncertainty to determine which additional evidence could resolve it.

Such an account would make recovery more selective.
Ambiguity in intent may call for clarification, uncertainty about an effect may call for observation, and an infeasible plan may call for a different skill or abstraction.
The robot should also be able to recognize when none of its available actions has adequate support and seek assistance.
Evaluations should examine whether these decisions improve task outcomes and reduce unsupported commitments, alongside the time, observations, and human intervention they require.
This would make assurance part of the robot's decision process rather than a judgment applied only after execution.

\section{Concluding Perspective}
\label{sec:conclusion:perspective}

Generally capable robots will encounter tasks and circumstances that cannot all be described in advance.
The work in this dissertation shows how planning can draw on learned semantic knowledge, persistent spatial representations, and learned behavior to address parts of that variability.
Its central position is that \emph{scaling trusted autonomy requires unifying grounded language understanding, composable state-reporting skills, and evolving representations of the world}.
Each contribution makes explicit a connection needed to interpret a request, reason about behavior, or carry useful state into the next task.

The broader lesson concerns how that structure is obtained.
By deferring task-specific commitments, constructing relevant abstractions when they are needed, and maintaining the information from which they are derived, a robot can extend the reach of explicit planning beyond a single world description fixed in advance.
Learning can help supply the missing semantics and behavior; reasoning can make their consequences available for inspection and testing.
Continued autonomy will depend on carrying this relationship further, so that the robot can recognize when its abstractions are inadequate and use experience to revise them.
Progress toward generally capable, verifiable robots requires both a growing range of useful behavior and a growing ability to establish when that behavior can be relied upon.


\appendix
\numberwithin{algorithm}{chapter}
\chapter{Supplementary Methods for LIMP}
\label{app:limp}
\section{Overview}
\label{sec:appendix}
This appendix supplements Chapter~\ref{ch:limp} with prompting details, spatial comparators, planning algorithms, robot skill implementations, and evaluation materials for LIMP. The authorial ``we'' refers to the collaboration credited at the beginning of that chapter. Prompts and evaluation instructions are reproduced as used in the study; project demonstrations are available at \href{https://robotlimp.github.io/}{robotlimp.github.io}.

\section{Language Instruction Module}
\label{sec:instruction-module-details}
We implement a two-stage prompting strategy in our language instruction module to translate natural language instructions into LTL specifications. The first stage translates a given instruction into a conventional LTL formula, where propositions refer to open-vocabulary objects. For any given instruction, we dynamically generate $K$ in-context translation examples from a standard dataset \citep{pan_data-efficient_2023} of natural language and LTL pairs, based on cosine similarity with the given instruction. Here is the exact text prompt used:
\par
\noindent\begin{minipage}{\linewidth}
\begin{lstlisting}[language=gptprompt, basicstyle=\ttfamily\footnotesize, captionpos=t, caption=Base prompt used to obtain a conventional LTL formula from a natural language query]
You are a LLM that understands operators involved with Linear Temporal Logic (LTL), such as F, G, U, &, |, ~ , etc. Your goal is translate language input to LTL output.
Input:<generated_example_instruction>
Output:<generated_example_LTL>
...
Input:<given_instruction>
Output:
\end{lstlisting}
\end{minipage}
\par

The second stage takes the given instruction and the LTL response from the first stage as input to generate a new LTL formula with predicate functions that correspond to parameterized robot skills. Skill parameters are instruction referent objects expressed in our novel Composable Referent Descriptor (CRD) syntax. CRDs enable referent disambiguation by chaining comparators that encode descriptive spatial information. We define eight spatial comparators and provide their descriptions as part of the second stage prompt. The prompt supplies descriptions of the supported comparators and examples of how they can be composed. Here is the exact prompt used: \looseness=-1
\begin{lstlisting}[language=gptprompt, basicstyle=\ttfamily\footnotesize, captionpos=t, caption= Second stage prompt to output our LTL syntax with CRD parameterized robot skills]
You are an LLM for robot planning that understands operators involved with Linear Temporal Logic (LTL), such as F, G, U, &, |, ~ , etc. You have a finite set of robot predicates and spatial predicates, given a language instruction and an LTL formula that represents the given instruction, your goal is to translate the ltl formula into one that uses appropriate composition of robot and spatial predicates in place of propositions with relevant details from original instruction as arguments.
Robot predicate set (near,pick,release). 
Usage:
near[referent_1]:returns true if the desired spatial relationship is for robot to be near referent_1.
pick[referent_1]:can only execute picking skill on referent_1 and return True when near[referent_1].
release[referent_1,referent_2]:can only execute release skill on referent_1 and return True when near[referent_2].
Spatial predicate set (isbetween,isabove,isbelow,isleftof,isrightof,isnextto,isinfrontof,isbehind). 
Usage:
referent_1::isbetween(referent_2,referent_3):returns true if referent_1 is between referent_2 and referent_3.
referent_1::isabove(referent_2):returns True if referent_1 is above referent_2.
referent_1::isbelow(referent_2):returns True if referent_1 is below referent_2.
referent_1::isleftof(referent_2):returns True if referent_1 is left of referent_2.
referent_1::isrightof(referent_2):returns True if referent_1 is right of referent_2.
referent_1::isnextto(referent_2):returns True if referent_1 is close to referent_2.
referent_1::isinfrontof(referent_2):returns True if referent_1 is in front of referent_2.
referent_1::isbehind(referent_2):returns True if referent_1 is behind referent_2.
Rules:
Strictly only use the finite set of robot and spatial predicates!
Strictly stick to the usage format!
Compose spatial predicates where necessary!
You are allowed to modify the structure of Input_ltl for the final Output if it does not match the intended Input_instruction!
You should strictly only stick to mentioned objects, however you are allowed to propose and include plausible objects if and only if not mentioned in instruction but required based on context of instruction! 
Pay attention to instructions that require performing certain actions multiple times in generating and sequencing the predicates for the final Output formula!
Example:
Input_instruction: Go to the orange building but before that pass by the coffee shop, then go to the parking sign.
Input_ltl: F (coffee_shop & F (orange_building & F parking_sign ) )
Output: F ( near[coffee_shop] & F ( near[orange_building] & F near[parking_sign] ))
Input_instruction: Go to the blue sofa then the laptop, after that bring me the brown bag between the television and the kettle on the left of the green seat, I am standing by the sink.
Input_ltl: F ( blue_sofa & F ( laptop & F ( brown_bag & F ( sink ) ) ) )
Output: F ( near[blue_sofa] & F ( near[laptop] & F ( near[brown_bag::isbetween(television,kettle::isleftof(green_seat))] & F (pick[brown_bag::isbetween(television,kettle::isleftof(green_seat))] & F ( near[sink] & F ( release[brown_bag,sink] ) ) ) ) ) )
Input_instruction: Hey need you to pass by chair between the sofa and bag, pick up the bag and go to the orange napkin on the right of the sofa.
Input_ltl: F ( chair & F ( bag & F ( orange_napkin ) ) )
Output: F ( near[chair::isbetween(sofa,bag)] & F ( near[bag] & F ( pick[bag] & F ( near[orange_napkin::isrightof(sofa)] ) ) ) )
Input_instruction:  Go to the chair between the green laptop and the yellow box underneath the play toy
Input_ltl: F ( green_laptop & F ( yellow_box & F ( play_toy & F ( chair ) ) ) )
Output: F ( near[chair::isbetween(green_laptop,yellow_box::isbelow(play_toy))] )
Input_instruction: Check the table behind the fridge and bring two beers to the couch one after the other
Input_ltl: F ( check_table & F ( bring_beer1 ) & F ( bring_beer2 ) & F ( couch ) )
Output: F ( near[table::isbehind(fridge)] & F ( pick[beer] & F ( near[couch] & F ( release[beer,couch] & F ( near[table::isbehind(fridge)] & F ( pick[beer] & F ( near[couch] & F ( release[beer,couch] ))))))))
Input_instruction: <given_instruction>
Input_ltl: <stage1_ltl_response>
Output:
\end{lstlisting}

\subsection{Interactive Symbol Verification}
\label{sec:interactive-verification-details}
Verifying sampled LTL formulas is essential, as such we implement an interactive dialog system that presents users with extracted referent composable referent descriptors (CRDs) in sampled formulas as well as the implied task structure––encoded in the sequence of state-machine transition expressions that must hold to progressively solve the task. We translate the task structure into English statements via a simple deterministic strategy that replaces logical connectives and skill predicates from the formula with equivalent English phrases. Users can verify a formula as correct or provide corrective statements which are used to reprompt the LLM to obtain new formulas.  Below is the exact prompt used for regenerating formulas. \looseness=-1
\par
\noindent\begin{minipage}{\linewidth}
\begin{lstlisting}[language=gptprompt, basicstyle=\ttfamily\footnotesize, captionpos=t, caption=Corrective reprompting prompt used to obtain new LTL formulas]
There was a mistake with your output LTL formula: Error with <verification_type>! Consider the clarification feedback and regenerate the correct output for the Input_instruction. Make sure to adhere to all rules and instructions in your original prompt!
previous_output:<last_response>
error_clarification: <given_error_clarification>
correct_output:
\end{lstlisting}
\end{minipage}
\par
To illustrate, the instruction \textit{``Bring the green plush toy to the whiteboard in front of it"} yields the interactive Referent and Task Structure Verification dialog below:
\begin{lstlisting}[language=gptprompt, basicstyle=\ttfamily\footnotesize, captionpos=t, caption=Interactive referent and task structure verification dialog.]
**************************
Instruction Following
**************************
Input_instruction: "Bring the green plush toy to the whiteboard in front of it" 
Sampled LTL formula: F(A & F(B & F(C & FD)))
    A: near[green_plush_toy]
    B: pick[green_plush_toy]
    C: near[whiteboard::isinfrontof(green_plush_toy)]
    D: release[green_plush_toy, whiteboard::isinfrontof(green_plush_toy)]

***************************
Referent Verification
*************************** 
I extracted this list of relevant objects based on your instruction:
    * whiteboard::isinfrontof(green_plush_toy) 
    * green_plush_toy
Does this match your intention? (y/n)

****************************
Task Structure Verification
****************************
Based on my understanding here is the sequence of subgoal objectives needed to satisfy the task:
Subgoal_1:
    Logical Expression: A&!B
    English translation: I should be near the [green_plush_toy] and not have picked up the [green_plush_toy] 
Subgoal_2:
    Logical Expression: B&!C
    English translation: I should have picked up the [green_plush_toy] and not be near the [whiteboard::isinfrontof(green_plush_toy)] 
Subgoal_3:
    Logical Expression: C&!D
    English translation: I should be near the [whiteboard::isinfrontof(green_plush_toy)] and not have released the [green_plush_toy] at the [whiteboard::isinfrontof(green_plush_toy)]
Subgoal_4:
    Logical Expression: D
    English translation: I should have released the [green_plush_toy] at the [whiteboard::isinfrontof(green_plush_toy)] 
Does this match your intention? (y/n)
\end{lstlisting}

\section{Spatial Grounding Module}
\label{sec:spatial-grounding-details}
The spatial grounding module detects and localizes specific instances of objects referenced in a given instruction by first detecting, segmenting and back-projecting all referent occurrences and then filtering based on the descriptive spatial details captured by each referent's composable referent descriptor (CRD). We use the Owl-Vit model \citep{minderer_simple_2022} to detect bounding boxes of open-vocabulary referents and SAM \citep{kirillov_segment_2023} to generate masks from detected bounding boxes. To illustrate referent filtering via spatial information, consider an example scenario where the goal is to resolve the composable referent descriptor below: 
\begin{equation}
\text{whiteboard}::isinfrontof( \text{green\_plush\_toy}).
\end{equation}
Let $W = \{w_1, w_2, \ldots, w_n\}$ and $G = \{g_1, g_2, \ldots, g_m\}$ represent the set of representative 3D positions of detected whiteboards and green\_plush\_toys respectively. The cartesian product of these sets enumerates all possible pairs $(w, g)$ for comparison.
\begin{equation}
    W \times G = \{ (w, g) \mid w \in W, g \in G \}
\end{equation}

The `isinfrontof(w, g)' comparator is applied to each pair, yielding a subset \( S \) that contains only those `whiteboards` that satisfy the `isinfrontof' condition with at least one `green\_plush\_toy'.
\begin{equation}
    S = \{ w \in W \mid \exists g \in G \text{ such that } \text{isinfrontof}(w, g) \text{ is true} \}
\end{equation}

\subsection{3D Spatial Comparators} 
Our 3D spatial comparators enable Relative Frame of Reference (FoR) spatial reasoning between referents, based on their backprojected 3D positions. Threshold values in the spatial comparators give users the ability to specify the sensitivity or resolution at which spatial relationships are resolved, we keep all threshold values fixed across all experiments. Below is a description of each spatial comparator. \looseness=-1

\begin{lstlisting}[language=gptprompt, basicstyle=\ttfamily\footnotesize, captionpos=t, numbers=none, caption= Implementation description of 3D spatial comparators]
1. isbetween(referent_1_pos, referent_2_pos, referent_3_pos, threshold): Returns true if referent_1 is within 'threshold' distance from the line segment connecting referent_2 to referent_3, ensuring it lies in the directional path between them without extending beyond.
2. isabove(referent_1_pos, referent_2_pos, threshold): Returns true if the z-coordinate of referent_1 exceeds that of referent_2 by at least 'threshold'.
3. isbelow(referent_1_pos, referent_2_pos, threshold): Returns true if the z-coordinate of referent_1 is less than that of referent_2 by more than 'threshold'.
4. isleftof(referent_1_pos, referent_2_pos, threshold): Returns true if the y-coordinate of referent_1 exceeds that of referent_2 by at least 'threshold', indicating referent_1 is to the left of referent_2.
5. isrightof(referent_1_pos, referent_2_pos, threshold): Returns true if the y-coordinate of referent_1 is less than that of referent_2 by more than 'threshold', indicating referent_1 is to the right of referent_2.
6. isnextto(referent_1_pos, referent_2_pos, threshold): Returns true if the Euclidean distance between referent_1 and referent_2 is less than 'threshold', indicating they are next to each other.
7. isinfrontof(referent_1_pos, referent_2_pos, threshold): Returns true if the x-coordinate of referent_1 is less than that of referent_2 by more than 'threshold', indicating referent_1 is in front of referent_2.
8. isbehind(referent_1_pos, referent_2_pos, threshold): Returns true if the x-coordinate of referent_1 exceeds that of referent_2 by at least 'threshold', indicating referent_1 is behind referent_2.
\end{lstlisting}

\section{Task and Motion Planning Module} 
\label{sec:planning-module-details}
We present pseudocode for our Progressive Motion Planner (Algorithm~\ref{alg:tamp-algo}) and our algorithm for generating Task Progression Semantic Maps (Algorithm~\ref{alg:tpsm-algo}). Algorithm~\ref{alg:tpsm-algo} generates a TPSM $\mathcal{M}_{\text{tpsm}}$ by integrating an environment map ($\mathcal{M}$) and a referent semantic map ($\mathcal{M}_{\text{rsm}}$) given a logical transition expression ($\mathcal{T}$), a desired automaton state ($\mathcal{S'}$), and a nearness threshold ($\theta$). The algorithm first initializes $\mathcal{M}_{\text{tpsm}}$ with a copy of $\mathcal{M}$ and extracts relevant instruction predicates from $\mathcal{T}$. For each predicate (parameterized skill), the algorithm identifies satisfying referent positions in $\mathcal{M}_{\text{rsm}}$, generates a spherical grid of surrounding points within a radius $\theta$, and assesses how these points affect the progression of the task automaton towards $\mathcal{S'}$. These points demarcate regions of interest, and are assigned a value of \textit{1} if they cause the automaton to transition to the desired state, \textit{-1} if they lead to a different automaton state or violate the automaton, and \textit{0} if they do not affect the automaton. The points are then integrated into $\mathcal{M}_{\text{tpsm}}$, yielding a semantic map that identifies goal and constraint violating regions.
\vspace{2mm}
\begin{algorithm}[p]
\caption{Progressive Motion Planning Algorithm}

\label{alg:tamp-algo}
\begin{algorithmic}[1]
\STATE \textbf{procedure} \textsc{PMP}$(X_{start}, \varphi, \mathcal{M}, \mathcal{M}_{rsm}, \theta)$
    \item[] \textbf{Input:}
    \item[] \quad \( X_{start} \): Start position in the environment.
    \item[] \quad \( \varphi \): CRD syntax LTL formula specifying task objectives.
    \item[] \quad \( \mathcal{M} \): Environment map.
    \item[] \quad \( \mathcal{M}_{rsm} \): Referent semantic map.
    \item[] \quad \( \theta \): Nearness threshold.

    \item[] \textbf{Output:}
    \item[] \quad \( \Pi \): Generated task and motion plan.

    \STATE \( \mathcal{A} \leftarrow \text{ConstructAutomaton}(\varphi) \)
    \STATE \( \text{path} \leftarrow \text{SelectAutomatonPath}(\mathcal{A}) \) \COMMENT{Task plan}

    \WHILE{$\Pi$.\text{status is active}}
        \WHILE{$\mathcal{A}$.state != path.acceptingState}
        \STATE \( \mathcal{S}, \mathcal{T}, \mathcal{S'} \leftarrow \mathcal{A}.\text{GetTransition(path.currentStep)} \)
            \STATE \( \text{objective} \leftarrow \text{NextObjectiveType}(\mathcal{T}) \)

            \IF{\( \text{objective} = \text{``skill''} \)}
                \STATE \( \Pi.\text{UpdateWithSkill}(\mathcal{T}) \)
                \STATE $\mathcal{A}$.UpdateAutomatonState($\mathcal{S'}$)
            \ELSIF{\( \text{objective} = \text{``navigation''} \)}
                \STATE \( \mathcal{M}_{params} \leftarrow \mathcal{M}, \mathcal{M}_{rsm}, \mathcal{T}, \mathcal{A}, \mathcal{S'}, \theta \)
                \STATE \( \mathcal{M}_{tpsm} \leftarrow \textsc{GenerateTPSM}(\mathcal{M}_{params}) \)
                \STATE \( \mathcal{O} \leftarrow \text{GenerateObstacleMap}(\mathcal{M}_{tpsm}) \)
                \STATE \( \text{plan} \leftarrow \text{FMT}^*(X_{start}, \mathcal{O}) \) \COMMENT{Path plan}
                \IF{\( \text{plan.exists} \)}
                    \STATE \( \Pi.\text{UpdateWithPlan}(\text{plan}) \)
                    \STATE \( X_{start} \leftarrow \text{plan.endPosition} \)
                    \STATE $\mathcal{A}$.UpdateAutomatonState($\mathcal{S'}$)
                \ELSE
                    \STATE \( \Pi, \mathcal{A}, \text{path} \leftarrow \text{Backtrack}(\Pi, \mathcal{A}, \text{path}) \)
                \ENDIF
            \ENDIF
        \ENDWHILE
    \ENDWHILE
    \STATE \textbf{return} \( \Pi \)
\STATE \textbf{end procedure}
\end{algorithmic}
\end{algorithm}

\begin{algorithm}[p]
\caption{Task Progression Semantic Mapping Algorithm}
\label{alg:tpsm-algo}
\begin{algorithmic}[1]
\STATE \textbf{procedure} \textsc{GenerateTPSM}$(\mathcal{M}, \mathcal{M}_{rsm}, \mathcal{T}, \mathcal{A}, \mathcal{S'}, \theta)$
    \item[] \textbf{Input:}
    \item[] \quad \( \mathcal{M} \): Environment map.
    \item[] \quad \( \mathcal{M}_{rsm} \): Referent semantic map.
    \item[] \quad \( \mathcal{T} \): Automaton transition expression.
    \item[] \quad \( \mathcal{A} \): Task Automaton.
    \item[] \quad \( \mathcal{S'} \): Desired State.
    \item[] \quad \( \theta \): Nearness threshold.

    \item[] \textbf{Output:}
    \item[] \quad \( \mathcal{M}_{tpsm} \): Task Progression Semantic Map.

    \STATE \( \mathcal{M}_{tpsm} \leftarrow \text{Copy}(\mathcal{M}) \)
    \STATE \( \mathcal{P} \leftarrow \text{ExtractRelevantPredicates}(\mathcal{T}) \)

    \FOR{\( p \) in \( \mathcal{P} \)}
        \STATE \( \mathcal{R} \leftarrow \text{QueryPositions}(\mathcal{M}_{rsm}, p) \)
        \FOR{\( r \) in \( \mathcal{R} \)}
            \STATE \( G \leftarrow \left\{ g \mid g = r + \delta, \lVert \delta \rVert \leq \theta \right\} \)  \COMMENT{spherical grid of surrounding points}        
            \FOR{\( g \) in \( G \)}
                \STATE \( \mathcal{Q} \leftarrow \text{TruePredicatesAt}(g, \mathcal{M}_{rsm}, \theta) \)
                \STATE \( \mathcal{S}_{next} \leftarrow \text{ProgressAutomaton}(\mathcal{A}, \mathcal{Q}) \)
                \IF{\( \mathcal{S}_{next} = \mathcal{S'} \)}
                    \STATE \( g.value \leftarrow 1 \) \COMMENT{Goal value}
                \ELSIF{\( \text{IsUndesired}(\mathcal{S}_{next}) \)}
                    \STATE \( g.value \leftarrow -1 \) \COMMENT{Avoidance value}
                \ELSE
                \STATE \( g.\text{value} \leftarrow 0 \)
                \ENDIF
            \ENDFOR
            \STATE \( \text{AddPoints}(\mathcal{M}_{tpsm}, G) \)
        \ENDFOR
    \ENDFOR

    \STATE \textbf{return} \( \mathcal{M}_{tpsm} \)
\STATE \textbf{end procedure}
\end{algorithmic}
\end{algorithm}

\clearpage
\section{Robot Skills}
\label{sec:robo-skills}
We define three predicate functions: \textbf{near}, \textbf{pick} and \textbf{release} for the navigation, picking and placing skills required for multi-object goal navigation and mobile pick-and-place. As described in Chapter~\ref{ch:limp}, we formalize navigation as continuous path planning problems and manipulation as object parameterized options. Chapter~\ref{ch:limp} discusses navigation, so here we focus on the pick and place manipulation skills.

\vspace{2mm}
\noindent\textbf{Pick Skill}: Once the robot has executed the near skill and is at the object to be manipulated, it takes a photo of the current environment to detect the object using the Owl-Vit model. The path planner uses the backprojected object position to compute yaw angles that orient the robot toward the target. After detecting the object, the robot obtains a segmentation mask from the detected bounding box using the Segment Anything model and computes the center pixel of this mask. It passes this pixel to the Boston Dynamics grasping API to compute a grasp motion plan. 

\vspace{2mm}
\noindent\textbf{Release Skill}: We implement a simple routine for the release skill which takes two parameters: the object to be placed and the place receptacle. Once a navigation skill gets the robot to the place receptacle, the robot gently moves its arm up or down to release the grasped object, based on the place receptacle 3D position. More expressive placement strategies could extend the use of LIMP's spatial grounding of instruction-specific receptacles. Demonstrations are available on the \href{https://robotlimp.github.io/}{\color{blue}{website}}. \looseness=-1

\section{Evaluation and Baseline Details} 
\label{sec:eval-details}
All computation including planning, loading and running pretrained visual language models was done on a single computer equipped with one NVIDIA GeForce RTX 3090 GPU. We leverage GPT-4-0613 as the underlying LLM for our instruction understanding module and all our baselines. In all experiments we set the LLM temperature to 0, however since deterministic greedy token decoding is not guaranteed with GPT4, we perform three (3) queries for each instruction and evaluate on the most recurring response (mode response). \looseness=-1

We compare LIMP with baseline implementations of NLMap-Saycan \citep{chen_open-vocabulary_2023} and Code-as-policies \citep{liang_code_2023}. Both baselines use the same GPT-4 LLM, prompting structure, and in-context learning examples as our language understanding module. We integrate our composable referent descriptor syntax, spatial grounding module and low-level robot control into these baselines as APIs. This enables baselines to execute plans by querying relevant object positions, using our FMT* path planner to find paths to said positions and executing manipulation options. \looseness=-1

\begin{figure}[tbp]
\centering

 \includegraphics[width=\textwidth]{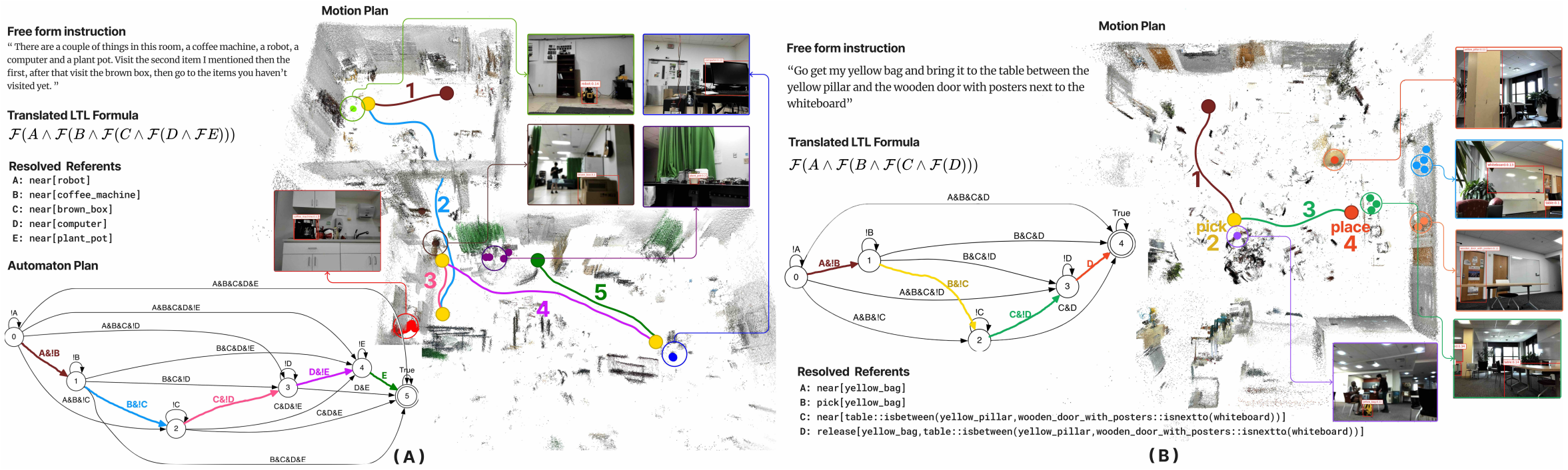}%
   \vspace*{1mm}
  \caption[Example plans for navigation and mobile pick-and-place]{\textbf{[A]} Sample generated plan for a multi object-goal navigation task. \textbf{[B]} Sample generated plan for a mobile pick-and-place task.}
  \label{fig:qualitative}
    
\end{figure}

\begin{figure}
\centering
\includegraphics[width=0.85\textwidth]{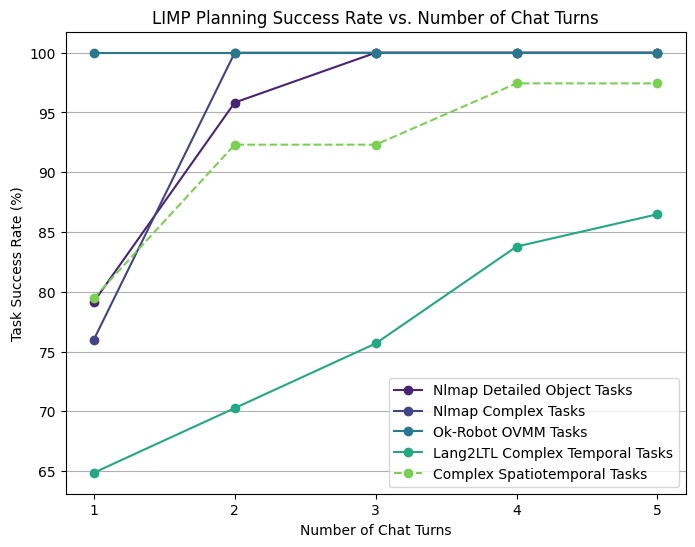}
\caption[Plan success with interactive reprompting]{Our interactive reprompting strategy implemented in the symbol verification node regenerates corrective formulas that improve plan success rates with minimal chat turns.}
\label{fig:language_module_chatturns}

\end{figure}

Figure~\ref{fig:qualitative} shows qualitative planning results. Figure~\ref{fig:language_module_chatturns} reports task success across successive chat turns with the interactive verification and reprompting procedure described in Section~\ref{sec:interactive-verification-details}. \looseness=-1

\clearpage
\subsection{NLMap-Saycan Implementation Prompt}
\begin{lstlisting}[language=gptprompt, basicstyle=\ttfamily\footnotesize, captionpos=t, caption= Exact prompt to implement NlMap-Saycan LLM planner]
You are an LLM for robot planning that understands logical operators such as &, |, ~ , etc. You have a finite set of robot predicates and spatial predicates, given a language instruction, your goal is to generate a sequence of actions that uses appropriate composition of robot and spatial predicates with relevant details from the instruction as arguments.
Robot predicate set (near,pick,release). 
Usage:
near[referent_1]:returns true if the desired spatial relationship is for robot to be near referent_1.
pick[referent_1]:can only execute picking skill on referent_1 and return True when near[referent_1].
release[referent_1,referent_2]:can only execute release skill on referent_1 and return True when near[referent_2].
Spatial predicate set (isbetween,isabove,isbelow,isleftof,isrightof,isnextto,isinfrontof,isbehind). 
Usage:
referent_1::isbetween(referent_2,referent_3):returns true if referent_1 is between referent_2 and referent_3.
referent_1::isabove(referent_2):returns True if referent_1 is above referent_2.
referent_1::isbelow(referent_2):returns True if referent_1 is below referent_2.
referent_1::isleftof(referent_2):returns True if referent_1 is left of referent_2.
referent_1::isrightof(referent_2):returns True if referent_1 is right of referent_2.
referent_1::isnextto(referent_2):returns True if referent_1 is close to referent_2.
referent_1::isinfrontof(referent_2):returns True if referent_1 is in front of referent_2.
referent_1::isbehind(referent_2):returns True if referent_1 is behind referent_2.
Rules:
Strictly only use the finite set of robot and spatial predicates!
Strictly stick to the usage format!
Compose spatial predicates where necessary!
You should strictly stick to mentioned objects, however you are allowed to propose and include plausible objects if and only if not mentioned in instruction but required based on context of instruction! 
Pay attention to instructions that require performing certain actions multiple times in generating and sequencing the predicates for the final Output!
Example:
Input_instruction: Go to the orange building but before that pass by the coffee shop, then go to the parking sign.
Output: 
1. near[coffee_shop] 
2. near[orange_building]
3. near[parking_sign]
Input_instruction: Go to the blue sofa then the laptop, after that bring me the brown bag between the television and the kettle on the left of the green seat, I am standing by the sink.
Output: 
1. near[blue_sofa]
2. near[laptop]
3. near[brown_bag::isbetween(television,kettle::isleftof(green_seat))] 
4. pick[brown_bag::isbetween(television,kettle::isleftof(green_seat))] 
5. near[sink] 
6. release[brown_bag,sink]
Input_instruction: Hey need you to pass by chair between the sofa and bag, pick up the bag and go to the orange napkin on the right of the sofa.
Output: 
1. near[chair::isbetween(sofa,bag)] 
2. near[bag] 
3. pick[bag]
4. near[orange_napkin::isrightof(sofa)]
Input_instruction:  Go to the chair between the green laptop and the yellow box underneath the play toy
Output: 
1. near[chair::isbetween(green_laptop,yellow_box::isbelow(play_toy))]
Input_instruction: Check the table behind the fridge and bring two beers to the couch one after the other
Output: 
1. near[table::isbehind(fridge)] 
2. pick[beer] 
3. near[couch] 
4. release[beer,couch] 
5. near[table::isbehind(fridge)] 
6. pick[beer] 
7. near[couch] 
8. release[beer,couch]
Input_instruction: <given_instruction>
Output:
\end{lstlisting}

\subsection{Code-as-Policies Implementation Prompt}
\begin{lstlisting}[language=gptprompt, basicstyle=\ttfamily\footnotesize, captionpos=t, caption= Exact prompt to implement Code-as-policies planner]
##Python robot planning script
from robotactions import near, pick, release
spatial_relationships = [
"isbetween", #referent_1::isbetween(referent_2,referent_3):returns true if referent_1 is between referent_2 and referent_3.
"isabove",   #referent_1::isabove(referent_2):returns True if referent_1 is above referent_2.
"isbelow",   #referent_1::isbelow(referent_2):returns True if referent_1 is below referent_2.
"isleftof",  #referent_1::isleftof(referent_2):returns True if referent_1 is left of referent_2.
"isrightof", #referent_1::isrightof(referent_2):returns True if referent_1 is right of referent_2.
"isnextto",  #referent_1::isnextto(referent_2):returns True if referent_1 is close to referent_2.
"isinfrontof", #referent_1::isinfrontof(referent_2):returns True if referent_1 is in front of referent_2.
"isbehind"  #referent_1::isbehind(referent_2):returns True if referent_1 is behind referent_2.]
##Rules:
##Strictly only use the finite set of robot and spatial predicates!
##Strictly stick to the usage format!
##Compose spatial predicates where necessary!
##You should strictly stick to mentioned objects, however you are allowed to propose and include plausible objects if and only if not mentioned in instruction but required based on context of instruction! 
##Pay attention to instructions that require performing certain actions multiple times in generating and sequencing the predicates for the final Output!
# Go to the orange building but before that pass by the coffee shop, then go to the parking sign.
ordered_navigation_goal_referents = ["coffee_shop", "orange_building", "parking_sign"]
for referent in ordered_navigation_goal_referents:
    near(referent)
# Go to the blue sofa then the laptop, after that bring me the brown bag between the television and the kettle on the left of the green seat, I am standing by the sink.
ordered_navigation_goal_referents = ["blue_sofa", "laptop", "brown_bag::isbetween(television,kettle::isleftof(green_seat))", "sink"]
referents_to_pick= ["brown_bag::isbetween(television,kettle::isleftof(green_seat))"]
release_location_referents = ["sink"]
picked_item = None
for referent in ordered_navigation_goal_referents:
    near(referent)
    if referent in referents_to_pick:
        pick(referent)
        picked_item = referent
    if referent in release_location_referents:
        release(picked_item, referent)
#Hey need you to pass by chair between the sofa and bag, pick up the bag and go to the orange napkin on the right of the sofa.
ordered_navigation_goal_referents = ["chair::isbetween(sofa,bag)", "bag", "orange_napkin::isrightof(sofa)"]
referents_to_pick= ["bag"]
picked_item = None
for referent in ordered_navigation_goal_referents:
    near(referent)
    if referent in referents_to_pick:
        pick(referent)
        picked_item = referent
#Go to the chair between the green laptop and the yellow box underneath the play toy
near("chair::isbetween(green_laptop,yellow_box::isbelow(play_toy))")
#Check the table behind the fridge and bring two beers to the couch one after the other
for i in range(2):
    near("table::isbehind(fridge)")
    pick("beer")
    near("couch")
    release("beer", "couch")

#<given_instruction>
\end{lstlisting}

\subsection{Instruction set}
\label{sec:instruction-set-details}
We perform a large-scale evaluation on 150 instructions across five real-world environments. The taskset includes 24 tasks with fine-grained object descriptions (NLMD) from \citet{chen_open-vocabulary_2023}, 25 tasks with complex language (NLMC) from \citet{chen_open-vocabulary_2023}, 25 tasks with simple structured phrasing (OKRB) from \citet{liu_demonstrating_2024}, 37 tasks with complex temporal structures (CT) from \citet{liu_grounding_2023}, and an additional 39 tasks we propose that have descriptive spatial constraints and temporal structures (CST).

\begin{lstlisting}[language=instructions, basicstyle=\ttfamily\footnotesize, captionpos=t, numbers=none, caption=Nlmap Detailed Object Tasks (NLMD)]
1: put the red can in the trash bin
2: put the brown multigrain chip bag in the woven basket
3: find the succulent plant
4: pick up the up side down mug
5: put the apple on the macbook with yellow stickers
6: use the dyson vacuum cleaner"
7: bring the kosher salt to the kitchen counter
8: put the used towels in washing machine
9: move the used mug to the dish washer
10: place the pickled cucumbers on the shelf
11: find my mug with the shape of a donut
12: put the almonds in the almond jar
13: fill the zisha tea pot with a coke from the cabinet
14: take the slippery floor sign with you
15: take the slippers that have holes on them to the shoe rack
16: find the mug on the mini fridge
17: bring the mint flavor gum to the small table
18: find some n95 masks
19: grab the banana with most black spots
20: fill the empty bottle with lemon juice
21: throw away the apple that's about to rot
22: throw away the rotting banana
23: take the box of organic blueberries out of the fridge
24: give a can of diet coke to the toy cat
\end{lstlisting}

\begin{lstlisting}[language=instructions, basicstyle=\ttfamily\footnotesize, captionpos=t, numbers=none, caption=Nlmap Complex Language Understanding Tasks (NLMC)]
1: I opened a pepsi earlier, bring an open can to the orange table
2: I spilled my coke, can you put a replacement on the kitchen counter
3: I spilled some coke on the television, go and bring something to clean it up
4: I accidentally dropped that jalapeno chips after eating it. Would you mind throwing it away
5: I like fruits, can you put something I would like on the yellow sofa for me
6: There is a green counter, a yellow counter, and a table. visit all the locations
7: There is a green counter, a trash can, and a table. visit all the locations
8: Redbull is my faviorite drink, can you put one on the desk please
9: Would you bring a coke can to the door for me
10: Please, move the pepsi to the red counter
11: Can you move the coke can to the orange counter
12: Would you throw away the bag of chips for me
13: Put an energy bar and water bottle on the table
14: Bring a lime soda and a bag of chips to the sofa
15: Can you throw away the apple and bring a coke to the bed
16: Bring a 7up can and a tea to the office desk
17: Move the multigrain chips to the table and an apple to the yellow counter
18: Move the lime soda, the sponge, and the water bottle to the table
19: Bring two sodas to the table
20: Move three cokes to the trash can
21: Throw away two cokes from the counter
22: Bring two different sodas to the cabinet, there is a coke, pepsi, soup, tea and 7up in the fridge
23: Bring an apple, a coke, and water bottle to the sofa
24: I spilled my coke on the table, throw it away and then bring something to help clean
25: I just worked out, can you bring me a drink and a snack to recover, i am on the sof
\end{lstlisting}

\begin{lstlisting}[language=instructions, basicstyle=\ttfamily\footnotesize, captionpos=t, numbers=none, caption=Ok-Robot Tasks (OKRB)]
1: Move the Takis on the desk to the nightstand
2: Move the soda can to the box
3: Move the purple shampoo to the red bag
4: Move the white meds box to the trash bin
5: Move the power adapter to the chair
6: Move the blue gloves to the sink
7: Move the McDonalds paper bag to the stove
8: Move the herbal tea can to the box
9: Move the cooking oil bottle to the marble surface
10: Move the milk bottle to the chair
11: Move the purple shampoo to the white rack
12: Move the purple lightbulb box to the sofa chair
13: Pick up purple medicine, drop it on chair
14: Pick up eyeglass case, drop it on chair
15: Pick up grey rag, drop it in sink
16: Pick up golden can rag, drop it on table
17: Pick up red navaratna oil, drop it on table
18: Pick up purple shampoo, drop it on green rack
19: Pick up taki chips, drop it on countertop
20: Pick up bandage box, drop it in dustbin
21: Pick up white aerosol, drop it in trash can
22: Pick up peanut butter, drop it on countertop
23: Pick up blue gloves, drop it in sink
24: Pick up brown box, drop it on chair
25: Pick up axe body spray, drop it on shel
\end{lstlisting}

\begin{lstlisting}[language=instructions, basicstyle=\ttfamily\footnotesize, captionpos=t, numbers=none, caption=Lang2LTL Complex temporal Tasks (CT)]
1: go to brown bookshelf, metal desk, wooden desk, kitchen counter, and the blue couch in any order
2: move to grey door, then bookrack, then go to the brown desk, then counter, then white desk  
3: visit brown wooden desk but only after bookshelf  
4: go from brown bookshelf to white metal desk and only visit each landmark one time  
5: go to brown wooden desk exactly once and do not visit brown desk before bookshelf  
6: go to brown desk only after visiting bookshelf, in addition go to brown desk only after visiting white desk  
7: visit the blue IKEA couch, in addition never go to the big steel door  
8: visit white kitchen counter then go to brown desk, in addition never visit white table  
9: go to the grey door, and only then go to the bookshelf, in addition always avoid the table  
10: go to kitchen counter then wooden desk, in addition after going to counter, you must avoid white table  
11: Go to bookshelf, alternatively go to metal desk  
12: Go to counter, alternatively go to metal desk  
13: Go to the counter, but never visit the counter  
14: do not go to the wooden desk until bookshelf, and do not go to bookshelf until wooden desk  
15: go to brown desk exactly once, in addition go to brown desk at least twice  
16: move to couch exactly twice, in addition pass by counter at most once
17: navigate to the counter then the brown desk, in addition after going to the counter, you must avoid doorway  
18: visit counter at least six times
19: either go to bookshelf then the brown desk, or go to couch
20: navigate to the wooden door, the glass door and the table, kitchen counter, and the blue couch in any order  
21: go to the painting, then find the kitchen table, then front desk, then staircase  
22: navigate to classroom but do not visit classroom before the white table  
23: only visit classroom once, and do not visit classroom until you visit elevator first  
24: Go to the staircase, front desk and the white table in that exact order. You are not permitted to revisit any of these locations  
25: go to the front desk then the yellow office door, in addition do not visit the glass door  
26: go to the stairs then the front desk, in addition avoid purple elevator  
27: move to elevator then front desk, in addition avoid staircase  
28: go to front desk then the cabinet, always avoid the elevator  
29: Go to elevator, alternatively go to staircase  
30: Visit the elevator exactly once, in addition visit the front desk on at least 2 separate occasions  
31: Go to the office, in addition avoid visiting the elevator and the classroom  
32: Visit the front desk, in addition you are not permitted to visit elevator and staircase  
33: Visit the purple door elevator, then go to the front desk and then go to the kitchen table, in addition you can never go to the elevator once you have seen the front desk
34: Visit the front desk then the sofa then the white table, in addition if you visit the sofa you must avoid the television after that
35: Go to the glass door, but never visit the glass door  
36: do not go to the white table until classroom, and do not go to the classroom until white table  
37: find the office, in addition avoid visiting the front desk and the classroom and the table
\end{lstlisting}

\begin{lstlisting}[language=instructions, basicstyle=\ttfamily\footnotesize, captionpos=t, numbers=none, caption=Complex Spatiotemporal Tasks (CST)]
1: Go to the red sofa but dont pass by any door then go to the computer, the sofa is behind the pillar and on the right side
2: Find the door with posters in front of the yellow pillar then go stand by the tree but avoid any sofa
3: There is a cabinet, a television and a tree in this room. I want you to go to the last item I mentioned then pass by the second item but dont go close to the first when you are going to thsecond item.",
4: Go to the trash bin, after that go to the large television. Actually can you go to the laptop on the table before doing all that?
5: Try to go to the tree without going near the trash bin.
6: Find the yellow trash bin and go to it, then go by the white table. But before doing the first thing visit where the fridge is then after the last thing, stop by the red sofa but avoid theooden door with posters",
7: Go to the television but avoid the tree.
8: Go to any trash bin but not the yellow one
9: Visit one of the televisions in this room then go to the fridge to the left of the cabinet
10: I need you to first go to the table between the pillar and the door with posters, then go to the tree. You know what, ignore the last thing I asked you to go to, after the first thing go tthe computer next to the green box instead.
11: Go to the red sofa then the blue one next to the cabinet but first pass by the book shelf
12: Go to water filter, dont pass by any computer to the left of the red sofa
13: Visit these things in the following order the microwave, then the black chair infront of the yellow robot then the fridge, but before doing any of this go to the robot
14: Go to the cabinet between the blue sofa and the brown box. 
15: I want you to go the cabinet but dont go near any black chair. Hold on, actually can you pass by the whiteboard before doing all that?
16: I need you to go to the red sofa but dont go anywhere near the sink
17: You can either go as close as you can to the sink or the plant pot but dont go near the yellow robot.
18: Go to the green curtain after that the large television but try not to go near the red sofa
19: Go to the sink then the red sofa, then the guitar. After that I want you to return to the first thing you visited.
20: I need you to go to the sink then any computer on a table but before you start that can you first go to the bookshelf, when you are done with all that, return to brown box
21: There are a couple of things in this room, a coffee machine, a robot, a computer and a plant pot. Visit the second item I mentioned then the first, after that visit the brown box, then go  the items you haven't visited yet.
22: Avoid any cabinet but go get the mug on the blue sofa and take it to red one
23: Please get me my plant pot from between the sofa and the sink, bring it to the white cabinet
24: Can you pick up my book from the left side of the television and bring it to me? I am sitting on the couch which is to the right of the door with posters. Make sure to never go near the trhbin while doing all this, it on the yellow table.
25: Bring the green plush toy to the whiteboard infront of it, watchout for the robot infront of the toy
26: Go get my yellow bag and bring it to the table between the yellow pillar and the wooden door with posters next to the whiteboard
27: Find the bottle that is on the table to the left of the computer and bring it to the wardrobe that is next to the glass door
28: Go to the cabinet between the blue sofa and the yellow robot or the cabinet to the left of the blue sofa. 
29: Go take the green toy that is next to the sofa under the poster and bring it to the bookshelf
30: I have a white cabinet, a green toy, a bookshelf and a red chair around here somewhere. Take the second item I mentioned from between the first item and the third. Bring it the cabinet butvoid the last item at all costs.
31: Hey I am standing by the whiteboard infront of the bookshelf, can you bring me the mug from the table to the left of the fridge?
32: Never pass by a robot, but i need you to bring the bag on the table to the cabinet
33: Take a soda can to the fridge, you can find one on the table to the left of the blue sofa
34: Hey, can you pick up my bag for me? It is under the table infront of the glass door, bring it to the cabinet. Actually you know what, forget what i asked for and bring the green toy instea its at the same place",
35: Go to the sink and take the mug beside the coffee maker, drop it off at the red sofa
36: Go to the sofa behind the chair next to the orange counter in front of the pillar
37: I want you to go visit the chair behind the sofa then the green bag, do that 3 times but during the second time avoid the fridge 
38: Go bring the plush toy between the sofa and the television to the couch, my cat is on the brown table so please dont pass near the table when you are returning with the toy
39: I think I left my wallet on the kitchen counter, go get it i will meet you at the bed. There is a lantern near the bed, make sure you dont hit it
\end{lstlisting}

\chapter{Supplementary Methods for REALM}
\label{app:realm}
\section{State Lifecycle and Promotion}
\label{app:lifecycle}

Table~\ref{tab:schema} summarizes the logical contents of a world revision. Each stored artifact is addressed relative to one run root and paired with a SHA-256 digest. A revision is registered after its checkpoint, twin, planning cloud, semantic registry, transition delta, and validation report are staged successfully.

\begin{table}[h]
\centering
\caption{Logical contents of a persistent world revision.}
\label{tab:schema}
\small
\begin{tabularx}{0.96\textwidth}{@{}p{0.18\textwidth}p{0.20\textwidth}X@{}}
\toprule
Field & Examples & Role \\
\midrule
Identity & state ID, logical time, revision type & Distinguishes observed, predicted, reconciled-synthetic, and reconciled-real worlds. \\
Authority & status, current-state index & Identifies the one state from which subsequent views may be generated. \\
Persistent state & Gaussian checkpoint, object registry & Stores renderable scene content, stable objects, poses, and ownership. \\
Derived views & digital twin, planning cloud, grounding/render bundle & Supplies instruction-specific geometry and observations bound to the state. \\
Action evidence & plan, trajectory binding, released pose & Explains the intervention used to predict the successor. \\
Observation evidence & calibrated RGB-D, masks, provenance & Explains the appearance and disocclusion information used in reconciliation. \\
History & parent ID, reversible delta & Connects each successor to the world from which it evolved. \\
Validation & locality, fidelity, authority, integrity & Determines whether a candidate may become current. \\
\bottomrule
\end{tabularx}
\end{table}

After checks on the plan binding, rigid transformation, ownership, and history, a prediction can become the current revision before post-action evidence arrives. This is provisional authority based on the planned outcome, not observation-based validation. Reconciliation creates a candidate at the same logical timestep without overwriting the prediction. Promotion revalidates the lineage, marks the prediction historical, and atomically grants the reconciled successor current authority. Alternative observation and fit variants can remain as sibling candidates; the demonstrated sequence promotes one mainline.

Promotion of a reconciled candidate proceeds as follows: (1) re-hash registered artifacts; (2) check finite and shape-consistent Gaussian tensors; (3) verify stable owner IDs and agreement between twin indices and checkpoint ownership; (4) replay inverse deltas to reconstruct every ancestor; (5) enforce action-owned geometry and opacity; (6) evaluate held-out changed-region and outside-region criteria; and (7) update the current-state index atomically. Failure leaves the preceding current state unchanged.

\section{Ownership Diagnostics}
\label{app:ownership}

The selector in Eq.~\ref{eq:selector} combines complementary evidence. Positive depth-consistent voting locates rows that project through the object; visible contradictions reject support inconsistent with other cameras; covariance-footprint overlap tests whether the rendered row actually covers the mask; surface and oriented-box distances bound the object spatially; and scale and anisotropy prevent scene-scale splats from becoming movable object support.

The legacy ball support illustrates their interaction. Of 724 selected centers, 470 lie outside the fitted ball OBB, 443 lie more than 0.01 world units outside it, and 180 have fewer than three footprint-supported views. Median anisotropy is 95.529 and the maximum is 177,390.8. These rows visibly include table and ground content. The combined selector retains 254 rows and is used throughout both transitions. Exact thresholds and per-row diagnostics are stored in the ownership-quality artifact.

Ownership construction rejects empty or insufficient supports, duplicate and out-of-bounds indices, and overlap between independently movable objects. The current analysis covers one ball; extending ownership to thin, articulated, reflective, transparent, and heavily occluded objects remains a central generalization problem.

\section{Task Specifications and Plan Evidence}
\label{app:tasks}

The temporal-logic specification for instruction 1 is
\begin{align*}
\varphi_1 &= \mathbf{F}(A\wedge \mathbf{F}(B\wedge \mathbf{F}(C\wedge \mathbf{F}D))),\\
A&=\mathrm{near}[\mathrm{blue\_ball::isbelow(table)}],\\
B&=\mathrm{pick}[\mathrm{blue\_ball::isbelow(table)}],\\
C&=\mathrm{near}[\mathrm{potted\_lime\_green\_conifer}],\\
D&=\mathrm{release}[\mathrm{blue\_ball},\mathrm{potted\_lime\_green\_conifer}].
\end{align*}
Instruction 2 is encoded as
\begin{align*}
\varphi_2 &= \mathbf{F}(A\wedge \mathbf{F}(B\wedge \mathbf{F}(C\wedge \mathbf{F}(D\wedge \mathbf{F}E)))),\\
A&=\mathrm{near}[\mathrm{small\_garden\_ornament\_in\_bushes}],\\
B&=\mathrm{near}[\mathrm{blue\_ball}],\quad C=\mathrm{pick}[\mathrm{blue\_ball}],\\
D&=\mathrm{near}[\mathrm{table}],\quad E=\mathrm{release}[\mathrm{blue\_ball},\mathrm{table}].
\end{align*}
Here $\mathbf{F}$ is the temporal ``eventually'' operator. Its nesting expresses nondecreasing temporal order, and $A,\ldots,E$ are local abbreviations defined separately for each specification. Instruction 2's ball referent is evaluated on views rendered from promoted $W_1$. The selected record retains \texttt{blue\_ball\_obj0}; the saved plan binds $W_1$, its checkpoint, grounding record, and planning cloud.

The first manipulation has 572 frames (323 pick and 249 release); the second has 751 (443 pick and 308 place). A plan-execution manifest assigns every final plan step a unique artifact and verifies the plan and artifact digests. The central collision checker covers 184 pick frames and all 249 release frames for task 1, and 92 pick frames and all 308 place frames for task 2. Approach, attachment, contact, and release are also checked under their declared policies.

\section{Observation and Reconciliation Protocol}
\label{app:reconciliation}

Each observation frame records intrinsics, a rigid camera transform in the \method world frame, a fit or holdout role, and hashed RGB, metric depth, and masks. Physical bundle types cannot carry synthetic-pixel masks; synthetic bundles require one for every frame. Implemented physical adapters accept registered robot/OSG frames, Record3D, COLMAP, or an already prepared bundle.

For the reported study, rendered geometry determines source-disocclusion and target-insertion masks. Generated color is allowed only inside those masks, after which source pixels are restored exactly outside them. Target depth remains the predicted metric depth rather than generated depth. Stage 1 contains 21 image-edit requests (12 source and 9 target); stage 2 contains 24 (12 source and 12 target). Requests, model identity, masks, and results are content-addressed and cached.

Reconciliation inserts 12,170 unowned disocclusion seeds at stage 1 and 3,551 at stage 2, activates mask-supported scene regions, and optimizes the selected Gaussians. Densification and pruning are disabled for action-owned rows. Predicted owner means, scales, orientations, and opacities are restored after fitting; spherical-harmonic appearance may change. The demonstrated candidates require non-increasing changed-region L1 and an increase of at most $10^{-4}$ in outside-region L1, both measured against the same synthetic targets on validation views.

\section{Complete State Sequence}
\label{app:ledger}

\begin{table}[h]
\centering
\caption{The five-state mainline produced by the sequential garden study.}
\small
\begin{tabularx}{0.96\textwidth}{@{}p{0.20\textwidth}p{0.13\textwidth}p{0.12\textwidth}X@{}}
\toprule
State & Revision & Status & Direct evidence \\
\midrule
\texttt{t0000-observed} & observed & historical & 6,125,086 Gaussian rows from the captured garden. \\
\texttt{t0001-predicted} & predicted & historical & 254 moved rows; 6,124,832 inactive rows exact. \\
\texttt{t0001-\allowbreak reconciled-\allowbreak syn.} & reconciled-synthetic & historical & 12,170 seeds; improved held-out masked L1; preserved action geometry. \\
\texttt{t0002-predicted} & predicted & historical & 254 moved rows; 6,137,002 inactive rows exact. \\
\texttt{t0002-\allowbreak reconciled-\allowbreak syn.} & reconciled-synthetic & current & 3,551 seeds; improved held-out masked L1; final task views agree. \\
\bottomrule
\end{tabularx}
\end{table}

At $W_1$, stable-identity association has 15 detection supports, lies 0.0187933\,m from the predicted pose, and lies 3.7198327\,m from the old pose. Instruction 2's pick location equals the re-grounded position. At $W_2$, the twin reports final ball position $[1.0219336,\,0.0055544,\,1.4529843]$ in run world units; checkpoint ownership, twin, registry, and planning cloud agree.

\bibliographystyle{apalike}
\bibliography{IEEEabrv,mainref}
\clearpage

\end{document}